\documentclass{article}

\usepackage{microtype}
\usepackage{pifont} 
\usepackage{booktabs}
\usepackage[table]{xcolor}
\usepackage{graphicx}
\usepackage{animate}
\usepackage{booktabs}
\usepackage{fancyhdr}
\usepackage{algorithm}
\usepackage{amssymb}
\usepackage{subcaption}
\usepackage{tikz}
\usepackage{array}
\usepackage[utf8]{inputenc}
\usepackage{tcolorbox}

\usepackage{hyperref}
\usepackage{fourier}
\usepackage{multirow}

\usepackage[accepted]{icml2025}
\usepackage{amsmath}
\usepackage{amssymb}
\usepackage{mathtools}
\usepackage{amsthm}
\usepackage{tocbibind}
\usepackage{arydshln}

\usepackage[capitalize,noabbrev]{cleveref}
\theoremstyle{plain}

\theoremstyle{definition}

\theoremstyle{remark}

\usepackage[textsize=tiny]{todonotes}

\begin{document}

\twocolumn[
\icmltitle{Señorita-2M: A High-Quality Instruction-based Dataset for General Video Editing by Video Specialists}

\begin{icmlauthorlist}
\icmlauthor{Bojia Zi$^{*}$}{cuhk}
\icmlauthor{Penghui Ruan$^{*}$}{polyu}
\icmlauthor{Marco Chen}{thu}
\icmlauthor{Xianbiao Qi$^{\dagger}$}{illf}
\icmlauthor{Shaozhe Hao}{hku}
\icmlauthor{Shihao Zhao}{hku}
\icmlauthor{Youze Huang}{uestc}
\icmlauthor{Bin Liang}{cuhk}
\icmlauthor{Rong Xiao}{illf}
\icmlauthor{Kam-Fai Wong}{cuhk}
\end{icmlauthorlist}

\icmlaffiliation{cuhk}{The Chinese University of Hong Kong}
\icmlaffiliation{polyu}{The Hong Kong Polytechnic University}
\icmlaffiliation{thu}{Tsinghua University}
\icmlaffiliation{hku}{The University of Hong Kong}
\icmlaffiliation{illf}{IntelliFusion Inc.}
\icmlaffiliation{uestc}{University of Electronic Science and Technology of China}

\icmlcorrespondingauthor{Xianbiao Qi}{qixianbiao@gmail.com}

\icmlkeywords{Machine Learning, ICML}

\vskip 0.3in
]
\printAffiliationsAndNotice{\icmlEqualContribution}

\newcommand{\ie}{i.e., }

\begin{figure*}[!htp]
  \centering

    \begin{minipage}{1.05\textwidth}
    \begin{subfigure}{0.49\textwidth}
    \begin{subfigure}{0.48\textwidth}
      \animategraphics[width=1\textwidth]{8}{data/teasers/remove_human/12/frame_}{0}{15}
    \end{subfigure}
    \begin{subfigure}{0.48\textwidth}
      \animategraphics[width=1\textwidth]{8}{data/teasers/remove_human/output_0/frame_}{0}{15}
    \end{subfigure}
    \vspace{-0.6em}
    \subcaption*{\emph{Remove the girl.}}
    \end{subfigure}
    \begin{subfigure}{0.49\textwidth}
    \begin{subfigure}{0.48\textwidth}
      \animategraphics[width=1\textwidth]{8}{data/teasers/swap_for_cat/0/frame_}{0}{15}
    \end{subfigure}
    \begin{subfigure}{0.48\textwidth}
      \animategraphics[width=1\textwidth]{8}{data/teasers/swap_for_cat/output_0/frame_}{0}{15}
    \end{subfigure}
    \vspace{-0.6em}
    \subcaption*{\emph{Swap the tiger for cat.}}
    \end{subfigure}
    \end{minipage}

    \begin{minipage}{1.05\textwidth}
    \begin{subfigure}{0.49\textwidth}
    \begin{subfigure}{0.48\textwidth}
      \animategraphics[width=1\textwidth]{8}{data/teasers/add_hat/2/frame_}{0}{16}
    \end{subfigure}
    \begin{subfigure}{0.48\textwidth}
      \animategraphics[width=1\textwidth]{8}{data/teasers/add_hat/output_0/frame_}{0}{16}
    \end{subfigure}
    \vspace{-0.6em}
    \subcaption*{\emph{Add a hat on girl's head.}}
    \end{subfigure}
    \begin{subfigure}{0.49\textwidth}
    \begin{subfigure}{0.48\textwidth}
      \animategraphics[width=1\textwidth]{8}{data/teasers/make_it_anime_style/4/frame_}{0}{16}
    \end{subfigure}
    \begin{subfigure}{0.48\textwidth}
      \animategraphics[width=1\textwidth]{8}{data/teasers/make_it_anime_style/output_0/frame_}{0}{15}
    \end{subfigure}
    \vspace{-0.6em}
    \subcaption*{\emph{Make it anime style.}}
    \end{subfigure}
    \end{minipage}

    \begin{minipage}{1.05\textwidth}
    \begin{subfigure}{0.49\textwidth}
    \begin{subfigure}{0.48\textwidth}
      \animategraphics[width=1\textwidth]{8}{data/teasers/make_it_watercolor2/16/frame_}{0}{16}
    \end{subfigure}
    \begin{subfigure}{0.48\textwidth}
      \animategraphics[width=1\textwidth]{8}{data/teasers/make_it_watercolor2/output_0/frame_}{0}{16}
    \end{subfigure}
    \vspace{-0.6em}
    \subcaption*{\emph{Make it watercolor style.}}
    \end{subfigure}
    \begin{subfigure}{0.49\textwidth}
    \begin{subfigure}{0.48\textwidth}
      \animategraphics[width=1\textwidth]{8}{data/teasers/add_rainbow/8/frame_}{0}{16}
    \end{subfigure}
    \begin{subfigure}{0.48\textwidth}
      \animategraphics[width=1\textwidth]{8}{data/teasers/add_rainbow/output_0/frame_}{0}{15}
    \end{subfigure}
    \vspace{-0.6em}
    \subcaption*{\emph{Add a rainbow.}}
    \end{subfigure}
    
    \end{minipage}
    
    \vspace{-1em}
\caption{The visual results given by editing models trained on our Señorita-2M. \emph{Best viewed with Acrobat Reader. Click the images to play the animation clips.}}
\end{figure*}

\begin{abstract}

Recent advancements in video generation have spurred the development of video editing techniques, which can be divided into inversion-based and end-to-end methods. However, current video editing methods still suffer from several challenges. Inversion-based methods, though training-free and flexible, are time-consuming during inference, struggle with fine-grained editing instructions, and produce artifacts and jitter. On the other hand, end-to-end methods, which rely on edited video pairs for training, offer faster inference speeds but often produce poor editing results due to a lack of high-quality training video pairs. In this paper, to close the gap in end-to-end methods, we introduce Señorita-2M, a high-quality video editing dataset. Señorita-2M consists of approximately 2 millions of video editing pairs. It is built by crafting four high-quality, specialized video editing models, each crafted and trained by our team to achieve state-of-the-art editing results. We also propose a filtering pipeline to eliminate poorly edited video pairs. Furthermore, we explore common video editing architectures to identify the most effective structure based on current pre-trained generative model. Extensive experiments show that our dataset can help to yield remarkably high-quality video editing results. More details are available at \url{https://senorita-2m-dataset.github.io}.

\end{abstract}

\section{Introduction}

In recent years, diffusion-based generative techniques have made significant strides~\citep{sora, pikalab, keling, gen3, chen2024videocrafter2, xing2025dynamicrafter, cogvideox_yang2024cogvideox, hunyuanvideosystematicframeworklargekong2025, stablediffusion_blattmann2023align}. Models like Stable 
Diffusion~\citep{stablediffusion_blattmann2023align, animatediff_guo2023animatediff} and Kolors~\citep{kolors}, which use the UNet~\citep{unet_ronneberger2015u} architecture, have achieved excellent text-to-image generation results. More recently, Pixart~\citep{pixart_chen2023} and Flux~\citep{flux} have employed the DiT~\citep{dit_peebles2023scalable} architecture to create powerful text-to-image models. Meanwhile, video generation has also advanced rapidly. Open-source models like VideoCrafter~\citep{videocrafter2_chen2024videocrafter2} and AnimateDiff~\citep{animatediff_guo2023animatediff} have garnered attention for their impressive visual effects. Other models, such as CogVideoX~\citep{cogvideox_yang2024cogvideox} and HunyuanVideo~\citep{hunyuanvideosystematicframeworklargekong2025}, with more parameters and training data, have surpassed previous models in motion consistency and visual quality. Additionally, closed-source models like SORA~\citep{sora}, Kling~\citep{keling}, and Gen3~\citep{gen3} have captivated users with their exceptional performance in video production. Simultaneously, editing techniques have also progressed significantly. Image editing has been widely studied and has yielded excellent results, but video editing, studied more recently, still requires further development to achieve satisfactory outcomes.

Image editing can be categorized into two types: Inversion-based methods~\citep{ddim_based_gal2022image, ddim_based_kawar2023imagic, ddim_based_parmar2023zero, ddim_based_tumanyan2023plug} and End-to-end methods~\citep{emu_edit_sheynin2024emu, omni_citation_geng2024instructdiffusion, omniedit_wei2024omniedit, ultraedit_zhao2024ultraedit, hq_edit_hui2024hq, instructpix2pix_brooks2023instructpix2pix, magic_brush_zhang2024magicbrush}. Inversion-based methods rely on converting the image to latent and then edited by a prompt. In contrast, end-to-end methods are trained on image editing datasets, often yielding more pleasing results. Among these end-to-end methods, the quality and number of training pairs play an important role for their effectiveness. Instruct-pix2pix~\citep{instructpix2pix_brooks2023instructpix2pix} uses data from a diffusion model for training, enabling diffusion to edit images. MagicBrush~\citep{magic_brush_zhang2024magicbrush} introduces manually annotated editing data, enhancing the capabilities of the diffusion model. EmuEdit~\citep{emu_edit_sheynin2024emu} surpasses previous methods by using smaller biases and higher-quality data for training. UltraEdit~\citep{ultraedit_zhao2024ultraedit} constructs a large-scale dataset using an inpainting model and Inversion methods. Omni-Edit~\citep{omniedit_wei2024omniedit} improved UltraEdit by employing more experts to generate higher-quality datasets, resulting in better-performing models.

The field of video editing differs significantly from image editing, with most video editing techniques being inversion-based, while only a few belong to the latter category. Inversion-based methods typically require long editing durations and often result in frame inconsistencies in the edited videos. As a result, end-to-end methods have gained increasing popularity and attention. InsV2V~\citep{insv2v_cheng2024consistent} trains an editing model using generated video pairs, while Revideo~\citep{revideo_mou2024revideo} utilizes motion and content to control the generated video output. Propgen~\citep{propgen_liu2024generative} supervises model training with inexpensive video masks, and ViVid-10M~\citep{vivid_10m_hu2024vivid} provides region editing data from both images and videos, training an inpainting model. \textbf{\emph{However, these methods suffer from poor performance due to the shortage of high-quality instruction-based editing dataset.}}

To address the issue of data shortage, we build a dataset by using high-quality video editing experts. Specially, we trained four high-quality video editing experts using CogVideoX~\citep{cogvideox_yang2024cogvideox}: a global stylizer, a local stylizer, an inpainting model, and a remover. These experts, along with other specialized models, are used to construct a large-scale dataset of high-quality video editing samples. Additionally, we designed a filtering pipeline that effectively removes failed video samples. We also utilized a large language model to convert video editing prompts, achieving clear and effective instructions. As a result, our dataset, \textbf{Señorita-2M}, contains approximately 2 million high-quality video editing pairs. Furthermore, we trained multiple video editors based on different video editing architectures using this dataset to evaluate the effectiveness of various editing frameworks, ultimately achieving impressive editing capabilities.

Our main contributions can be summarized into three folders:

\begin{itemize}
    \item We introduce \textbf{Señorita-2M}, the first truly large-scale instruction-based video editing dataset. Existing datasets either focus on local edits (\ie RACCooN and VIVID-10M) or are synthetically generated (\ie InsV2V). In contrast, our dataset comprises two million video pairs, with the original data sourced from the Internet.

    \item To build \textbf{Señorita-2M} dataset, we craft four expert models with each model specializing in a particular editing task, \ie global stylizer, local stylizer, object remover and object swap. Each expert achieves state-of-the-art performance in its own task. 

    \item Experiments have shown that our dataset can help to train a high-quality video editing model. The resulting model demonstrates high visual quality, strong frame consistency and text alignment. Our dataset and models will be open-sourced upon acceptance.

\end{itemize}

\begin{figure*}[!ht]
    \centering
    \includegraphics[width=0.92\linewidth]{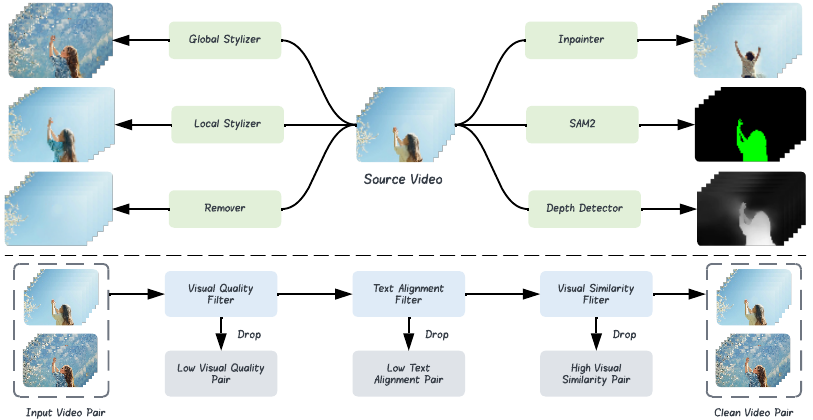}
\caption{Top: The data construction pipeline of the Señorita-2M dataset. Bottom: The filtering pipeline of Señorita-2M. Further details are provided in the Appendix \ref{sec_dataset_construction_global_editing}.}
    \label{fig:dataset_construction}
\end{figure*}

\renewcommand{\arraystretch}{1.4}
\setlength{\tabcolsep}{5pt}

\begin{table*}[!ht]\small
\centering 
\caption{Comparison between Señorita-2M, InsV2V, VIVID-10M. InsV2V uses VideoP2P~\citep{videop2p_liu2023video} to create their dataset. Señorita-2M uses videos downloaded from Pexels~\citep{pexels}, produced by editing experts and vision experts.}
\begin{tabular}{ccccccccc} 
\toprule \textbf{Datasets} & \textbf{Sources} & \textbf{Editing Types} & \textbf{Experts}  & \textbf{Frames} & \textbf{Resolution}  &  \textbf{Real Videos} &\textbf{Edited Pairs} &  \textbf{OpenSource} \\ \hline

VIVID-10M & Panda-70M &Local & 1  & 30 & $1280\times 720$  &73,737 & 1.5M & \ding{55}\\ 
InsV2V & Synthesis & Free-Form  & 1 & 16 & $256\times256$  & 0 & 0.06M& \ding{52} \\
\midrule
\multirow{2}{*}{Señorita-2M} & Crawled\vspace{-0.6em} & \multirow{2}{*}{Local+Global} & \multirow{2}{*}{6}  & \multirow{2}{*}{33 - 64} & $336\times592$ & \multirow{2}{*}{388,909} & \multirow{2}{*}{2M} & \multirow{2}{*}{\ding{52}} \\
& from Internet &&&&- $1120\times1984$&&& \\
\bottomrule

\end{tabular} 
\label{tab:dataset_parameter_comparison} 
\end{table*}

\section{Related Works}
\subsection{Image and Video Editing}

\textbf{Image Editing.}
Recent image editing methods have emerged, such as DDIM inversion, which edits by converting images to latent space and adding prompts to regenerate them. Research has focused on reducing discretization errors in the inversion process~\cite{other_ddim_huberman2024edit, other_ddim_lu2022dpm, other_ddim_wallace2023edict}. SDEdit~\cite{sdedit_meng2021sdedit} introduces noise to images and denoises them according to a target text. Prompt-to-Prompt~\cite{prompt_to_prompt_hertz2022prompt} modifies attention maps during diffusion steps. Null-Text Inversion~\cite{null_text_inversion_mokady2023null} adjusts textual embeddings for classifier-free guidance. Recent supervised methods, including InstructP2P~\cite{instructpix2pix_brooks2023instructpix2pix}, HIVE~\cite{hive_zhang2024hive}, and MagicBrush~\cite{magic_brush_zhang2024magicbrush}, integrate well-crafted instructions within end-to-end frameworks.

\textbf{Video Editing.}
Video editing has gained great attention from the public~\cite{fatezero_qi2023fatezero, stablev2v_liu2024stablev2v}. Tune-A-Video~\cite{tuneavideo_wu2023tune} fine-tunes diffusion models on specific videos to generate edited videos based on target prompts. Methods like Pix2Video~\cite{pix2video_ceylan2023pix2video} and TokenFlow~\cite{tokenflow_geyer2023tokenflow} focus on consistency across frames by using attention across frames or editing key frames. AnyV2V~\cite{ku2024anyv2v} generates edited videos by injecting features, guided by the first frame. New models like Gen3~\cite{gen3} and SORA~\cite{sora} perform style transfer through adding noise and regenerating by target prompts. In contrast, few video editing approaches use supervised methods. InsV2V~\cite{insv2v_cheng2024consistent} trains on video pairs, while EVE~\cite{eve_singer2025video} uses an SDS loss~\cite{dreamfusion_poole2022dreamfusion} for distillation. RACCooN~\cite{raccon_yoon2024raccoonremoveaddchange} and VIVID-10M~\cite{vivid_10m_hu2024vivid} use inpainting models and video annotations to produce local editing models. Similarly, Propgen~\cite{propgen_liu2024generative} is used for local editing, applies segmentation models to propagate edits across frames.

\subsection{Image and Video Editing Datasets}
Image editing datasets often rely on synthetic data. InstructPix2Pix~\cite{instructpix2pix_brooks2023instructpix2pix} introduced CLIP-score-based prompt-to-prompt filtering to build large-scale datasets. MagicBrush~\cite{magic_brush_zhang2024magicbrush} improves data quality with human annotations from DALLE-2~\cite{dalle2_ramesh2022hierarchical}, while HQ-Edit~\cite{hq_edit_hui2024hq} uses DALLE-3~\cite{dalle3_betker2023improving} for high-quality edited pairs. Emu-Edit~\cite{emu_edit_sheynin2024emu} expanded its dataset to 10 million image pairs, combining free-form and local editing. UltraEdit~\cite{ultraedit_zhao2024ultraedit} contributed 4 million samples with LLM-generated instructions, blending creativity with human input. Omni-Edit~\cite{omniedit_wei2024omniedit} diversified editing capabilities using multiple expert models and multimodal frameworks for quality control.

In contrast, only a few video editing datasets exist. RACCooN~\cite{raccon_yoon2024raccoonremoveaddchange} and VIVID-10M use inpainting models for video annotation. InsV2V~\cite{insv2v_cheng2024consistent} builds its dataset with pairs of generated original and target videos, though the data quality was insufficient for strong performance.

\section{Methodology}
This section outlines the construction methods of four video experts: global stylizer, local stylizer, text-guided video inpainter, and object remover. Besides, we also introduce the pipeline for building our Señorita-2M, which  includes data collection, the inference processes for local and global video pairs, and the filtering pipeline. More details are shown in Appendix \ref{sec_design_of_video_editing_experts_appdendix}, \ref{sec_app_senorita_2m_dataset} and \ref{sec_data_selection}.

\subsection{The Construction of Video Experts}
\subsubsection{The Training Data for Video Experts}
We use Webvid-10M~\citep{webvid_bain2021frozen} dataset for training. CogVLM2~\citep{Cogvlm2_hong2024cogvlm2} generates captions, each with around 50 words, and recognizes objects in the videos. These objects are segmented and tracked using Grounded-SAM2~\citep{groundingdino_liu2023grounding, sam2_ravi2024sam2}.

\subsubsection{The Design and Training for Video Experts}
\textbf{Global Stylizer}. The current video generation models struggle to understand the style prompt. Thus, the controlnet built on these generation models cannot perform the stylization according to the text prompt. To improve, we first edit the initial frame with an image ControlNet (ControlNet-SD1.5~\citep{controlnet_zhang2023adding}) and then guide the video ControlNet to complete the remaining frames. The video ControlNet uses multiple control conditions to get robust style transfer results\cite{unicontrolnet_zhao2024uni}, including Canny, HED, and Depth, each transformed into latent space via 3D-VAE. More details are in the Appendix \ref{sec_app_global_stylizer}.

\textbf{Local Stylizer}. Inspired by the inpainting methods, such as AVID~\citep{avid_zhang2023avid}, we trained a local stylizer using both inpainting and ControlNet. The model uses three control conditions, same as the global stylizer, inputted into the ControlNet branch. Besides, the mask conditions are fed into the main branch. The pretrained model used is CogVideoX-2B. More details are in the Appendix \ref{sec_app_local_stylizer}.

\textbf{Text-guided Video Inpainter}. Existing methods like AVID~\citep{avid_zhang2023avid} and COCOCO~\citep{cococo_zi2024cococo} suffer from outdated models, causing artifacts. Besides, the VIVID-10M has not been opensoureced. Therefore, we trained an inpainter based on CogVideoX-5B-I2V, guided by a first frame edited with Flux-Fill~\citep{flux}. The inpainter was trained with four types of masks to avoid overfitting, including random and precise shapes. More details are in the Appendix \ref{sec_app_video_inpainter}.

\textbf{Video Remover.} Current video inpainters like Propinater~\citep{zhou2023propainter} generate blur when removing objects, which highly reduces its usability. Thus, we trained a powerful video remover based on CogVideoX-2B, using a novel mask selection strategy. 90\% of masks are randomly sampled from unrelated videos with positive instructions, while 10\% precisely cover objects with negative instructions. After training, classifier-free guidance is used with both types of instructions. This results in content generation unrelated to the mask shape. More details are in the Appendix \ref{sec_app_video_inpainter}.

\begin{figure*}[!ht]
  \centering

    \begin{minipage}{0.02\textwidth}
        \rotatebox{90}{\scriptsize \centering (a) Local Stylization}
    \end{minipage}
    \begin{minipage}{0.94\textwidth}
    \begin{subfigure}{0.49\textwidth}
    \begin{subfigure}{0.48\textwidth}
      \animategraphics[width=1\textwidth]{8}{data/local_style_transfer/1179810-hd_1280_720_30fps_mask_id_1_0_0_guidance_scale_6.0_output1215_org/image_}{0}{15}
    \end{subfigure}
    \begin{subfigure}{0.48\textwidth}
      \animategraphics[width=1\textwidth]{8}{data/local_style_transfer/1179810-hd_1280_720_30fps_mask_id_1_0_0_guidance_scale_6.0_output1215/image_}{0}{15}
    \end{subfigure}
    \vspace{-0.6em}
    \subcaption*{\emph{Paint the purple flower pink color.}}
    \end{subfigure}
    \begin{subfigure}{0.49\textwidth}
    \begin{subfigure}{0.48\textwidth}
      \animategraphics[width=1\textwidth]{8}{data/local_style_transfer/4039046-hd_1280_720_30fps_mask_id_1_0_0_guidance_scale_6.0_output1657_org/image_}{0}{15}
    \end{subfigure}
    \begin{subfigure}{0.48\textwidth}
      \animategraphics[width=1\textwidth]{8}{data/local_style_transfer/4039046-hd_1280_720_30fps_mask_id_1_0_0_guidance_scale_6.0_output1657/image_}{0}{15}
    \end{subfigure}
    \vspace{-0.6em}
    \subcaption*{\emph{Paint this cat's fur purple.}}
    \end{subfigure}
    \end{minipage}

    \begin{minipage}{0.02\textwidth}
        \rotatebox{90}{\scriptsize  \centering (b) Object Removal}
    \end{minipage}
    \begin{minipage}{0.94\textwidth}
    \begin{subfigure}{0.49\textwidth}
    \begin{subfigure}{0.48\textwidth}
      \animategraphics[width=1\textwidth]{8}{data/obj_removal/obj-3/4321851-hd_1280_720_25fps_mask_id_1_0_0_guidance_scale_2.0_output1961_org/image_}{0}{15}
    \end{subfigure}
    \begin{subfigure}{0.48\textwidth}
      \animategraphics[width=1\textwidth]{8}{data/obj_removal/obj-3/4321851-hd_1280_720_25fps_mask_id_1_0_0_guidance_scale_2.0_output1961/image_}{0}{15}
    \end{subfigure}
    \vspace{-0.6em}
    \subcaption*{\emph{Remove/Add a flowers.}}
    \end{subfigure}
    \begin{subfigure}{0.49\textwidth}
    \begin{subfigure}{0.48\textwidth}
      \animategraphics[width=1\textwidth]{8}{data/obj_removal/obj-2/4265399-hd_1280_720_30fps_mask_id_1_0_0_guidance_scale_2.0_output1080_org/image_}{0}{15}
    \end{subfigure}
    \begin{subfigure}{0.48\textwidth}
      \animategraphics[width=1\textwidth]{8}{data/obj_removal/obj-2/4265399-hd_1280_720_30fps_mask_id_1_0_0_guidance_scale_2.0_output1080/image_}{0}{15}
    \end{subfigure}
    \vspace{-0.6em}
    \subcaption*{\emph{Remove/Add a man.}}
    \end{subfigure}
    
    \end{minipage}

    \begin{minipage}{0.02\textwidth}
        \rotatebox{90}{\scriptsize \centering  (c) Style Transfer}
    \end{minipage}
    \begin{minipage}{0.94\textwidth}
    \begin{subfigure}{0.49\textwidth}
    \begin{subfigure}{0.48\textwidth}
      \animategraphics[width=1\textwidth]{8}{data/style_transfer/van_gogh/transfer-2/9133278-hd_2048_1080_25fps_style_id_14_0_0_guidance_scale_6_output5178_org/image_}{0}{15}
    \end{subfigure}
    \begin{subfigure}{0.48\textwidth}
      \animategraphics[width=1\textwidth]{8}{data/style_transfer/van_gogh/transfer-2/9133278-hd_2048_1080_25fps_style_id_14_0_0_guidance_scale_6_output5178/image_}{0}{15}
    \end{subfigure}
    \vspace{-0.6em}
    \subcaption*{\emph{Turn it Van Gogh style.}}
    \end{subfigure}
    \begin{subfigure}{0.49\textwidth}
    \begin{subfigure}{0.48\textwidth}
      \animategraphics[width=1\textwidth]{8}{data/style_transfer/waterpainting/transfer-2/7248428-hd_1280_720_24fps_style_id_266_0_0_guidance_scale_6_output7583_org/image_}{0}{15}
    \end{subfigure}
    \begin{subfigure}{0.48\textwidth}
      \animategraphics[width=1\textwidth]{8}{data/style_transfer/waterpainting/transfer-2/7248428-hd_1280_720_24fps_style_id_266_0_0_guidance_scale_6_output7583/image_}{0}{15}
    \end{subfigure}
    \vspace{-0.6em}
    \subcaption*{\emph{Make it watercolor style.}}
    \end{subfigure}

    \end{minipage}

    \begin{minipage}{0.02\textwidth}
        \rotatebox{90}{\scriptsize \centering (d) Object Swap}
    \end{minipage}
    \begin{minipage}{0.94\textwidth}
    \begin{subfigure}{0.49\textwidth}
    \begin{subfigure}{0.48\textwidth}
      \animategraphics[width=1\textwidth]{8}{data/obj_swap/animal/best/6853331-hd_2048_1080_25fps_mask_id_1_0_0_guidance_scale_6.0_output79_org/image_}{0}{15}
    \end{subfigure}
    \begin{subfigure}{0.48\textwidth}
      \animategraphics[width=1\textwidth]{8}{data/obj_swap/animal/best/6853331-hd_2048_1080_25fps_mask_id_1_0_0_guidance_scale_6.0_output79/image_}{0}{15}
    \end{subfigure}
    \vspace{-0.6em}
    \subcaption*{\emph{Turn the cat into bear.}}
    \end{subfigure}
    \begin{subfigure}{0.49\textwidth}
    \begin{subfigure}{0.48\textwidth}
      \animategraphics[width=1\textwidth]{8}{data/obj_swap/food/4978863-hd_1280_720_30fps_mask_id_1_0_0_guidance_scale_6.0_output22_org/image_}{0}{15}
    \end{subfigure}
    \begin{subfigure}{0.48\textwidth}
      \animategraphics[width=1\textwidth]{8}{data/obj_swap/food/4978863-hd_1280_720_30fps_mask_id_1_0_0_guidance_scale_6.0_output22/image_}{0}{15}
    \end{subfigure}
    \vspace{-0.6em}
    \subcaption*{\emph{Turn the light into fire.}}
    \end{subfigure}
    \end{minipage}

    \begin{minipage}{0.02\textwidth}
        \rotatebox{90}{\scriptsize \centering\ \ \ \ \ \ (e) Others }
    \end{minipage}
    \begin{minipage}{0.94\textwidth}
    \begin{subfigure}{0.49\textwidth}
    \begin{subfigure}{0.48\textwidth}
      \animategraphics[width=1\textwidth]{8}{data/object_detection/videos/4105436-uhd_3840_2160_25fps_org/image_}{0}{15}
    \end{subfigure}
    \begin{subfigure}{0.48\textwidth}
      \animategraphics[width=1\textwidth]{8}{data/object_detection/videos/4105436-uhd_3840_2160_25fps/image_}{0}{15}
    \end{subfigure}
    \vspace{-0.6em}
    \subcaption*{\emph{Ground the women.}}
    \end{subfigure}
    \begin{subfigure}{0.49\textwidth}
    \begin{subfigure}{0.48\textwidth}
      \animategraphics[width=1\textwidth]{8}{data/object_detection/videos/8055354-hd_1280_720_30fps_org/image_}{0}{15}
    \end{subfigure}
    \begin{subfigure}{0.48\textwidth}
      \animategraphics[width=1\textwidth]{8}{data/object_detection/videos/8055354-hd_1280_720_30fps/image_}{0}{15}
    \end{subfigure}
    \vspace{-0.6em}
    \subcaption*{\emph{Detect depth.}}
    \end{subfigure}
    \end{minipage}
    \vspace{-1em}
\caption{Visualization of our Señorita-2M. \emph{Best viewed with Acrobat Reader. Click the images to play the animation clips.}}

\label{fig:visualization_of_senorita}
\end{figure*}

\begin{figure*}[htp]
\centering
\begin{subfigure}{0.475\textwidth}
\includegraphics[width=0.95\textwidth]{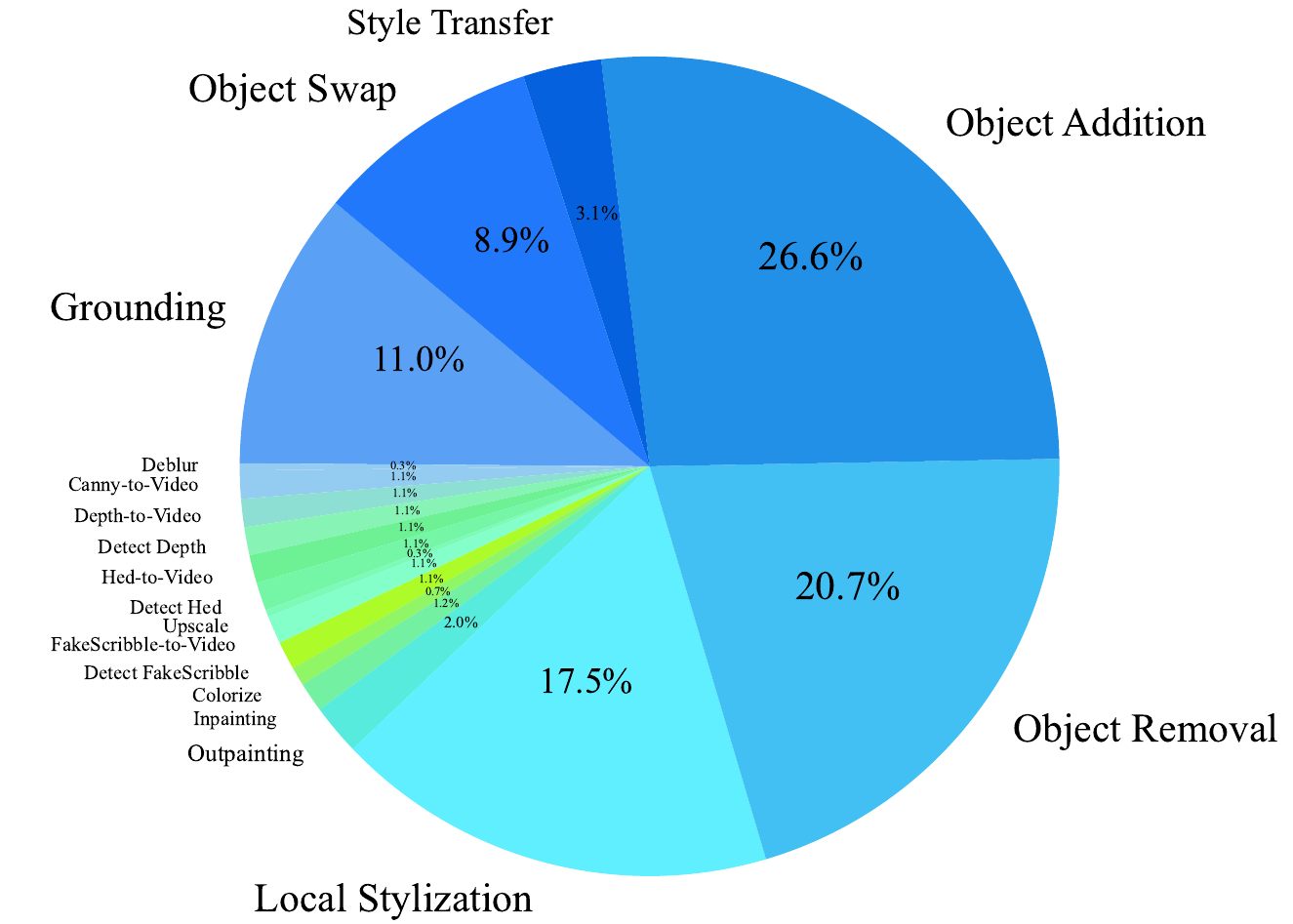}
\subcaption{Distribution of editing types.}
\end{subfigure}
\begin{subfigure}{0.475\textwidth}
\includegraphics[width=0.975\textwidth]{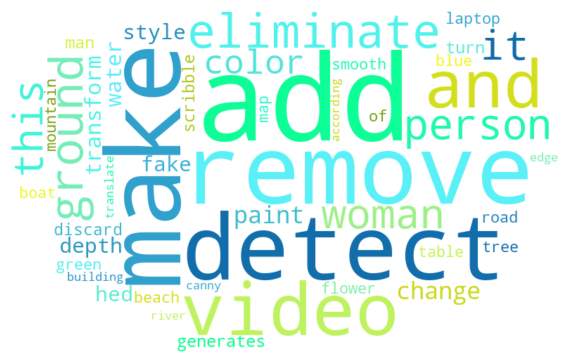}
\subcaption{Word cloud for the frequency of terms.}
\end{subfigure}
\vspace{-0.1in}
\caption{Overview of Señorita-2M dataset: a statistical analysis.}

\label{fig:pie_chart2}
\vspace{-0.175in}
\end{figure*}

\subsection{The Construction of Señorita-2M}
Here, we introduce the data source and two types editing tasks, including both local editing and global editing tasks. The comparison between different datasets are provided in Table \ref{tab:dataset_parameter_comparison}. The visual results of our dataset is provided in Figure \ref{fig:visualization_of_senorita}, while  statistical analysis of our dataset is in Figure \ref{fig:pie_chart2}. 
\subsubsection{The Data Source in Señorita-2M}
We crawled videos from Pexels~\citep{pexels}, a video-sharing website with high-resolution and high quality videos, by authenticated APIs. The total number of videos in this part is around 390,000. We use the BLIP-2~\citep{blip2_li2023blip} to caption the videos, in order to carter to the length restriction of CLIP. Besides, the mask regions and their corresponding phrases are obtained by CogVLM2\cite{Cogvlm2_hong2024cogvlm2} and Grounded-SAM2\cite{groundingdino_liu2023grounding, sam2_ravi2024sam2}.

\subsubsection{Local Edit}

Local editing includes 6 tasks: object swap, local style transfer, object addition, object removal, inpainting, and outpainting. More details can be found in the Appendix \ref{sec_dataset_construction_local_editing}.

\textbf{Object Swap.} Object swap uses FLUX-Fill and our trained inpainter. To begin with, the LLaMA-3~\citep{llama3_dubey2024llama} suggests a replacement object, which is then swapped in the first frame by FLUX-Fill. The inpainter generates the remaining frames guided by the first. Finally, the LLM generates instructions referring to both the old and new objects.

\textbf{Local Style Transfer.} We construct a prompt by asking the LLM to add descriptive adjectives to the object name. This prompt is fed into the local stylizer to modify the masked region, and the LLM converts it into the final instruction.

\textbf{Object Removal.} Our remover model is used for object removal. Positive and negative instructions are generated with ``Remove'' or ``Generate'' added before the object name. We use classifier-free guidance technique to use the both positive and negative instruction for inference of our remover. The LLM then generates the instruction, referring the inference instructions.

\textbf{Object Addition.} The object addition task is the reverse of object removal, where the source and target videos are swapped. LLM assists in rewriting and enhancing the instructions.

\textbf{Video Inpainting and Outpainting.} For inpainting, a region is removed from the first frame and replaced with zeros. The masked region’s position is shifted over time. Instructions are generated by adding ``inpaint'' before the caption. Outpainting is similar but uses a black background, with instructions prefixed by ``outpaint''.

\subsubsection{Global Edit}

Our global edit involves three key components: 1. Style transfer 2. Object grounding 3. Conditional generation. More details are shown in the Appendix \ref{sec_dataset_construction_global_editing}.

\textbf{Style Transfer.} We began by combining style prompts provided by Midjourney~\citep{midjourney} with BLIP-2 captions to generate the prompts with style information, which were then input into ControlNet-SD1.5-HED for style transfer on the first video frame. We integrated the edited first frame with control conditions, including canny, depth, and hed, and processed it through our global stylizer to generate the rest frames. Finally, an LLM was used to convert the style prompts into actionable instructions for further content optimization.

\textbf{Object Grounding.} We provide video pairs for object grounding to help the video editor accurately localize relevant regions according to instructions. Areas unrelated to the prompt are marked in black, while prompt-related instances are highlighted in distinct colors. Initial instructions are created by prepending words like ``Detect'' or ``Ground'' before the object name, and an LLM is then used to refine these instructions.

\textbf{Conditional Generation.} This component consists of 10 tasks aimed at supporting video-to-video translation, including: Deblur, Canny-to-Video, Depth-to-Video, Depth Detection, Hed-to-Video, Hed Detection, Upscaling, FakeScribble-to-Video, FakeScribble Detection, and Colorization.

\subsubsection{Data Selection}

We proposed a filtering pipeline to select proper edited videos. First, we apply quality filtering to recognize successful edits, followed by removal of poorly aligned text. Finally, we exclude videos with subtle or no changes.

\textbf{Quality filtering}. We trained classifiers to filter out failure cases. First, we manually annotated a dataset to classify failed and successful samples. The classifier training began by extracting features using a frozen CLIP vision encoder from 17 frames per video. These features were then classified with MLP classifiers. To enhance robustness, we ensembled classifiers trained with various strategies. Different thresholds were applied for different tasks.

\textbf{Removing poor text-alignment videos}. We found that some generated content misaligned with the text prompt. To address this, we compared edited samples with their corresponding text prompts using CLIP. For object swapping and local stylization, we compared the inpainting prompt with the edited video, and for stylization, we compared the style prompt with the edited video. Since object removal lacks a suitable prompt, no comparison was made. In practice, different thresholds are used for different tasks.

\textbf{Removing subtle video pairs}. Some video pairs show subtle edits or regenerate content similar to the original. These could cause overfitting during training. To filter these out, we used CLIP’s vision encoder to extract features and compare the original and edited videos, excluding pairs with a similarity score above a set threshold.

\renewcommand{\arraystretch}{1.1}
\setlength{\tabcolsep}{5pt}

\begin{table*}[!ht]
\vspace{-0.175in}
\centering 
\caption{Comparison with previous methods. The best results are \textbf{blodfaced}.}
\begin{tabular}{c|ccc|c} 
\toprule 
Methods &  \textbf{Ewarp}($10^{-3}$) ($\downarrow$) & \textbf{CLIPScore} ($\uparrow$) & \textbf{Temporal Consistency} ($\uparrow$) & \textbf{User Preference} ($\uparrow$) \\ \hline

Tokenflow & 16.31  & 0.2637 & 0.9752 & 6.74\% \\ 
Flatten & 16.31  & 0.2461 & 0.9690 & 5.95\% \\ 
AnyV2V & 20.48  & 0.2723 & 0.9709 & 19.40\% \\ 
\hline
\hline
InsV2V & 16.50  & 0.1675 & 0.9727 & 14.68\% \\
\hline
\rowcolor{gray!30} Ours & \textbf{9.42}  & \textbf{0.2895} & \textbf{0.9775} & \textbf{53.17\%} \\ 
\hline
\bottomrule

\end{tabular} 
\label{tab:comparison1} 
\vspace{-0.175in}
\end{table*}

\section{Experiments}
This section provides the construction details of the dataset, training details of the editing models and the experimental results and analysis.
More information can be found in the Appendix \ref{sec_design_of_video_editing_experts_appdendix}.

\subsection{Construction of Dataset}
The dataset construction pipeline involves four steps: First, videos are  collected and annotated. Next, editing experts are applied to the videos. Instructions are then generated using LLMs, followed by a filtering pipeline to remove failure cases.

\textbf{Data preparation}. The original videos are resized to 336$\times$592 or 592$\times$336 according to their aspect ratio. The brief descriptions obtained by BLIP-2, 810K masks and corresponding phrases are prepared for inference.

\textbf{Inference of experts}. The inference was performed using Nvidia 4090 GPUs. The Local Stylizer was configured with a classifier-free guidance (CFG) of 6 and a resolution of 336$\times$592, processing 33 frames. The Global Stylizer was set to a resolution of 256$\times$496 and then resized to 336$\times$592, also handling 33 frames. The Inpainter, using a CFG of 6 and a resolution of 336$\times$592, processed 33 frames as well. The Remover was set with a CFG of 2 and a resolution of 336$\times$592. Additionally, depth estimators, HED, Canny detectors, and other computer vision techniques were employed to generate video pairs, all at a resolution of 1120$\times$1984.

\begin{table*}[!ht]
\centering 
\vspace{-0.175in}
\caption{The results of the ablation study are presented. All three models are fine-tuned based on CogVideoX-5B. \textbf{Temp-Cons} is the abbreviated form of Temporal Consistency. The best results are \textbf{blodfaced}.}
\begin{tabular}{c|cccccc} 
\toprule Methods & \textbf{Dataset} & \textbf{Training Samples} & \textbf{Epochs} &\textbf{Ewarp}($10^{-3}$) ($\downarrow$)   & \textbf{CLIPScore} ($\uparrow$) & \textbf{Temp-Cons} ($\uparrow$) \\ \hline

Ablation-1 & InsV2V & 60K & 8 & 8.51 & 0.2366 & 0.9712  \\
Ablation-2 & Señorita-2M & 60K & 8 & 8.44 & 0.2596 & 0.9783  \\
Ablation-3 & Señorita-2M & 120K & 4 & \textbf{7.95} & \textbf{0.2641} & \textbf{0.9785}  \\
\hline
\bottomrule

\end{tabular} 
\label{tab:ab_study} 
\vspace{-0.2in}
\end{table*}

\textbf{Instruction generation}. Instructions are generated by LLMs\cite{llama3_dubey2024llama}, transforming source and target object names or editing prompts into clear instructions. 

\textbf{Determining the source and target videos}. In the Object Swap task, the edited video serves as the source, and the original as the target. The Object Addition task follows a similar approach. For Object Removal, local and global stylization, the edited video is the target.

\textbf{Filtering pipeline}. We first apply a quality filter to remove the failure cases with the threshold of 0.6. Then, we use CLIP to compare the similarity and remove those with low similarity. The object removal and local stylization use the threshold of 0.22. While the global stylization and object addition use threshold of 0.2. Finally, CLIP is used to compare the difference between the original and edited videos, removing those with value higher than 0.95. More details can be found in the Appendix \ref{sec_data_selection}.

\subsection{Training Details of Editing Model}
We use CogVideoX-5B-I2V\cite{cogvideox_yang2024cogvideox} as the base model and integrate it with ControlNet to leverage the edited first frame to guide the editing process. The batch size of the editing model's training is 32, the learning rate is 1e-5, and weight decay is 1e-4. We train model for 2 epoch. We sample 33 frames of the videos to train with a resolution of 336$\times$592 in first stage. Different from the first stage, we use higher resolution of 448$\times$768 and batch size of 16 in stage two, fintuning with 1 epoch to help model edit high resolution.

\subsection{Experimental Results}
We compared the editing model trained on our dataset with previous editing methods to demonstrate the effectiveness of our dataset. Additionally, we conducted an ablation study to show that our editing dataset significantly aids in training an effective editor. Furthermore, we conducted experiments by training 6 models with different architectures to understand the impact of architectures on editing.

\begin{table*}[!htp]
\centering 
\caption{Exploration of different editing architectures. \textit{Ins-Edit} refers to the InstructPix2Pix architecture, \textit{Control-Edit} denotes the ControlNet architecture for video editing. $^*$ indicates the use of the Omni-Edit dataset for enhancement. \textit{FF-} are first-frame guided editing models. The best results are \textbf{boldfaced}.}
\vspace{0.15em}
\begin{tabular}{c|ccc|c} 
\toprule Methods &  \textbf{Ewarp}($10^{-3}$) ($\downarrow$)& \textbf{CLIPScore} ($\uparrow$) & \textbf{Temporal Consistency} ($\uparrow$) & \textbf{User Preference} ($\uparrow$)  \\ \hline

Ins-Edit & 13.18  & 0.2648 & 0.9797 & 3.87\% \\ 
Control-Edit & 12.81  & 0.2882 & 0.9769 & 14.40\%\\ 
Ins-Edit$^*$ & 13.83  & 0.2789 & 0.9784 & 8.86\% \\ 
Control-Edit$^*$ & 10.46  & 0.2866 & \textbf{0.9802} & 23.26\% \\ 
FF-Ins-Edit & \textbf{8.44}  & 0.2861 & 0.9783 & 12.46\% \\ 
FF-Control-Edit & 9.42  & \textbf{0.2895} & 0.9775 & \textbf{37.12\%} \\ 
\hline
\bottomrule

\end{tabular} 
\label{tab:6_models_comparison} 
\vspace{-0.2in}
\end{table*}

\subsubsection{Quantitative Comparison}

For model evaluation, we utilize the DAVIS dataset\cite{davis2017pont} with randomly generated editing prompts. We evaluate the stability of the edited videos using Ewarp and Temporal Consistency, while the CLIPScore is used to assess the text-video alignment. To evaluate Ewarp, we use a resolution of $336 \times 592$. CLIP is used to extract features from each frame, and we compute the similarity between adjacent frames for Temporal Consistency.

Based on the provided experimental results in Table~\ref{tab:comparison1}, our method outperforms all others across several metrics. Specifically, in terms of Ewarp, our method achieves the lowest value at 9.42, significantly outperforming Tokenflow, Flatten, InsV2V, and AnyV2V, which have higher Ewarp values. In terms of CLIPScore, our method also leads with a score of 0.2895, surpassing the highest score from AnyV2V, which is 0.2723. Furthermore, our method excels in Temporal Consistency with a score of 0.9775, higher than the other approaches.

To further demonstrate the effectiveness of our dataset, we compare our results with InsV2V, as both methods require video pairs to train editors, while the other editing methods are zero-shot. In terms of Ewarp, InsV2V achieves a value of 16.50, significantly worse than our 9.42. In CLIPScore, InsV2V scores 0.1675, which is notably lower than our 0.2895. Additionally, in Temporal Consistency, InsV2V scores 0.9727, whereas our method leads with 0.9775. These comparisons highlight that our approach consistently outperforms InsV2V across all metrics, demonstrating the superior quality and effectiveness of our dataset in achieving better editing performance.

\subsubsection{Qualitative Results}

We conducted a user study to determine which video users preferred. The sequence of videos in the questionnaire was randomly shuffled to ensure fairness. As shown in Table \ref{tab:comparison1}, our method significantly outperforms all previous approaches, achieving an impressive user preference score of 53.17\%, which is notably higher than the next best score of 19.40\%. This highlights the superior appeal and relevance of our method in meeting user needs. Moreover, the results showcase the effectiveness of our datasets, solidifying their superiority over existing alternatives. The Figure \ref{visual_comparison} in Appendix \ref{sec_visual_comparison} provides the visual comparison results.

\subsubsection{Ablation Study}
The ablation study in Table~\ref{tab:ab_study} demonstrates that utilizing samples from the Señorita-2M dataset enhances the model's performance. Compared to experiments using videos from InsV2V dataset, the experiment with  Señorita-2M, conducted over the same number of epochs, yields superior results. Specifically, the CLIPScore improves from 0.2366 to 0.2596, and Temporal Consistency increases from 0.9712 to 0.9783. Additionally, increasing the number of training samples from 60K to 120K, sampled from the Señorita-2M dataset, leads to further improvements. The CLIPScore rises to 0.2641, while Temporal Consistency reaches 0.9785, and Ewarp decreases from 8.44 to 7.95. These changes suggest better text alignment and reduced warping errors. Overall, these results indicate that a larger and more diverse dataset significantly enhances the model's ability to learn a broader range of editing capabilities, improving both consistency and text alignment.

\subsubsection{Different Editing Architectures}
We explore different editing architectures by training 6 instruction-based models with various architectures. We draw on two widely used image editing architectures, InstructPix2Pix~\citep{instructpix2pix_brooks2023instructpix2pix} and ControlNet~\citep{controlnet_zhang2023adding}. Specifically, InstructPix2Pix concatenates conditions and input latents, outputting predicted noise, while ControlNet uses a control branch for editing conditions and a main branch for input latents. We also investigate strategies with and without first frame guidance. Additionally, we enhance the models by incorporating the Omni-Edit dataset\cite{omniedit_wei2024omniedit}.

We compare different video editing architectures, such as InstructPix2Pix, ControlNet, and their enhanced and first-frame guided versions. The results are shown in Table \ref{tab:6_models_comparison}. Control-Edit with the Omni-Edit dataset achieves the highest temporal consistency at 0.9802 and a user preference of 23.26\%. The first-frame guided models show notable improvements, with FF-Control-Edit achieving the highest user preference at 37.12\%. FF-Ins-Edit has the lowest error warp at 8.44. Without dataset enhancement, Ins-Edit and Control-Edit show similar results in user preference, with Control-Edit slightly ahead at 14.40\% compared to 3.87\%. These results indicate that first-frame guidance and dataset enhancements significantly boost the performance of video editing models.

\section{Conclusion}

In this paper, we trained a set of advanced video editing models and integrated them with various computer vision experts to create a high-quality, instruction-based video editing dataset. This dataset includes 18 distinct video editing tasks, comprising approximately 2 million video pairs in various resolutions and frame lengths. To ensure best video quality, we applied a cascade of multiple filtering algorithms. Additionally, we employed a large language model to transform prompts and object names into precise editing instructions. To validate the dataset's effectiveness, we trained four editing models using four widely adopted editing architectures. Experimental results demonstrate that our dataset can effectively produce high-quality video editing models, achieving notable improvements in visual quality, frame consistency, and text alignment.

\section*{Impact Statement}
In this paper, we propose a dataset conducted for training general video editing models. The real videos in the dataset are legally sourced from Pexels.com using their authenticated APIs. The dataset itself does not pose any harm to the community. However, models trained on this dataset is capable of editing videos. Fortunately, this risk could be reduced by deepfake detection methods.

\bibliography{main}

\begin{thebibliography}{69}
\providecommand{\natexlab}[1]{#1}
\providecommand{\url}[1]{\texttt{#1}}
\expandafter\ifx\csname urlstyle\endcsname\relax
  \providecommand{\doi}[1]{doi: #1}\else
  \providecommand{\doi}{doi: \begingroup \urlstyle{rm}\Url}\fi

\bibitem[pex()]{pexels}
https://www.pexels.com/.
\newblock URL \url{https://www.pexels.com/}.

\bibitem[pik()]{pika}
https://www.pika.art/.
\newblock URL \url{https://www.pika.art/}.

\bibitem[Bain et~al.(2021)Bain, Nagrani, Varol, and Zisserman]{webvid_bain2021frozen}
Bain, M., Nagrani, A., Varol, G., and Zisserman, A.
\newblock Frozen in time: A joint video and image encoder for end-to-end retrieval.
\newblock In \emph{Proceedings of the IEEE/CVF International Conference on Computer Vision}, pp.\  1728--1738, 2021.

\bibitem[Betker et~al.(2023)Betker, Goh, Jing, Brooks, Wang, Li, Ouyang, Zhuang, Lee, Guo, et~al.]{dalle3_betker2023improving}
Betker, J., Goh, G., Jing, L., Brooks, T., Wang, J., Li, L., Ouyang, L., Zhuang, J., Lee, J., Guo, Y., et~al.
\newblock Improving image generation with better captions.
\newblock \emph{Computer Science. https://cdn.openai.com/papers/dall-e-3. pdf}, 2\penalty0 (3):\penalty0 8, 2023.

\bibitem[{Black Forest Labs}(2024)]{flux}
{Black Forest Labs}.
\newblock Black forest labs.
\newblock \url{https://github.com/black-forest-labs/flux/}, 2024.

\bibitem[Blattmann et~al.(2023)Blattmann, Rombach, Ling, Dockhorn, Kim, Fidler, and Kreis]{stablediffusion_blattmann2023align}
Blattmann, A., Rombach, R., Ling, H., Dockhorn, T., Kim, S.~W., Fidler, S., and Kreis, K.
\newblock Align your latents: High-resolution video synthesis with latent diffusion models.
\newblock In \emph{Proceedings of the IEEE/CVF Conference on Computer Vision and Pattern Recognition}, pp.\  22563--22575, 2023.

\bibitem[Brooks et~al.(2023)Brooks, Holynski, and Efros]{instructpix2pix_brooks2023instructpix2pix}
Brooks, T., Holynski, A., and Efros, A.~A.
\newblock Instructpix2pix: Learning to follow image editing instructions.
\newblock In \emph{Proceedings of the IEEE/CVF Conference on Computer Vision and Pattern Recognition}, pp.\  18392--18402, 2023.

\bibitem[Ceylan et~al.(2023)Ceylan, Huang, and Mitra]{pix2video_ceylan2023pix2video}
Ceylan, D., Huang, C.-H.~P., and Mitra, N.~J.
\newblock Pix2video: Video editing using image diffusion.
\newblock In \emph{Proceedings of the IEEE/CVF International Conference on Computer Vision}, pp.\  23206--23217, 2023.

\bibitem[Chen et~al.(2024{\natexlab{a}})Chen, Zhang, Cun, Xia, Wang, Weng, and Shan]{chen2024videocrafter2}
Chen, H., Zhang, Y., Cun, X., Xia, M., Wang, X., Weng, C., and Shan, Y.
\newblock Videocrafter2: Overcoming data limitations for high-quality video diffusion models.
\newblock In \emph{Proceedings of the IEEE/CVF Conference on Computer Vision and Pattern Recognition}, 2024{\natexlab{a}}.

\bibitem[Chen et~al.(2024{\natexlab{b}})Chen, Zhang, Cun, Xia, Wang, Weng, and Shan]{videocrafter2_chen2024videocrafter2}
Chen, H., Zhang, Y., Cun, X., Xia, M., Wang, X., Weng, C., and Shan, Y.
\newblock Videocrafter2: Overcoming data limitations for high-quality video diffusion models.
\newblock \emph{arXiv preprint arXiv:2401.09047}, 2024{\natexlab{b}}.

\bibitem[Chen et~al.(2023)Chen, Yu, Ge, Yao, Xie, Wu, Wang, Kwok, Luo, Lu, et~al.]{pixart_chen2023}
Chen, J., Yu, J., Ge, C., Yao, L., Xie, E., Wu, Y., Wang, Z., Kwok, J., Luo, P., Lu, H., et~al.
\newblock Pixart-alpha: Fast training of diffusion transformer for photorealistic text-to-image synthesis.
\newblock \emph{arXiv preprint arXiv:2310.00426}, 2023.

\bibitem[Cheng et~al.(2024)Cheng, Xiao, and He]{insv2v_cheng2024consistent}
Cheng, J., Xiao, T., and He, T.
\newblock Consistent video-to-video transfer using synthetic dataset.
\newblock In \emph{The Twelfth International Conference on Learning Representations}, 2024.
\newblock URL \url{https://openreview.net/forum?id=IoKRezZMxF}.

\bibitem[Dubey et~al.(2024)Dubey, Jauhri, Pandey, Kadian, Al-Dahle, Letman, Mathur, Schelten, Yang, Fan, et~al.]{llama3_dubey2024llama}
Dubey, A., Jauhri, A., Pandey, A., Kadian, A., Al-Dahle, A., Letman, A., Mathur, A., Schelten, A., Yang, A., Fan, A., et~al.
\newblock The llama 3 herd of models.
\newblock \emph{arXiv preprint arXiv:2407.21783}, 2024.

\bibitem[Gal et~al.(2022)Gal, Alaluf, Atzmon, Patashnik, Bermano, Chechik, and Cohen-Or]{ddim_based_gal2022image}
Gal, R., Alaluf, Y., Atzmon, Y., Patashnik, O., Bermano, A.~H., Chechik, G., and Cohen-Or, D.
\newblock An image is worth one word: Personalizing text-to-image generation using textual inversion.
\newblock \emph{arXiv preprint arXiv:2208.01618}, 2022.

\bibitem[{Gen-3}(2024)]{gen3}
{Gen-3}.
\newblock Introducing gen-3 alpha: A new frontier for video generation.
\newblock \url{https://runwayml.com/research/introducing-gen-3-alpha/}, 2024.

\bibitem[Geng et~al.(2024)Geng, Yang, Hang, Li, Gu, Zhang, Bao, Zhang, Li, Hu, et~al.]{omni_citation_geng2024instructdiffusion}
Geng, Z., Yang, B., Hang, T., Li, C., Gu, S., Zhang, T., Bao, J., Zhang, Z., Li, H., Hu, H., et~al.
\newblock Instructdiffusion: A generalist modeling interface for vision tasks.
\newblock In \emph{Proceedings of the IEEE/CVF Conference on Computer Vision and Pattern Recognition}, pp.\  12709--12720, 2024.

\bibitem[Geyer et~al.(2023)Geyer, Bar-Tal, Bagon, and Dekel]{tokenflow_geyer2023tokenflow}
Geyer, M., Bar-Tal, O., Bagon, S., and Dekel, T.
\newblock Tokenflow: Consistent diffusion features for consistent video editing.
\newblock \emph{arXiv preprint arXiv:2307.10373}, 2023.

\bibitem[Guo et~al.(2023{\natexlab{a}})Guo, Yang, Rao, Agrawala, Lin, and Dai]{guo2023sparsectrl}
Guo, Y., Yang, C., Rao, A., Agrawala, M., Lin, D., and Dai, B.
\newblock Sparsectrl: Adding sparse controls to text-to-video diffusion models, 2023{\natexlab{a}}.

\bibitem[Guo et~al.(2023{\natexlab{b}})Guo, Yang, Rao, Wang, Qiao, Lin, and Dai]{animatediff_guo2023animatediff}
Guo, Y., Yang, C., Rao, A., Wang, Y., Qiao, Y., Lin, D., and Dai, B.
\newblock Animatediff: Animate your personalized text-to-image diffusion models without specific tuning.
\newblock \emph{arXiv preprint arXiv:2307.04725}, 2023{\natexlab{b}}.

\bibitem[Hertz et~al.(2022)Hertz, Mokady, Tenenbaum, Aberman, Pritch, and Cohen-Or]{prompt_to_prompt_hertz2022prompt}
Hertz, A., Mokady, R., Tenenbaum, J., Aberman, K., Pritch, Y., and Cohen-Or, D.
\newblock Prompt-to-prompt image editing with cross attention control.
\newblock \emph{arXiv preprint arXiv:2208.01626}, 2022.

\bibitem[Hong et~al.(2024)Hong, Wang, Ding, Yu, Lv, Wang, Cheng, Huang, Ji, Xue, et~al.]{Cogvlm2_hong2024cogvlm2}
Hong, W., Wang, W., Ding, M., Yu, W., Lv, Q., Wang, Y., Cheng, Y., Huang, S., Ji, J., Xue, Z., et~al.
\newblock Cogvlm2: Visual language models for image and video understanding.
\newblock \emph{arXiv preprint arXiv:2408.16500}, 2024.

\bibitem[Hu et~al.(2024)Hu, Zhong, Wang, Jiang, Tian, Yang, Wan, and Zhang]{vivid_10m_hu2024vivid}
Hu, J., Zhong, T., Wang, X., Jiang, B., Tian, X., Yang, F., Wan, P., and Zhang, D.
\newblock Vivid-10m: A dataset and baseline for versatile and interactive video local editing.
\newblock \emph{arXiv preprint arXiv:2411.15260}, 2024.

\bibitem[Huberman-Spiegelglas et~al.(2024)Huberman-Spiegelglas, Kulikov, and Michaeli]{other_ddim_huberman2024edit}
Huberman-Spiegelglas, I., Kulikov, V., and Michaeli, T.
\newblock An edit friendly ddpm noise space: Inversion and manipulations.
\newblock In \emph{Proceedings of the IEEE/CVF Conference on Computer Vision and Pattern Recognition}, pp.\  12469--12478, 2024.

\bibitem[Hui et~al.(2024)Hui, Yang, Zhao, Shi, Wang, Wang, Zhou, and Xie]{hq_edit_hui2024hq}
Hui, M., Yang, S., Zhao, B., Shi, Y., Wang, H., Wang, P., Zhou, Y., and Xie, C.
\newblock Hq-edit: A high-quality dataset for instruction-based image editing.
\newblock \emph{arXiv preprint arXiv:2404.09990}, 2024.

\bibitem[Kawar et~al.(2023)Kawar, Zada, Lang, Tov, Chang, Dekel, Mosseri, and Irani]{ddim_based_kawar2023imagic}
Kawar, B., Zada, S., Lang, O., Tov, O., Chang, H., Dekel, T., Mosseri, I., and Irani, M.
\newblock Imagic: Text-based real image editing with diffusion models.
\newblock In \emph{Proceedings of the IEEE/CVF Conference on Computer Vision and Pattern Recognition}, pp.\  6007--6017, 2023.

\bibitem[{Keling}(2024)]{keling}
{Keling}.
\newblock Kling video model.
\newblock \url{https://kling.kuaishou.com/en}, 2024.

\bibitem[Kong et~al.(2025)Kong, Tian, Zhang, Min, Dai, Zhou, Xiong, Li, Wu, Zhang, Wu, Lin, Yuan, Long, Wang, Wang, Li, Huang, Yang, Tan, Wang, Song, Bai, Wu, Xue, Wang, Wang, Liu, Li, Li, Wang, Yu, Deng, Li, Chen, Cui, Peng, Yu, He, Xu, Zhou, Xu, Tao, Lu, Liu, Zhou, Wang, Yang, Wang, Liu, Jiang, and Zhong]{hunyuanvideosystematicframeworklargekong2025}
Kong, W., Tian, Q., Zhang, Z., Min, R., Dai, Z., Zhou, J., Xiong, J., Li, X., Wu, B., Zhang, J., Wu, K., Lin, Q., Yuan, J., Long, Y., Wang, A., Wang, A., Li, C., Huang, D., Yang, F., Tan, H., Wang, H., Song, J., Bai, J., Wu, J., Xue, J., Wang, J., Wang, K., Liu, M., Li, P., Li, S., Wang, W., Yu, W., Deng, X., Li, Y., Chen, Y., Cui, Y., Peng, Y., Yu, Z., He, Z., Xu, Z., Zhou, Z., Xu, Z., Tao, Y., Lu, Q., Liu, S., Zhou, D., Wang, H., Yang, Y., Wang, D., Liu, Y., Jiang, J., and Zhong, C.
\newblock Hunyuanvideo: A systematic framework for large video generative models, 2025.
\newblock URL \url{https://arxiv.org/abs/2412.03603}.

\bibitem[Ku et~al.(2024)Ku, Wei, Ren, Yang, and Chen]{ku2024anyv2v}
Ku, M., Wei, C., Ren, W., Yang, H., and Chen, W.
\newblock Anyv2v: A plug-and-play framework for any video-to-video editing tasks.
\newblock \emph{arXiv preprint arXiv:2403.14468}, 2024.

\bibitem[Lee et~al.(2024)Lee, Cho, Shin, Lee, Yang, and Lee]{video_diffusion_inpainter_lee2024video}
Lee, M., Cho, S., Shin, C., Lee, J., Yang, S., and Lee, S.
\newblock Video diffusion models are strong video inpainter.
\newblock \emph{arXiv preprint arXiv:2408.11402}, 2024.

\bibitem[Li et~al.(2023)Li, Li, Savarese, and Hoi]{blip2_li2023blip}
Li, J., Li, D., Savarese, S., and Hoi, S.
\newblock Blip-2: Bootstrapping language-image pre-training with frozen image encoders and large language models.
\newblock In \emph{International conference on machine learning}, pp.\  19730--19742. PMLR, 2023.

\bibitem[Liu et~al.(2024{\natexlab{a}})Liu, Li, Zhang, Lan, and Liu]{stablev2v_liu2024stablev2v}
Liu, C., Li, R., Zhang, K., Lan, Y., and Liu, D.
\newblock Stablev2v: Stablizing shape consistency in video-to-video editing.
\newblock \emph{arXiv preprint arXiv:2411.11045}, 2024{\natexlab{a}}.

\bibitem[Liu et~al.(2023{\natexlab{a}})Liu, Zeng, Ren, Li, Zhang, Yang, Li, Yang, Su, Zhu, et~al.]{groundingdino_liu2023grounding}
Liu, S., Zeng, Z., Ren, T., Li, F., Zhang, H., Yang, J., Li, C., Yang, J., Su, H., Zhu, J., et~al.
\newblock Grounding dino: Marrying dino with grounded pre-training for open-set object detection.
\newblock \emph{arXiv preprint arXiv:2303.05499}, 2023{\natexlab{a}}.

\bibitem[Liu et~al.(2023{\natexlab{b}})Liu, Zhang, Li, Lin, and Jia]{videop2p_liu2023video}
Liu, S., Zhang, Y., Li, W., Lin, Z., and Jia, J.
\newblock Video-p2p: Video editing with cross-attention control.
\newblock \emph{arXiv preprint arXiv:2303.04761}, 2023{\natexlab{b}}.

\bibitem[Liu et~al.(2024{\natexlab{b}})Liu, Wang, Wang, Liu, Zhang, Lee, Li, Yu, Lin, Kim, et~al.]{propgen_liu2024generative}
Liu, S., Wang, T., Wang, J.-H., Liu, Q., Zhang, Z., Lee, J.-Y., Li, Y., Yu, B., Lin, Z., Kim, S.~Y., et~al.
\newblock Generative video propagation.
\newblock \emph{arXiv preprint arXiv:2412.19761}, 2024{\natexlab{b}}.

\bibitem[Loshchilov \& Hutter(2017)Loshchilov and Hutter]{adamw_loshchilov2017decoupled}
Loshchilov, I. and Hutter, F.
\newblock Decoupled weight decay regularization.
\newblock \emph{arXiv preprint arXiv:1711.05101}, 2017.

\bibitem[Lu et~al.(2022)Lu, Zhou, Bao, Chen, Li, and Zhu]{other_ddim_lu2022dpm}
Lu, C., Zhou, Y., Bao, F., Chen, J., Li, C., and Zhu, J.
\newblock Dpm-solver++: Fast solver for guided sampling of diffusion probabilistic models.
\newblock \emph{arXiv preprint arXiv:2211.01095}, 2022.

\bibitem[Meng et~al.(2021)Meng, He, Song, Song, Wu, Zhu, and Ermon]{sdedit_meng2021sdedit}
Meng, C., He, Y., Song, Y., Song, J., Wu, J., Zhu, J.-Y., and Ermon, S.
\newblock Sdedit: Guided image synthesis and editing with stochastic differential equations.
\newblock \emph{arXiv preprint arXiv:2108.01073}, 2021.

\bibitem[{Midjourney}(2024)]{midjourney}
{Midjourney}.
\newblock Midjourney.
\newblock \url{https://www.midjourney.com/}, 2024.

\bibitem[Mokady et~al.(2023)Mokady, Hertz, Aberman, Pritch, and Cohen-Or]{null_text_inversion_mokady2023null}
Mokady, R., Hertz, A., Aberman, K., Pritch, Y., and Cohen-Or, D.
\newblock Null-text inversion for editing real images using guided diffusion models.
\newblock In \emph{Proceedings of the IEEE/CVF Conference on Computer Vision and Pattern Recognition}, pp.\  6038--6047, 2023.

\bibitem[Mou et~al.(2024)Mou, Cao, Wang, Zhang, Shan, and Zhang]{revideo_mou2024revideo}
Mou, C., Cao, M., Wang, X., Zhang, Z., Shan, Y., and Zhang, J.
\newblock Revideo: Remake a video with motion and content control.
\newblock \emph{arXiv preprint arXiv:2405.13865}, 2024.

\bibitem[OpenAI(2024)]{sora}
OpenAI.
\newblock Sora: Creating video from text.
\newblock \url{https://openai.com/index/sora/}, 2024.

\bibitem[Parmar et~al.(2023)Parmar, Kumar~Singh, Zhang, Li, Lu, and Zhu]{ddim_based_parmar2023zero}
Parmar, G., Kumar~Singh, K., Zhang, R., Li, Y., Lu, J., and Zhu, J.-Y.
\newblock Zero-shot image-to-image translation.
\newblock In \emph{ACM SIGGRAPH 2023 Conference Proceedings}, pp.\  1--11, 2023.

\bibitem[Peebles \& Xie(2023)Peebles and Xie]{dit_peebles2023scalable}
Peebles, W. and Xie, S.
\newblock Scalable diffusion models with transformers.
\newblock In \emph{Proceedings of the IEEE/CVF International Conference on Computer Vision}, pp.\  4195--4205, 2023.

\bibitem[Pont-Tuset et~al.(2017)Pont-Tuset, Perazzi, Caelles, Arbel{\'a}ez, Sorkine-Hornung, and Van~Gool]{davis2017pont}
Pont-Tuset, J., Perazzi, F., Caelles, S., Arbel{\'a}ez, P., Sorkine-Hornung, A., and Van~Gool, L.
\newblock The 2017 davis challenge on video object segmentation.
\newblock \emph{arXiv preprint arXiv:1704.00675}, 2017.

\bibitem[Poole et~al.(2022)Poole, Jain, Barron, and Mildenhall]{dreamfusion_poole2022dreamfusion}
Poole, B., Jain, A., Barron, J.~T., and Mildenhall, B.
\newblock Dreamfusion: Text-to-3d using 2d diffusion.
\newblock \emph{arXiv preprint arXiv:2209.14988}, 2022.

\bibitem[Qi et~al.(2023)Qi, Cun, Zhang, Lei, Wang, Shan, and Chen]{fatezero_qi2023fatezero}
Qi, C., Cun, X., Zhang, Y., Lei, C., Wang, X., Shan, Y., and Chen, Q.
\newblock Fatezero: Fusing attentions for zero-shot text-based video editing.
\newblock \emph{arXiv preprint arXiv:2303.09535}, 2023.

\bibitem[Radford et~al.(2021)Radford, Kim, Hallacy, Ramesh, Goh, Agarwal, Sastry, Askell, Mishkin, Clark, et~al.]{clip_radford2021learning}
Radford, A., Kim, J.~W., Hallacy, C., Ramesh, A., Goh, G., Agarwal, S., Sastry, G., Askell, A., Mishkin, P., Clark, J., et~al.
\newblock Learning transferable visual models from natural language supervision.
\newblock In \emph{International conference on machine learning}, pp.\  8748--8763. PMLR, 2021.

\bibitem[Ramesh et~al.(2021)Ramesh, Pavlov, Goh, Gray, Voss, Radford, Chen, and Sutskever]{clip_ramesh2021zero}
Ramesh, A., Pavlov, M., Goh, G., Gray, S., Voss, C., Radford, A., Chen, M., and Sutskever, I.
\newblock Zero-shot text-to-image generation.
\newblock In \emph{International Conference on Machine Learning}, pp.\  8821--8831. PMLR, 2021.

\bibitem[Ramesh et~al.(2022)Ramesh, Dhariwal, Nichol, Chu, and Chen]{dalle2_ramesh2022hierarchical}
Ramesh, A., Dhariwal, P., Nichol, A., Chu, C., and Chen, M.
\newblock Hierarchical text-conditional image generation with clip latents.
\newblock \emph{arXiv preprint arXiv:2204.06125}, 1\penalty0 (2):\penalty0 3, 2022.

\bibitem[Ravi et~al.(2024)Ravi, Gabeur, Hu, Hu, Ryali, Ma, Khedr, R{\"a}dle, Rolland, Gustafson, Mintun, Pan, Alwala, Carion, Wu, Girshick, Doll{\'a}r, and Feichtenhofer]{sam2_ravi2024sam2}
Ravi, N., Gabeur, V., Hu, Y.-T., Hu, R., Ryali, C., Ma, T., Khedr, H., R{\"a}dle, R., Rolland, C., Gustafson, L., Mintun, E., Pan, J., Alwala, K.~V., Carion, N., Wu, C.-Y., Girshick, R., Doll{\'a}r, P., and Feichtenhofer, C.
\newblock Sam 2: Segment anything in images and videos.
\newblock \emph{arXiv preprint arXiv:2408.00714}, 2024.
\newblock URL \url{https://arxiv.org/abs/2408.00714}.

\bibitem[Ronneberger et~al.(2015)Ronneberger, Fischer, and Brox]{unet_ronneberger2015u}
Ronneberger, O., Fischer, P., and Brox, T.
\newblock U-net: Convolutional networks for biomedical image segmentation.
\newblock In \emph{Medical Image Computing and Computer-Assisted Intervention--MICCAI 2015: 18th International Conference, Munich, Germany, October 5-9, 2015, Proceedings, Part III 18}, pp.\  234--241. Springer, 2015.

\bibitem[Sheynin et~al.(2024)Sheynin, Polyak, Singer, Kirstain, Zohar, Ashual, Parikh, and Taigman]{emu_edit_sheynin2024emu}
Sheynin, S., Polyak, A., Singer, U., Kirstain, Y., Zohar, A., Ashual, O., Parikh, D., and Taigman, Y.
\newblock Emu edit: Precise image editing via recognition and generation tasks.
\newblock In \emph{Proceedings of the IEEE/CVF Conference on Computer Vision and Pattern Recognition}, pp.\  8871--8879, 2024.

\bibitem[Singer et~al.(2025)Singer, Zohar, Kirstain, Sheynin, Polyak, Parikh, and Taigman]{eve_singer2025video}
Singer, U., Zohar, A., Kirstain, Y., Sheynin, S., Polyak, A., Parikh, D., and Taigman, Y.
\newblock Video editing via factorized diffusion distillation.
\newblock In \emph{European Conference on Computer Vision}, pp.\  450--466. Springer, 2025.

\bibitem[Team(2024)]{kolors}
Team, K.
\newblock Kolors: Effective training of diffusion model for photorealistic text-to-image synthesis.
\newblock \emph{arXiv preprint}, 2024.

\bibitem[Tumanyan et~al.(2023)Tumanyan, Geyer, Bagon, and Dekel]{ddim_based_tumanyan2023plug}
Tumanyan, N., Geyer, M., Bagon, S., and Dekel, T.
\newblock Plug-and-play diffusion features for text-driven image-to-image translation.
\newblock In \emph{Proceedings of the IEEE/CVF Conference on Computer Vision and Pattern Recognition}, pp.\  1921--1930, 2023.

\bibitem[Wallace et~al.(2023)Wallace, Gokul, and Naik]{other_ddim_wallace2023edict}
Wallace, B., Gokul, A., and Naik, N.
\newblock Edict: Exact diffusion inversion via coupled transformations.
\newblock In \emph{Proceedings of the IEEE/CVF Conference on Computer Vision and Pattern Recognition}, pp.\  22532--22541, 2023.

\bibitem[Wei et~al.(2024)Wei, Xiong, Ren, Du, Zhang, and Chen]{omniedit_wei2024omniedit}
Wei, C., Xiong, Z., Ren, W., Du, X., Zhang, G., and Chen, W.
\newblock Omniedit: Building image editing generalist models through specialist supervision.
\newblock \emph{arXiv preprint arXiv:2411.07199}, 2024.

\bibitem[Wu et~al.(2023)Wu, Ge, Wang, Lei, Gu, Shi, Hsu, Shan, Qie, and Shou]{tuneavideo_wu2023tune}
Wu, J.~Z., Ge, Y., Wang, X., Lei, S.~W., Gu, Y., Shi, Y., Hsu, W., Shan, Y., Qie, X., and Shou, M.~Z.
\newblock Tune-a-video: One-shot tuning of image diffusion models for text-to-video generation.
\newblock In \emph{Proceedings of the IEEE/CVF International Conference on Computer Vision}, pp.\  7623--7633, 2023.

\bibitem[Xing et~al.(2025)Xing, Xia, Zhang, Chen, Yu, Liu, Liu, Wang, Shan, and Wong]{xing2025dynamicrafter}
Xing, J., Xia, M., Zhang, Y., Chen, H., Yu, W., Liu, H., Liu, G., Wang, X., Shan, Y., and Wong, T.-T.
\newblock Dynamicrafter: Animating open-domain images with video diffusion priors.
\newblock In \emph{European Conference on Computer Vision}, 2025.

\bibitem[Yang et~al.(2024)Yang, Teng, Zheng, Ding, Huang, Xu, Yang, Zhang, Gu, Feng, Yin, Hong, Wang, Cheng, Zhang, Liu, Xu, Dong, and Tang]{cogvideox_yang2024cogvideox}
Yang, Z., Teng, J., Zheng, W., Ding, M., Huang, S., Xu, J., Yang, Y., Zhang, X., Gu, X., Feng, G., Yin, D., Hong, W., Wang, W., Cheng, Y., Zhang, Y., Liu, T., Xu, B., Dong, Y., and Tang, J.
\newblock Cogvideox: Text-to-video diffusion models with an expert transformer.
\newblock 2024.

\bibitem[Yoon et~al.(2024)Yoon, Yu, and Bansal]{raccon_yoon2024raccoonremoveaddchange}
Yoon, J., Yu, S., and Bansal, M.
\newblock Raccoon: Remove, add, and change video content with auto-generated narratives, 2024.
\newblock URL \url{https://arxiv.org/abs/2405.18406}.

\bibitem[Zhang et~al.(2024{\natexlab{a}})Zhang, Mo, Chen, Sun, and Su]{magic_brush_zhang2024magicbrush}
Zhang, K., Mo, L., Chen, W., Sun, H., and Su, Y.
\newblock Magicbrush: A manually annotated dataset for instruction-guided image editing.
\newblock \emph{Advances in Neural Information Processing Systems}, 36, 2024{\natexlab{a}}.

\bibitem[Zhang et~al.(2023{\natexlab{a}})Zhang, Rao, and Agrawala]{controlnet_zhang2023adding}
Zhang, L., Rao, A., and Agrawala, M.
\newblock Adding conditional control to text-to-image diffusion models.
\newblock In \emph{Proceedings of the IEEE/CVF International Conference on Computer Vision}, pp.\  3836--3847, 2023{\natexlab{a}}.

\bibitem[Zhang et~al.(2024{\natexlab{b}})Zhang, Yang, Feng, Qin, Chen, Yu, Chen, Wang, Savarese, Ermon, et~al.]{hive_zhang2024hive}
Zhang, S., Yang, X., Feng, Y., Qin, C., Chen, C.-C., Yu, N., Chen, Z., Wang, H., Savarese, S., Ermon, S., et~al.
\newblock Hive: Harnessing human feedback for instructional visual editing.
\newblock In \emph{Proceedings of the IEEE/CVF Conference on Computer Vision and Pattern Recognition}, pp.\  9026--9036, 2024{\natexlab{b}}.

\bibitem[Zhang et~al.(2023{\natexlab{b}})Zhang, Wu, Wang, Luo, Zhang, Zhao, Vajda, Metaxas, and Yu]{avid_zhang2023avid}
Zhang, Z., Wu, B., Wang, X., Luo, Y., Zhang, L., Zhao, Y., Vajda, P., Metaxas, D., and Yu, L.
\newblock Avid: Any-length video inpainting with diffusion model.
\newblock \emph{arXiv preprint arXiv:2312.03816}, 2023{\natexlab{b}}.

\bibitem[Zhao et~al.(2024{\natexlab{a}})Zhao, Ma, Chen, Si, Wu, An, Yu, Zhang, Li, and Chang]{ultraedit_zhao2024ultraedit}
Zhao, H., Ma, X., Chen, L., Si, S., Wu, R., An, K., Yu, P., Zhang, M., Li, Q., and Chang, B.
\newblock Ultraedit: Instruction-based fine-grained image editing at scale.
\newblock \emph{arXiv preprint arXiv:2407.05282}, 2024{\natexlab{a}}.

\bibitem[Zhao et~al.(2024{\natexlab{b}})Zhao, Chen, Chen, Bao, Hao, Yuan, and Wong]{unicontrolnet_zhao2024uni}
Zhao, S., Chen, D., Chen, Y.-C., Bao, J., Hao, S., Yuan, L., and Wong, K.-Y.~K.
\newblock Uni-controlnet: All-in-one control to text-to-image diffusion models.
\newblock \emph{Advances in Neural Information Processing Systems}, 36, 2024{\natexlab{b}}.

\bibitem[Zhou et~al.(2023)Zhou, Li, Chan, and Loy]{zhou2023propainter}
Zhou, S., Li, C., Chan, K.~C., and Loy, C.~C.
\newblock Propainter: {I}mproving propagation and transformer for video inpainting.
\newblock In \emph{Proceedings of the IEEE/CVF International Conference on Computer Vision}, 2023.

\bibitem[Zi et~al.(2024)Zi, Zhao, Qi, Wang, Shi, Chen, Liang, Wong, and Zhang]{cococo_zi2024cococo}
Zi, B., Zhao, S., Qi, X., Wang, J., Shi, Y., Chen, Q., Liang, B., Wong, K.-F., and Zhang, L.
\newblock Cococo: Improving text-guided video inpainting for better consistency, controllability and compatibility.
\newblock \emph{arXiv preprint arXiv:2403.12035}, 2024.

\end{thebibliography}
\bibliographystyle{icml2025}

\newpage
\appendix
\onecolumn

\section{Qualitative Comparison of Video Editing Models}
\label{sec_visual_comparison}

\begin{figure*}[!htp]

    \begin{minipage}{0.02\textwidth}
        \rotatebox{90}{\footnotesize \centering \ \ \ \ \ \ Ours \ \ \ \ \ \ AnyV2V \ \ Tokenflow \ \ \ \ InsV2V \ \ \ \ \ \ Flatten \ \ \ \ \ Original }
    \end{minipage}
    \begin{minipage}{0.47\textwidth}

    \begin{minipage}{0.98\textwidth}
        \includegraphics[width=0.32\textwidth]{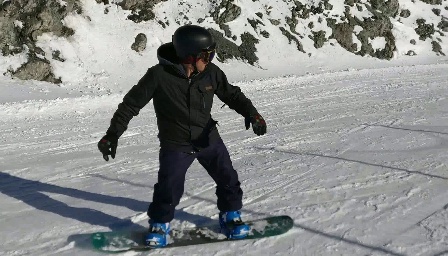}
        \includegraphics[width=0.32\textwidth]{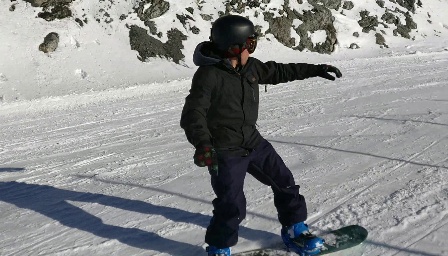}
        \includegraphics[width=0.32\textwidth]{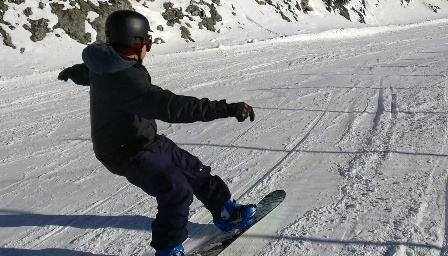}
    \end{minipage}

    \begin{minipage}{0.98\textwidth}
        \includegraphics[width=0.32\textwidth]{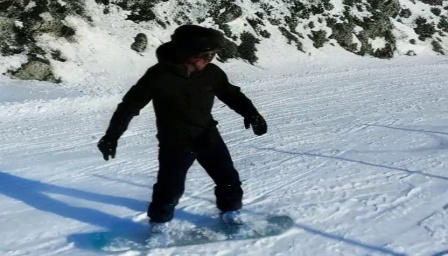}
        \includegraphics[width=0.32\textwidth]{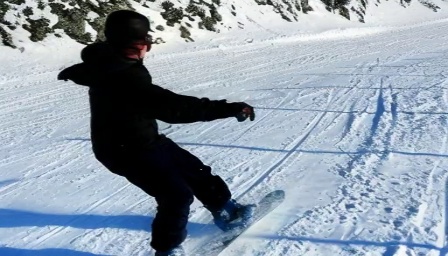}
        \includegraphics[width=0.32\textwidth]{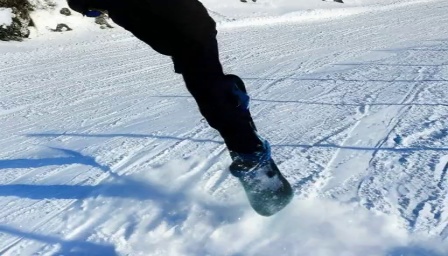}
    \end{minipage}
    
    \begin{minipage}{0.98\textwidth}
        \includegraphics[width=0.32\textwidth]{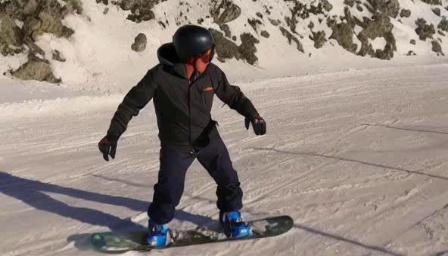}
        \includegraphics[width=0.32\textwidth]{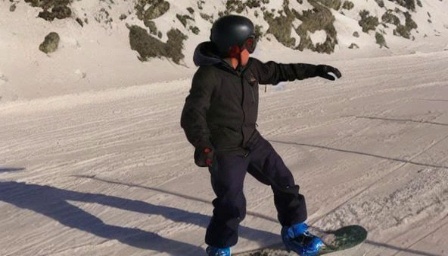}
        \includegraphics[width=0.32\textwidth]{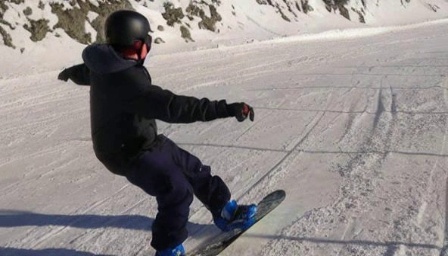}
    \end{minipage}

    \begin{minipage}{0.98\textwidth}
        \includegraphics[width=0.32\textwidth]{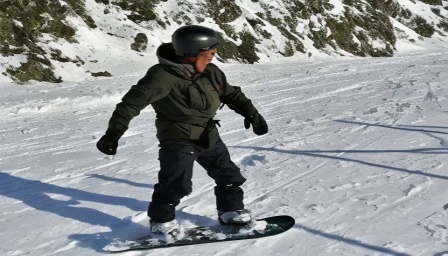}
        \includegraphics[width=0.32\textwidth]{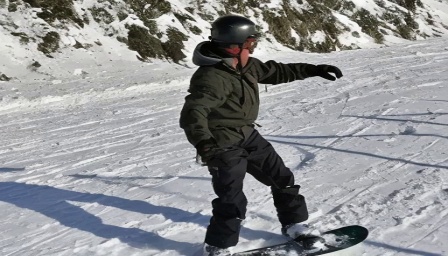}
        \includegraphics[width=0.32\textwidth]{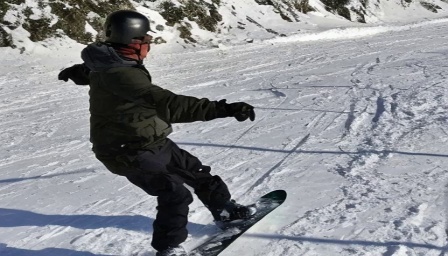}
    \end{minipage}
    
    \begin{minipage}{0.98\textwidth}
        \includegraphics[width=0.32\textwidth]{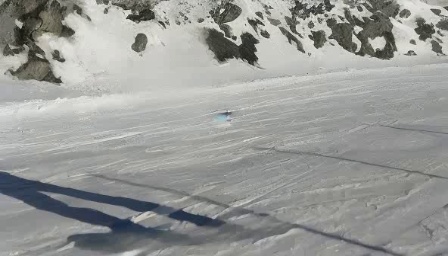}
        \includegraphics[width=0.32\textwidth]{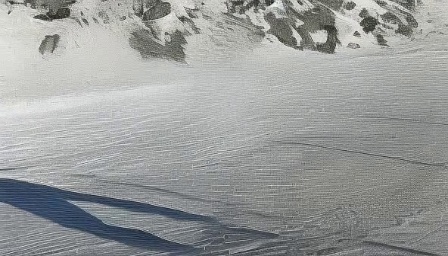}
        \includegraphics[width=0.32\textwidth]{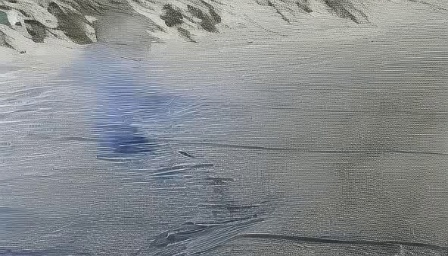}
    \end{minipage}

    \begin{minipage}{0.98\textwidth}
        \includegraphics[width=0.32\textwidth]{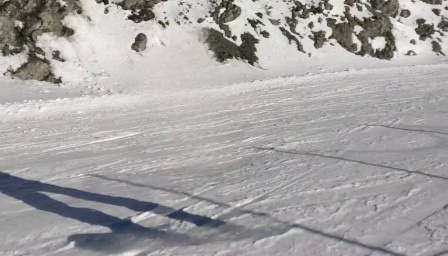}
        \includegraphics[width=0.32\textwidth]{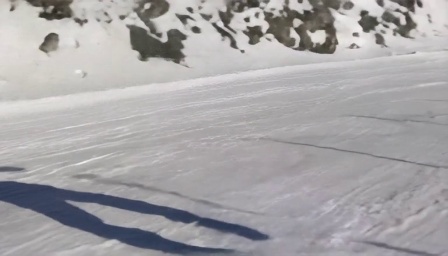}
        \includegraphics[width=0.32\textwidth]{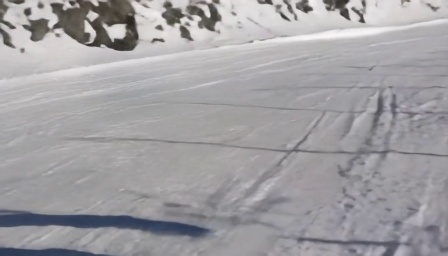}
        \vspace{-0.5em}
        \subcaption*{\small Remove the man.}
    \end{minipage}
    \end{minipage}
    \begin{minipage}{0.47\textwidth}
    \begin{minipage}{0.98\textwidth}
        \includegraphics[width=0.32\textwidth]{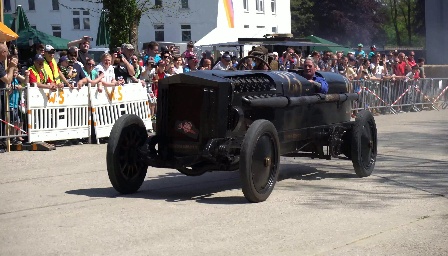}
        \includegraphics[width=0.32\textwidth]{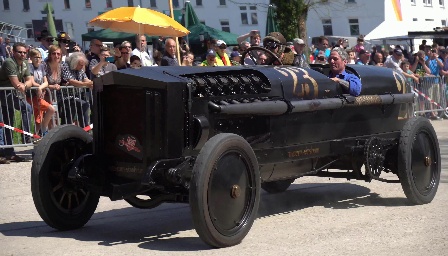}
        \includegraphics[width=0.32\textwidth]{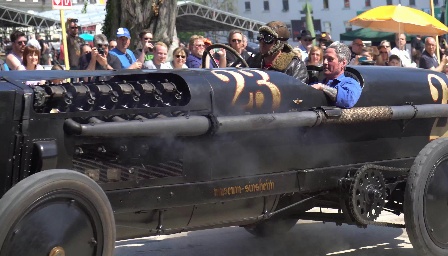}
    \end{minipage}
    
    \begin{minipage}{0.98\textwidth}
        \includegraphics[width=0.32\textwidth]{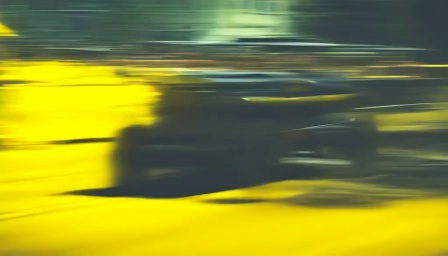}
        \includegraphics[width=0.32\textwidth]{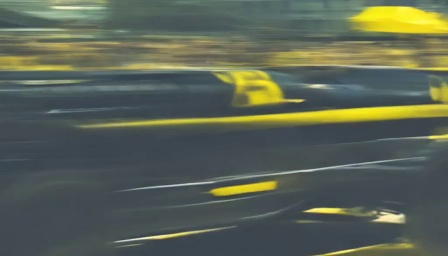}
        \includegraphics[width=0.32\textwidth]{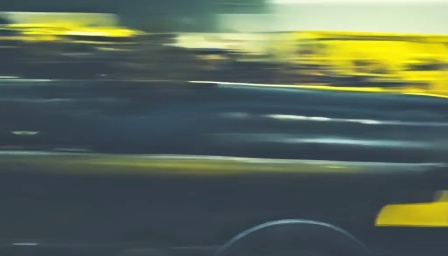}
    \end{minipage}
    
    \begin{minipage}{0.98\textwidth}
        \includegraphics[width=0.32\textwidth]{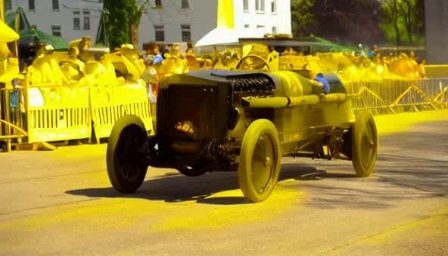}
        \includegraphics[width=0.32\textwidth]{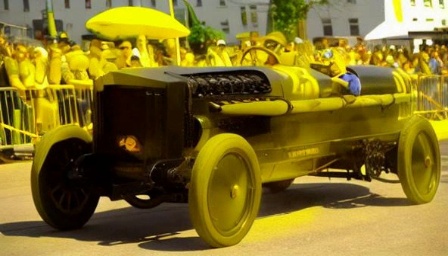}
        \includegraphics[width=0.32\textwidth]{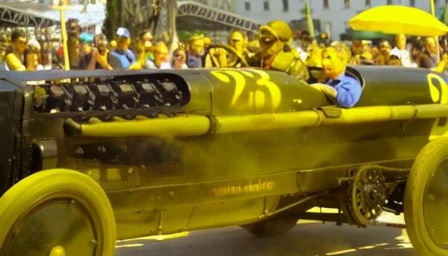}
    \end{minipage}

    \begin{minipage}{0.98\textwidth}
        \includegraphics[width=0.32\textwidth]{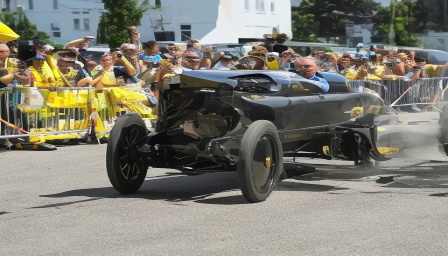}
        \includegraphics[width=0.32\textwidth]{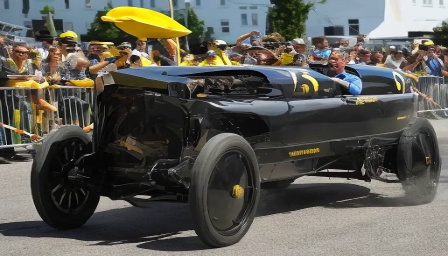}
        \includegraphics[width=0.32\textwidth]{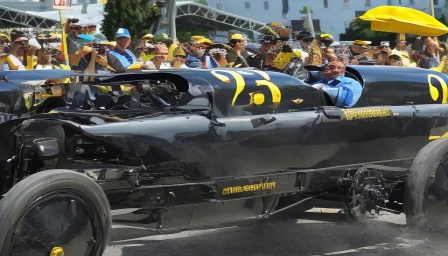}
    \end{minipage}
    
    \begin{minipage}{0.98\textwidth}
        \includegraphics[width=0.32\textwidth]{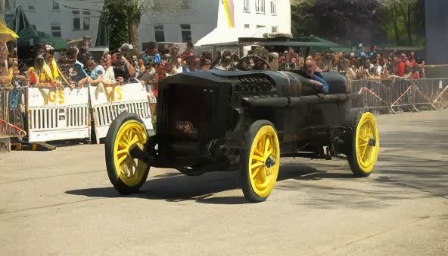}
        \includegraphics[width=0.32\textwidth]{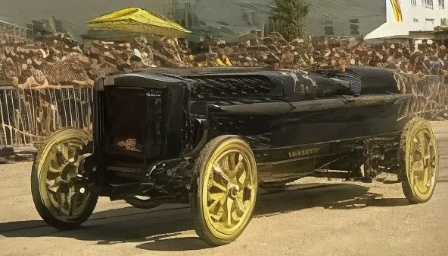}
        \includegraphics[width=0.32\textwidth]{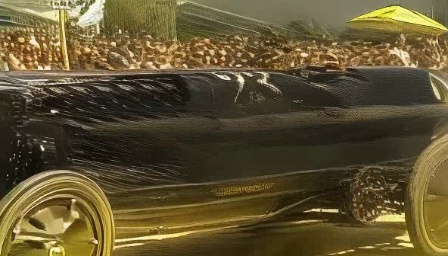}
    \end{minipage}

    \begin{minipage}{0.98\textwidth}
        \includegraphics[width=0.32\textwidth]{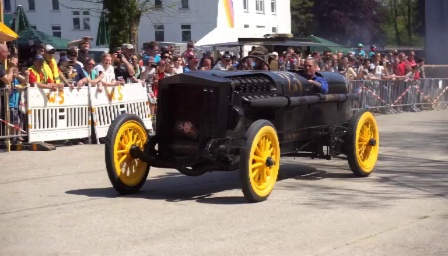}
        \includegraphics[width=0.32\textwidth]{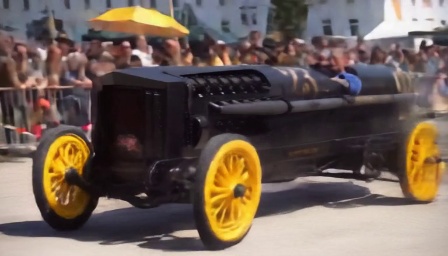}
        \includegraphics[width=0.32\textwidth]{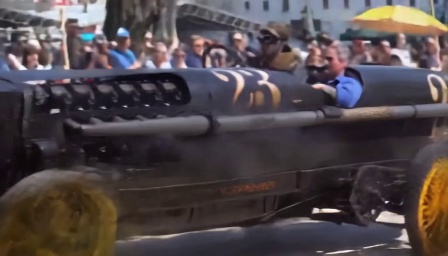}
        \vspace{-0.5em}
        \subcaption*{\small Make the wheels yellow.}
    \end{minipage}
    \end{minipage}

    \begin{minipage}{0.02\textwidth}
        \rotatebox{90}{\footnotesize \centering \ \ \ \ \ \ Ours \ \ \ \ \ \ AnyV2V \ \ Tokenflow \ \ \ \ InsV2V \ \ \ \ \ \ Flatten \ \ \ \ \ Original }
    \end{minipage}
    \begin{minipage}{0.47\textwidth}
    \begin{minipage}{0.98\textwidth}
        \includegraphics[width=0.32\textwidth]{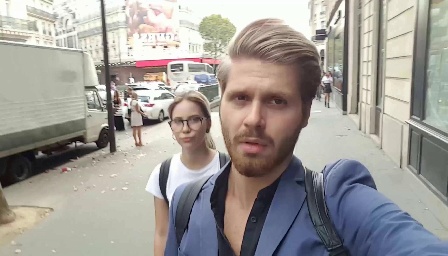}
        \includegraphics[width=0.32\textwidth]{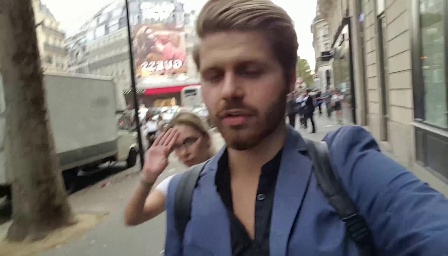}
        \includegraphics[width=0.32\textwidth]{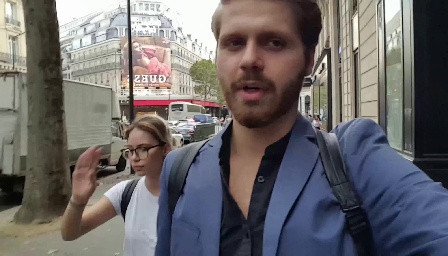}
    \end{minipage}
    
    \begin{minipage}{0.98\textwidth}
        \includegraphics[width=0.32\textwidth]{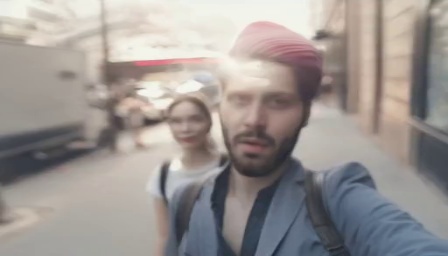}
        \includegraphics[width=0.32\textwidth]{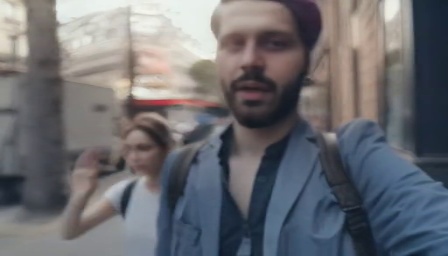}
        \includegraphics[width=0.32\textwidth]{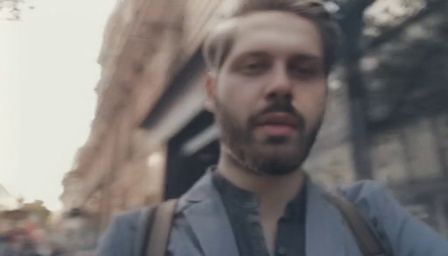}
    \end{minipage}
    
    \begin{minipage}{0.98\textwidth}
        \includegraphics[width=0.32\textwidth]{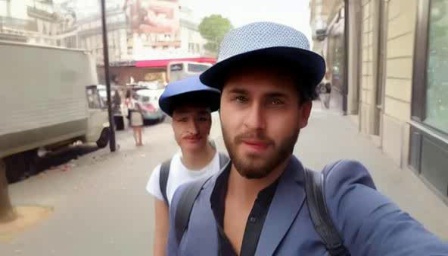}
        \includegraphics[width=0.32\textwidth]{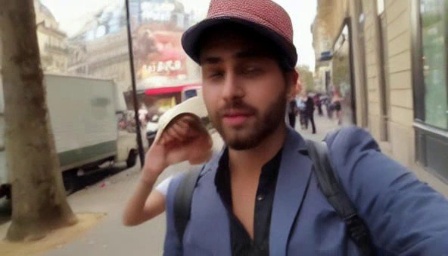}
        \includegraphics[width=0.32\textwidth]{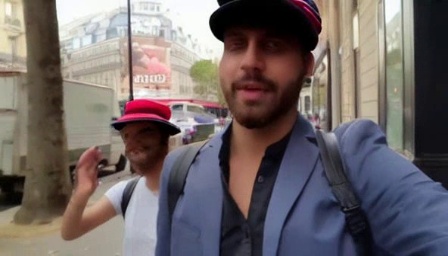}
    \end{minipage}

    \begin{minipage}{0.98\textwidth}
        \includegraphics[width=0.32\textwidth]{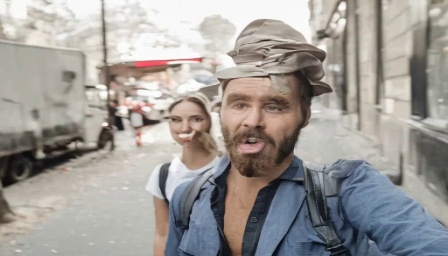}
        \includegraphics[width=0.32\textwidth]{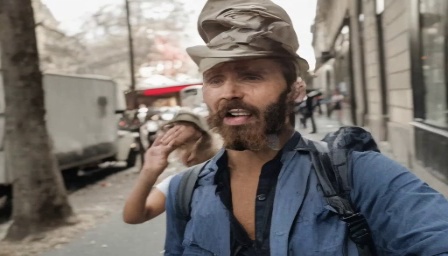}
        \includegraphics[width=0.32\textwidth]{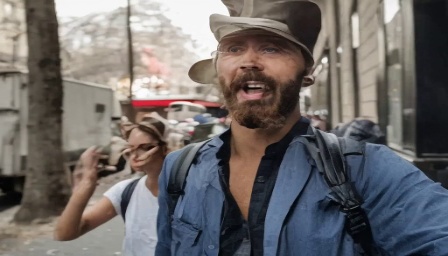}
    \end{minipage}
    
    \begin{minipage}{0.98\textwidth}
        \includegraphics[width=0.32\textwidth]{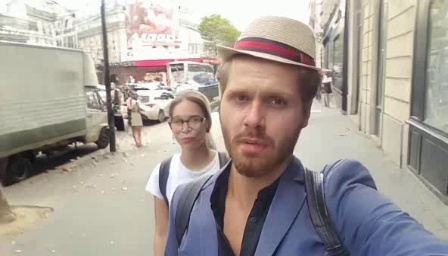}
        \includegraphics[width=0.32\textwidth]{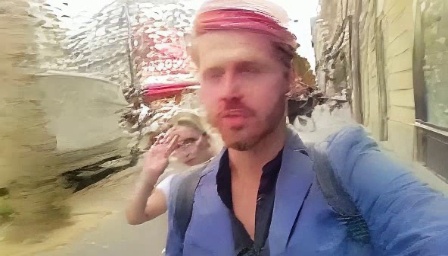}
        \includegraphics[width=0.32\textwidth]{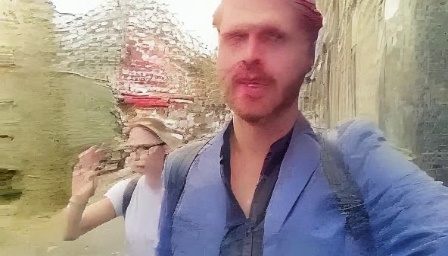}
    \end{minipage}

    \begin{minipage}{0.98\textwidth}
        \includegraphics[width=0.32\textwidth]{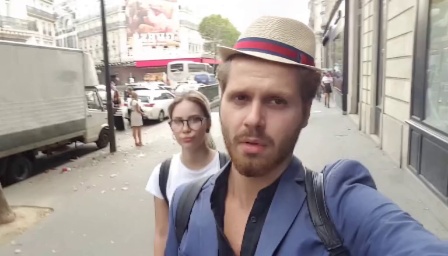}
        \includegraphics[width=0.32\textwidth]{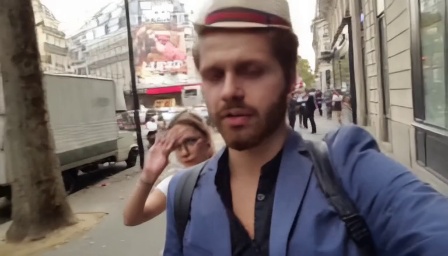}
        \includegraphics[width=0.32\textwidth]{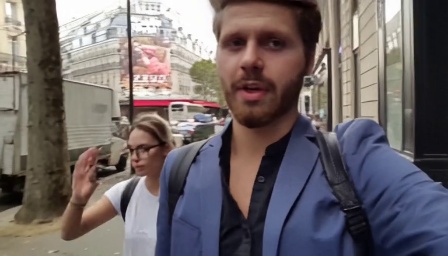}
        \vspace{-0.5em}
        \subcaption*{\small Add hat on his head.}
    \end{minipage}
    \end{minipage}
    \begin{minipage}{0.47\textwidth}
    \begin{minipage}{0.98\textwidth}
        \includegraphics[width=0.32\textwidth]{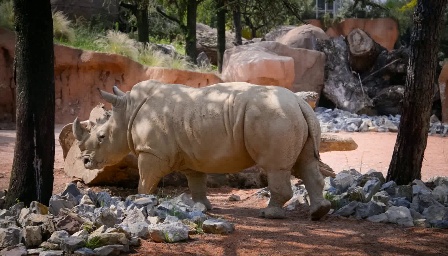}
        \includegraphics[width=0.32\textwidth]{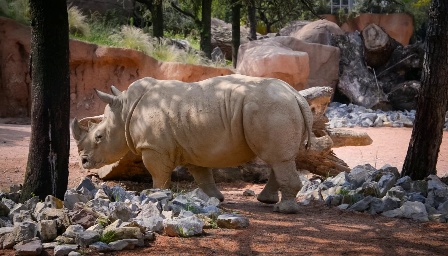}
        \includegraphics[width=0.32\textwidth]{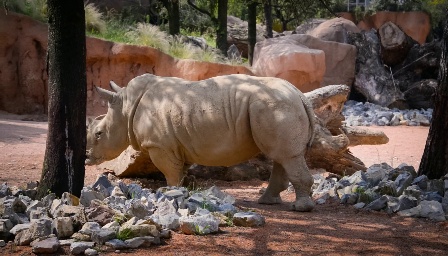}
    \end{minipage}

    \begin{minipage}{0.98\textwidth}
        \includegraphics[width=0.32\textwidth]{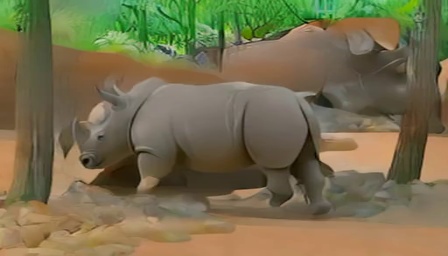}
        \includegraphics[width=0.32\textwidth]{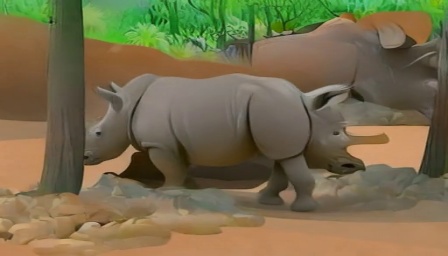}
        \includegraphics[width=0.32\textwidth]{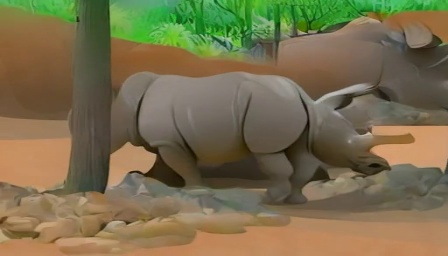}
    \end{minipage}
    
    \begin{minipage}{0.98\textwidth}
        \includegraphics[width=0.32\textwidth]{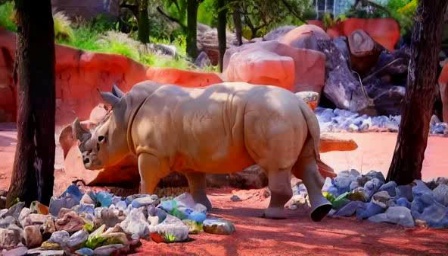}
        \includegraphics[width=0.32\textwidth]{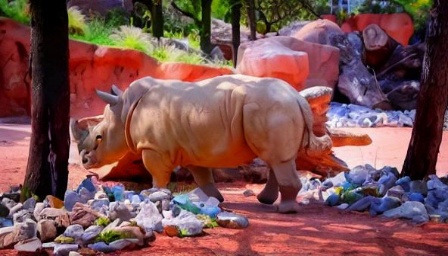}
        \includegraphics[width=0.32\textwidth]{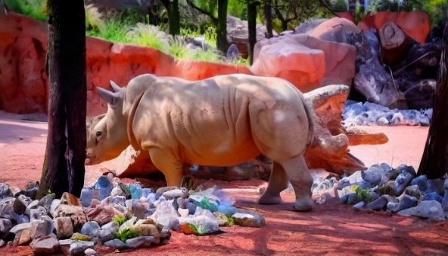}
    \end{minipage}

    \begin{minipage}{0.98\textwidth}
        \includegraphics[width=0.32\textwidth]{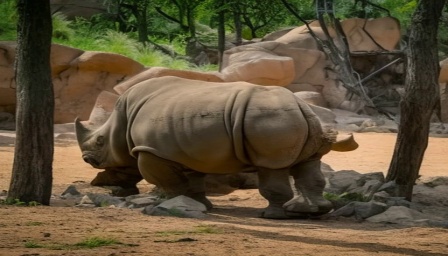}
        \includegraphics[width=0.32\textwidth]{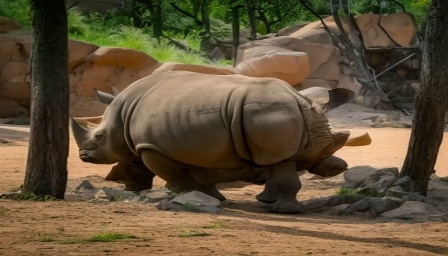}
        \includegraphics[width=0.32\textwidth]{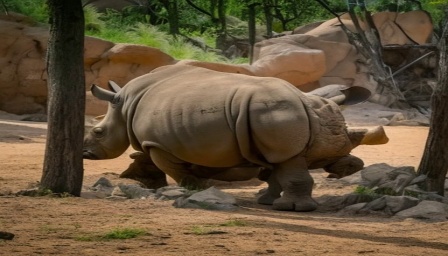}
    \end{minipage}
    
    \begin{minipage}{0.98\textwidth}
        \includegraphics[width=0.32\textwidth]{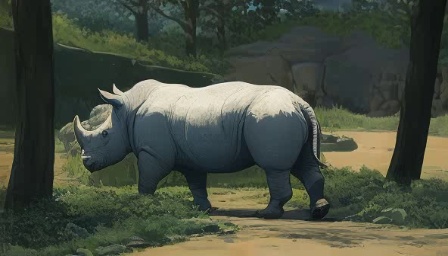}
        \includegraphics[width=0.32\textwidth]{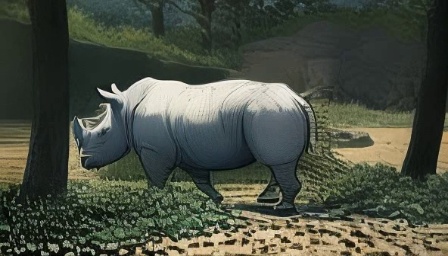}
        \includegraphics[width=0.32\textwidth]{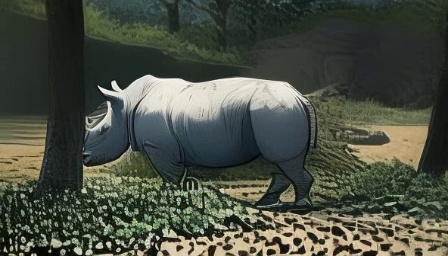}
    \end{minipage}

    \begin{minipage}{0.98\textwidth}
        \includegraphics[width=0.32\textwidth]{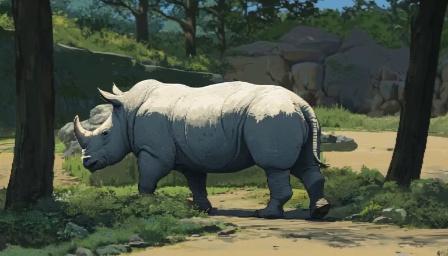}
        \includegraphics[width=0.32\textwidth]{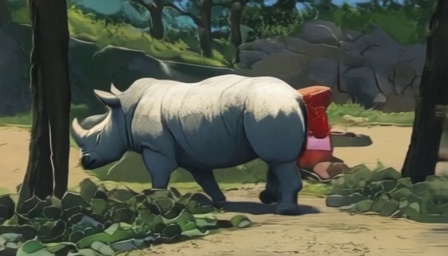}
        \includegraphics[width=0.32\textwidth]{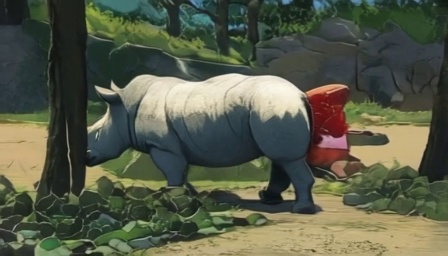}
        \vspace{-0.5em}
        \subcaption*{\small Make it anime style.}
    \end{minipage}
    
    \end{minipage}
    \caption{Editing results compared between different editing methods.}
    \label{visual_comparison}
\end{figure*}

\section{The Design of Video Editing Experts}
\label{sec_design_of_video_editing_experts_appdendix}
\begin{figure*}[!ht]
    \centering
    \includegraphics[width=0.95\linewidth]{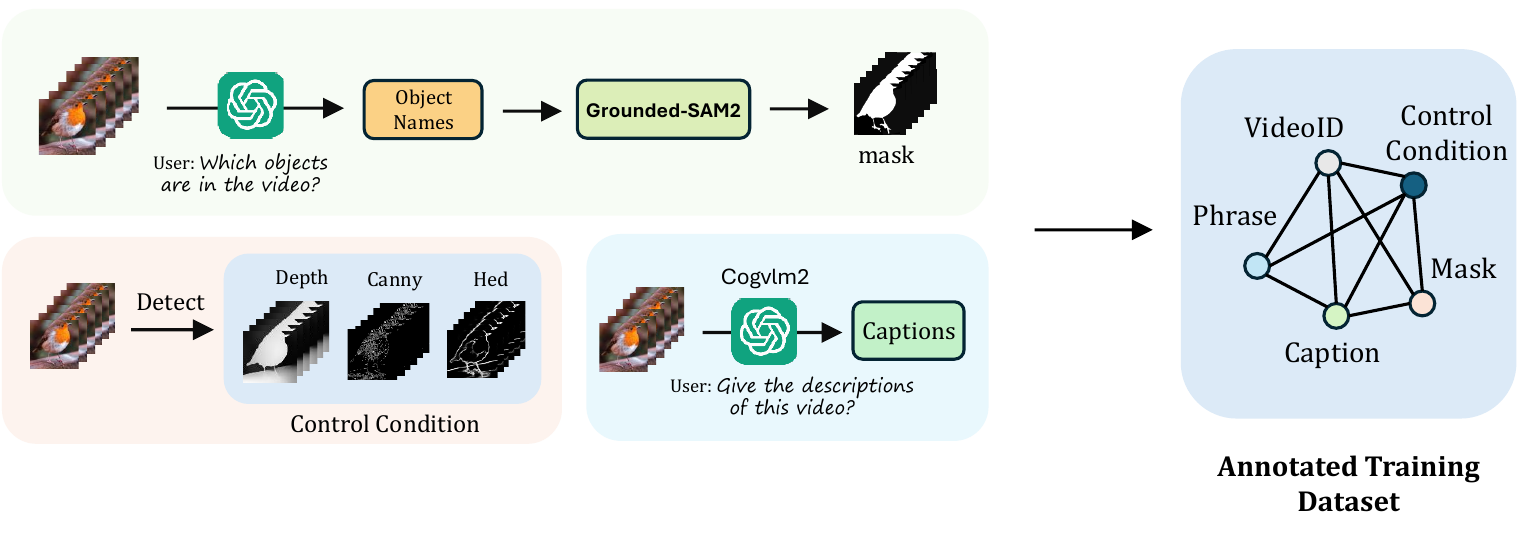}
    \caption{The construction pipeline of annotated training dataset for experts training.}
    \label{fig:data_annotation}
\end{figure*}

\subsection{The Construction of Expert Training Dataset}
As shown in Figure ~\ref{fig:data_annotation}, we built a well-annotated dataset based on WebVid-10M~\citep{webvid_bain2021frozen} to train our expert models. We use the CogVLM2~\citep{Cogvlm2_hong2024cogvlm2} to recognize objects in the video. The detected object names are separated by commas and used as input prompts for Grounded-SAM2~\citep{groundingdino_liu2023grounding, sam2_ravi2024sam2}. This process generates phrase names and corresponding object mask sequences within the video. These annotations are used to train the inpainter, remover, and local stylizer models. Moreover, We use canny, hed and depth detector to get control conditions to train our global stylizer and local stylizer. For inpainter training, we also leverage detailed video captions. Specifically, we use CogVLM2 to generate detailed, descriptive captions for each video. All data annotation processes are conducted on Nvidia 4090 GPUs. The prompts used for object recognition and video captioning are provided as followings:

\begin{tcolorbox}[colback=white,colframe=black!75!white,title=The Prompts for Video Captioning and Object Recognition]
Video Captioning: \textit{Please give the descriptions of this video. The answer should be more than 20 words but less than 60 words. The answer is:}\\
  
Object Recognition: \textit{What objects are in this video? Please list them by using comma to separate different words. Give me answers briefly. Do not give detailed descriptions, years or numbers. Give object names. The answer is:}
\end{tcolorbox}

\begin{figure*}[!ht]
    \centering
    \includegraphics[width=0.925\linewidth]{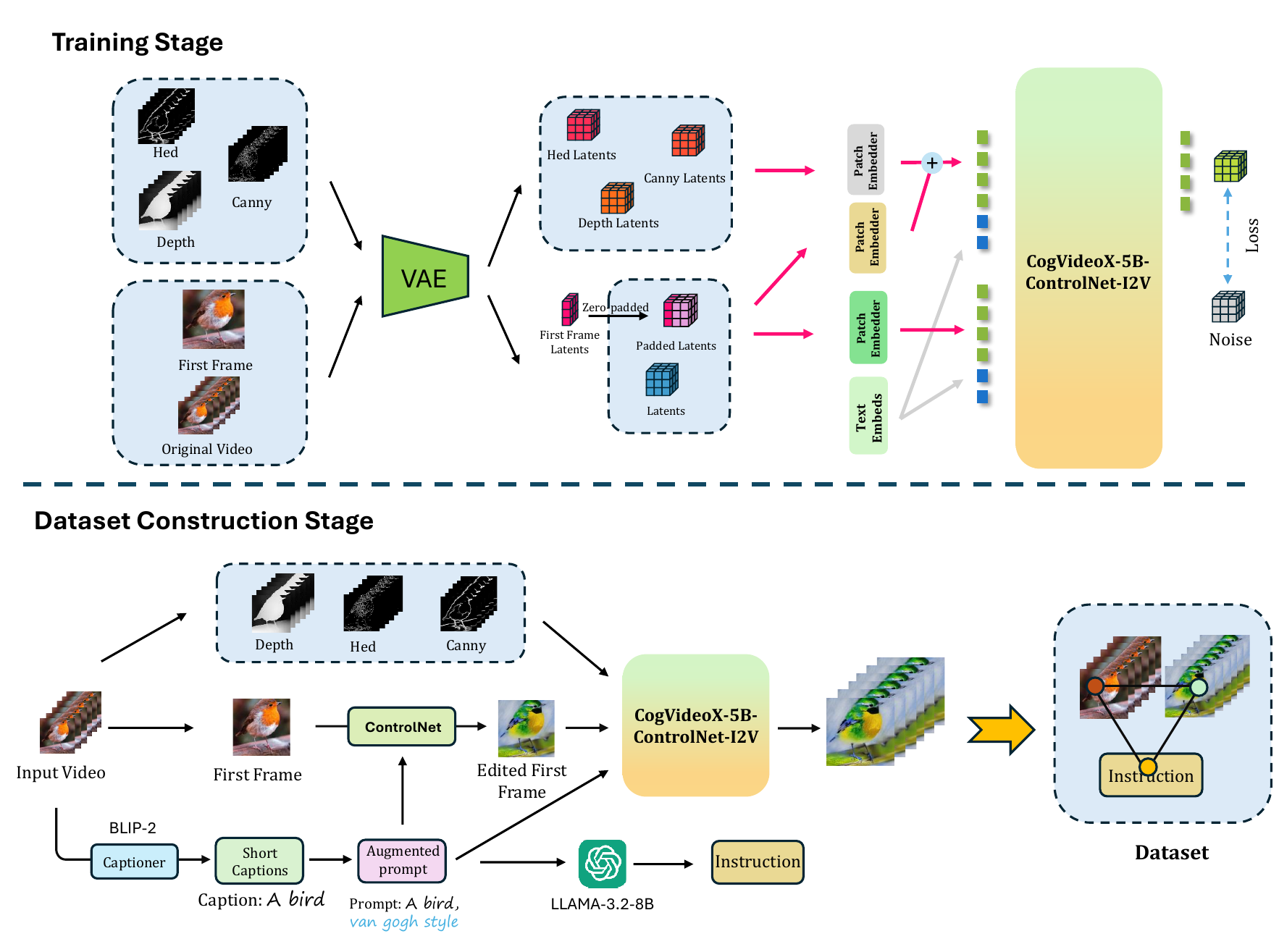}
    \caption{Top: The training pipeline of our global stylizer. Bottom: The data construction pipeline for Señorita-2M using our global stylizer.}
    \label{fig:global-style-transfer}
\end{figure*}

\subsection{The Construction of Global Stylizer}
\label{sec_app_global_stylizer}

\begin{table*}[!ht]
\centering
\caption{Quantitative Comparison on Global Stylization. The best results are \textbf{boldfaced}.}
\begin{tabular}{c|ccc} 
\toprule

\textbf{Methods} & Ewarp$(10^{-3})$($\downarrow$)& CLIPScore ($\uparrow$)& Temp-Cons ($\uparrow$)\\
\hline
Tokenflow & 19.99 & 0.3125 & 0.9752 \\ 
Flatten & 11.18 & 0.3127 & 0.9759 \\ 
InsV2V & 9.61 & 0.2864 & 0.9736 \\ 
AnyV2V & 34.94 & 0.2928 & 0.9687 \\ 
\hline
Our Expert & \textbf{9.02} & \textbf{0.3145} & \textbf{0.9781}  \\ 

\bottomrule
\end{tabular} 
\label{tab:global_stylization_comparison} 
\end{table*}

We find that two powerful video generation models, \ie CogVideoX~\citep{cogvideox_yang2024cogvideox} and HunyuanVideo~\citep{hunyuanvideosystematicframeworklargekong2025}, lack sufficient ability to generate videos that accurately follow style information. This limitation prevents us from applying techniques such as ControlNet to repaint a video effectively.
To address this, we shift our focus to image-based ControlNet to leverage the strong stylization capabilities of these models, enhancing the stylization of video generation. Specifically, we first apply an image ControlNet to process the first frame, then use a video ControlNet to propagate the style across the remaining frames. Since video generation models inherently maintain temporal consistency between frames, the style applied to the first frame can be effectively transferred to the rest of the video.

We utilize the architecture of ControlNet for DiT and integrate it with CogVideoX-5B-I2V to process subsequent frames, maintaining the video's structure consistently propagate the style from the first frame. Specifically, we inject first frame features from the control branch (ControlNet) into the main branch (base model) through zero convolution.
The control branch consists of N layers, while the main branch has M layers, with M being a multiple of N. We ensure that once the first N layers of the main branch have been added with hidden states, the K-th layer of the main branch receives the K\%N-th hidden state from the control branch. This process is repeated until all DiT blocks in the main branch have received the control hidden states.

\begin{figure*}[!ht]
  \centering
    \begin{subfigure}{0.49\textwidth}
      \animategraphics[width=1\textwidth]{8}{data/style_transfer/van_gogh/transfer-2/9133278-hd_2048_1080_25fps_style_id_14_0_0_guidance_scale_6_output5178_org/image_}{0}{15}
    \end{subfigure}
    \begin{subfigure}{0.49\textwidth}
      \animategraphics[width=1\textwidth]{8}{data/style_transfer/van_gogh/transfer-2/9133278-hd_2048_1080_25fps_style_id_14_0_0_guidance_scale_6_output5178/image_}{0}{15}
    \end{subfigure}

    \begin{subfigure}{0.24\textwidth}
      \animategraphics[width=1\textwidth]{8}{data/style_transfer/ghibli/transfer-1/3151434-hd_2048_1080_24fps_style_id_23_0_0_guidance_scale_6_output5560_org/image_}{0}{15}
    \end{subfigure}
    \begin{subfigure}{0.24\textwidth}
      \animategraphics[width=1\textwidth]{8}{data/style_transfer/ghibli/transfer-1/3151434-hd_2048_1080_24fps_style_id_23_0_0_guidance_scale_6_output5560/image_}{0}{15}
    \end{subfigure}
    \begin{subfigure}{0.24\textwidth}
      \animategraphics[width=1\textwidth]{8}{data/style_transfer/ghibli/transfer-2/9459890-hd_1280_720_30fps_style_id_23_0_0_guidance_scale_6_output5813_org/image_}{0}{15}
    \end{subfigure}
    \begin{subfigure}{0.24\textwidth}
      \animategraphics[width=1\textwidth]{8}{data/style_transfer/ghibli/transfer-2/9459890-hd_1280_720_30fps_style_id_23_0_0_guidance_scale_6_output5813/image_}{0}{15}
    \end{subfigure}

    \begin{subfigure}{0.24\textwidth}
      \animategraphics[width=1\textwidth]{8}{data/style_transfer/van_gogh/transfer-1/3978648-hd_1280_720_24fps_style_id_14_0_0_guidance_scale_6_output4830_org/image_}{0}{15}
    \end{subfigure}
    \begin{subfigure}{0.24\textwidth}
      \animategraphics[width=1\textwidth]{8}{data/style_transfer/van_gogh/transfer-1/3978648-hd_1280_720_24fps_style_id_14_0_0_guidance_scale_6_output4830/image_}{0}{15}
    \end{subfigure}
    \begin{subfigure}{0.24\textwidth}
      \animategraphics[width=1\textwidth]{8}{data/style_transfer/van_gogh/transfer-1/4170289-hd_1280_720_24fps_style_id_14_0_0_guidance_scale_6_output4834_org/image_}{0}{15}
    \end{subfigure}
    \begin{subfigure}{0.24\textwidth}
      \animategraphics[width=1\textwidth]{8}{data/style_transfer/van_gogh/transfer-1/4170289-hd_1280_720_24fps_style_id_14_0_0_guidance_scale_6_output4834/image_}{0}{15}
    \end{subfigure}

    \begin{subfigure}{0.24\textwidth}
      \animategraphics[width=1\textwidth]{8}{data/style_transfer/waterpainting/transfer-1/9733889-hd_2048_1080_30fps_style_id_266_0_0_guidance_scale_6_output7040_org/image_}{0}{15}
    \end{subfigure}
    \begin{subfigure}{0.24\textwidth}
      \animategraphics[width=1\textwidth]{8}{data/style_transfer/waterpainting/transfer-1/9733889-hd_2048_1080_30fps_style_id_266_0_0_guidance_scale_6_output7040/image_}{0}{15}
    \end{subfigure}
    \begin{subfigure}{0.24\textwidth}
      \animategraphics[width=1\textwidth]{8}{data/style_transfer/waterpainting/transfer-2/7248428-hd_1280_720_24fps_style_id_266_0_0_guidance_scale_6_output7583_org/image_}{0}{15}
    \end{subfigure}
    \begin{subfigure}{0.24\textwidth}
      \animategraphics[width=1\textwidth]{8}{data/style_transfer/waterpainting/transfer-2/7248428-hd_1280_720_24fps_style_id_266_0_0_guidance_scale_6_output7583/image_}{0}{15}
    \end{subfigure}

    \begin{subfigure}{0.24\textwidth}
      \animategraphics[width=1\textwidth]{8}{data/style_transfer/cyberpunk/transfer-1/852436-hd_1280_720_30fps_style_id_262_0_0_guidance_scale_6_output5293_org/image_}{0}{15}
    \end{subfigure}
    \begin{subfigure}{0.24\textwidth}
      \animategraphics[width=1\textwidth]{8}{data/style_transfer/cyberpunk/transfer-1/852436-hd_1280_720_30fps_style_id_262_0_0_guidance_scale_6_output5293/image_}{0}{15}
    \end{subfigure}
    \begin{subfigure}{0.24\textwidth}
      \animategraphics[width=1\textwidth]{8}{data/style_transfer/cyberpunk/transfer-1/5384656-hd_1280_720_25fps_style_id_262_0_0_guidance_scale_6_output5552_org/image_}{0}{15}
    \end{subfigure}
    \begin{subfigure}{0.24\textwidth}
      \animategraphics[width=1\textwidth]{8}{data/style_transfer/cyberpunk/transfer-1/5384656-hd_1280_720_25fps_style_id_262_0_0_guidance_scale_6_output5552/image_}{0}{15}
    \end{subfigure}

    \caption{The visual results of our global stylization. The video on the left depicts the original video, while the video on the right displays the edited videos. \emph{Best viewed with Acrobat Reader. Click the images to play the animation clips.}}
    \label{figure:global_stylization}
\end{figure*}

For the input to the DiT block in the transformer, we use two types of patch embedders. In the main branch, the patch embedder processes the input video condition. The control branch uses two patch embedders: the main patch embedder and the control patch embedder, to receive and process the control condition. The output embeddings of both embedders are combined and fed into the model. As shown in Figure~\ref{fig:global-style-transfer}, we pad the first frame along the frame dimension and concatenate it with the original latent representation along the channel dimension (32 channels) as input to the main branch. In the control branch, we concatenate the HED, Canny, and depth latents representations along the channel dimension (48 channels).

For training, the parameters of the main branch are initialized using CogVideoX-5B-I2V, and the control branch is copied from the main branch, except for the control patch embedder, which is zero-initialized. We select the first 6 DiT blocks from the main branch to serve as the control branch. We trained our global stylizer on our expert dataset for 1 epoch with a batch size of 8, learning rate of 1e-5, and weight decay of 1e-4. We freeze some training layers to reduce the training cost and keep generalization. Specifically, the norm and FFN layers in the backbone were frozen, while the first DiT block in the control branch was trained. Only the first DiT block, patch embedder, and attention layer in the control branch were trained. We train our global stylizer in two phases. In Phase 1, we train the global stylizer on videos with a resolution of $256\times448\times33$. Additionally, we incorporate a 10\% null prompt during training to enable classifier-free guidance. In Phase 2, we finetuned the model from Phase 1, increasing the spatial resolution to $448\times896$.

During inference, we append the required style prompt to the end of the video description, creating a new combined prompt. The first frame is generated by ControlNet-SD1.5, which is then fed into the model along with the prompt and control condition. We use the classifier-free guidance of 4. The model processes a video within 2 minutes on an Nvidia RTX 4090, at a resolution of 336$\times$592, producing 33 frames.

Table~\ref{tab:global_stylization_comparison} shows that our expert model outperforms all baselines, achieving the lowest Ewarp (9.02), highest CLIPScore (0.3145), and best Temporal Consistency (0.9781). While InsV2V performs well in Ewarp, it lags in text alignment. AnyV2V exhibits the highest distortion (Ewarp 34.94), indicating poor stylization. These results highlight our expert model's superior balance of visual quality, text alignment, and temporal smoothness.

\begin{figure*}[!ht]
    \centering
    \includegraphics[width=0.95\linewidth]{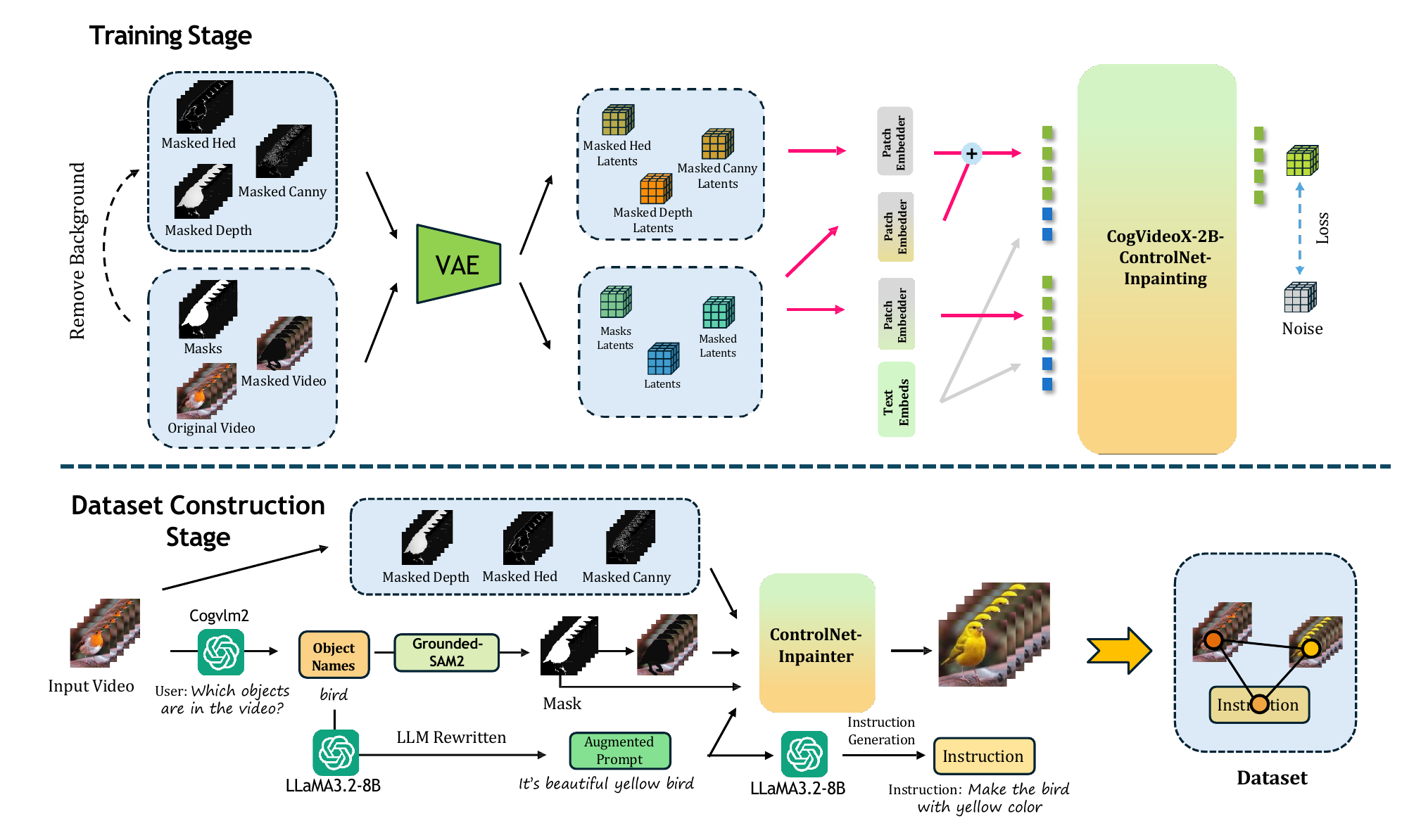}
    \caption{Top: The training pipeline of our local stylizer. Bottom: The data construction pipeline for Señorita-2M using our local stylizer.}
    \label{fig:local_stylization2}
\end{figure*}

\subsection{The Construction of Local Stylizer}
\label{sec_app_local_stylizer}

\begin{table*}[!htp]
\centering
\caption{Quantitative Comparison on Local Stylization. The best results are \textbf{boldfaced}.}
\begin{tabular}{c|ccccccc} 
\toprule 
\textbf{Methods} & Ewarp$(10^{-3})$ ($\downarrow$)& CLIPScore ($\uparrow$)& Temp-Cons ($\uparrow$)& PSNR ($\uparrow$)& SSIM ($\uparrow$)& LPIPS ($\downarrow$)& MSE ($\downarrow$)\\
\hline
Tokenflow & 16.60 & 0.2876 & 0.9810 & 18.79 & 0.8555 & 0.1483 & 987.90 \\ 
Flatten & 17.18 & 0.2923 & 0.9751 & 18.64 & 0.8605 & 0.1463 & 1068.95  \\ 
InsV2V & 7.40 & 0.2830 & 0.9783 & 20.81 & 0.9091 & 0.0985 & 829.83 \\ 
AnyV2V & 15.77 & 0.2920 & 0.9759 & 19.60 & 0.8884 & 0.1207 & 835.39 \\ 
\hline
Our Expert & \textbf{6.50} & \textbf{0.2944} & \textbf{0.9828} & \textbf{28.29} & \textbf{0.9843} & \textbf{0.0346} & \textbf{108.25} \\ 

\bottomrule
\end{tabular} 
\label{tab:comparison} 
\end{table*}

Inspired by SparseControl~\citep{guo2023sparsectrl}, CoCoCo~\citep{cococo_zi2024cococo}, and AVID~\citep{avid_zhang2023avid}, we trained a local stylizer by combining both inpainting and ControlNet, enabling appearance modification, stylization, and texture manipulation in specific regions of videos, while keeping the original background unchanged.

\begin{figure*}[!ht]
  \centering
  
    \begin{subfigure}{0.48\textwidth}
      \animategraphics[width=1\textwidth]{8}{data/local_style_transfer/1179810-hd_1280_720_30fps_mask_id_1_0_0_guidance_scale_6.0_output1215_org/image_}{0}{15}
    \end{subfigure}
    \begin{subfigure}{0.48\textwidth}
      \animategraphics[width=1\textwidth]{8}{data/local_style_transfer/1179810-hd_1280_720_30fps_mask_id_1_0_0_guidance_scale_6.0_output1215/image_}{0}{15}
    \end{subfigure}

    \begin{subfigure}{0.24\textwidth}
      \animategraphics[width=1\textwidth]{8}{data/local_style_transfer/4039046-hd_1280_720_30fps_mask_id_1_0_0_guidance_scale_6.0_output1657_org/image_}{0}{15}
    \end{subfigure}
    \begin{subfigure}{0.24\textwidth}
      \animategraphics[width=1\textwidth]{8}{data/local_style_transfer/4039046-hd_1280_720_30fps_mask_id_1_0_0_guidance_scale_6.0_output1657/image_}{0}{15}
    \end{subfigure}
    \begin{subfigure}{0.24\textwidth}
      \animategraphics[width=1\textwidth]{8}{data/local_style_transfer/4523106-hd_1280_720_25fps_mask_id_2_0_0_guidance_scale_6.0_output1367_org/image_}{0}{15}
    \end{subfigure}
    \begin{subfigure}{0.24\textwidth}
      \animategraphics[width=1\textwidth]{8}{data/local_style_transfer/4523106-hd_1280_720_25fps_mask_id_2_0_0_guidance_scale_6.0_output1367/image_}{0}{15}
    \end{subfigure}

    \begin{subfigure}{0.24\textwidth}
      \animategraphics[width=1\textwidth]{8}{data/local_style_transfer/5487156-hd_1280_720_30fps_mask_id_2_0_0_guidance_scale_6.0_output654_org/image_}{0}{15}
    \end{subfigure}
    \begin{subfigure}{0.24\textwidth}
      \animategraphics[width=1\textwidth]{8}{data/local_style_transfer/5487156-hd_1280_720_30fps_mask_id_2_0_0_guidance_scale_6.0_output654/image_}{0}{15}
    \end{subfigure}
    \begin{subfigure}{0.24\textwidth}
      \animategraphics[width=1\textwidth]{8}{data/local_style_transfer/5940882-hd_1280_720_60fps_mask_id_6_0_0_guidance_scale_6.0_output468_org/image_}{0}{15}
    \end{subfigure}
    \begin{subfigure}{0.24\textwidth}
      \animategraphics[width=1\textwidth]{8}{data/local_style_transfer/5940882-hd_1280_720_60fps_mask_id_6_0_0_guidance_scale_6.0_output468/image_}{0}{15}
    \end{subfigure}

    \begin{subfigure}{0.24\textwidth}
      \animategraphics[width=1\textwidth]{8}{data/local_style_transfer/6054878-hd_2048_1080_25fps_mask_id_1_0_0_guidance_scale_6.0_output821_org/image_}{0}{15}
    \end{subfigure}
    \begin{subfigure}{0.24\textwidth}
      \animategraphics[width=1\textwidth]{8}{data/local_style_transfer/6054878-hd_2048_1080_25fps_mask_id_1_0_0_guidance_scale_6.0_output821/image_}{0}{15}
    \end{subfigure}
    \begin{subfigure}{0.24\textwidth}
      \animategraphics[width=1\textwidth]{8}{data/local_style_transfer/6133943-hd_1280_720_30fps_mask_id_1_0_0_guidance_scale_6.0_output1679_org/image_}{0}{15}
    \end{subfigure}
    \begin{subfigure}{0.24\textwidth}
      \animategraphics[width=1\textwidth]{8}{data/local_style_transfer/6133943-hd_1280_720_30fps_mask_id_1_0_0_guidance_scale_6.0_output1679/image_}{0}{15}
    \end{subfigure}
    
    \begin{subfigure}{0.24\textwidth}
      \animategraphics[width=1\textwidth]{8}{data/local_style_transfer/6277842-hd_1280_720_25fps_mask_id_2_0_0_guidance_scale_6.0_output213_org/image_}{0}{15}
    \end{subfigure}
    \begin{subfigure}{0.24\textwidth}
      \animategraphics[width=1\textwidth]{8}{data/local_style_transfer/6277842-hd_1280_720_25fps_mask_id_2_0_0_guidance_scale_6.0_output213/image_}{0}{15}
    \end{subfigure}
    \begin{subfigure}{0.24\textwidth}
      \animategraphics[width=1\textwidth]{8}{data/local_style_transfer/6500331-hd_1280_720_25fps_mask_id_2_0_0_guidance_scale_6.0_output122_org/image_}{0}{15}
    \end{subfigure}
    \begin{subfigure}{0.24\textwidth}
      \animategraphics[width=1\textwidth]{8}{data/local_style_transfer/6500331-hd_1280_720_25fps_mask_id_2_0_0_guidance_scale_6.0_output122/image_}{0}{15}
    \end{subfigure}

\caption{The visual results of our local stylizer. The video on the left depicts the original video, while the video on the right displays the edited videos. \emph{Best viewed with Acrobat Reader. Click the images to play the animation clips.}}
\label{figure:local_stylization}
\end{figure*}

We use the same controlnet architecture as in our global stylizer~\ref{sec_app_global_stylizer}. The difference between two models mainly lies in the base model and input condition. For our local stylizer, we utilize the CogVideoX-2B model as the base. As shown in Figure~\ref{fig:local_stylization2}, the main branch takes the original video latents, masked video latents, and mask latents as input (48 channels). To mitigate the inflated channel dimension, we initialize our patch embedder using the first 16 channels from CogVideoX-2B, while the remaining 32 channels are zero-initialized. Similarly, the patch embedder for control branch are also zero-initialized.

Our control branch consists of 6 DiT blocks copied from main branch. For training data, we use the mask and phrases in the training dataset. We then combine the phrase with some pronouns randomly, to compose them as a sentence for training.
We trained our local stylizer for 1 epoch, with a batch size of 32, AdamW optimizer~\citep{adamw_loshchilov2017decoupled}, a learning rate of 1e-5, and a weight decay of 1e-4. The training videos consist of 33 frames at a resolution of 
$336\times592$. Similarly, to preserve generalization ability and accelerate training, we freeze the FFN layers except for the first DiT block.

For inference, we use a classifier-free guidance scale of 6. The inference process completes within 1 minute on an Nvidia RTX 4090 for a video with a resolution of $336\times592\times33$. We prepend the sentence prefix ``\textit{It's}'' to the detected object phrase and pronouns to form a complete prompt. For example, when we want to paint the house in the video to yellow, we should use the prompt: ``\textit{It's a yellow house.}''

Table~\ref{tab:comparison} presents a quantitative comparison of local stylization methods. Our expert model achieves the lowest Ewarp (6.50), indicating minimal warping artifacts, and the highest CLIPScore (0.2944), ensuring strong text alignment. It also attains the best Temporal Consistency (0.9828), preserving coherence across frames. For background preservation, our model outperforms all baselines with the highest PSNR (28.29) and SSIM (0.9843), signifying better structural similarity to the original background. The lowest LPIPS (0.0346) and MSE (108.25) further confirm minimal distortion. While InsV2V performs competitively, it falls short in CLIPScore and background fidelity. These results highlight our model's effectiveness in maintaining both stylization quality and background consistency.

\begin{figure*}[!ht]
    \centering
    \includegraphics[width=0.925\linewidth]{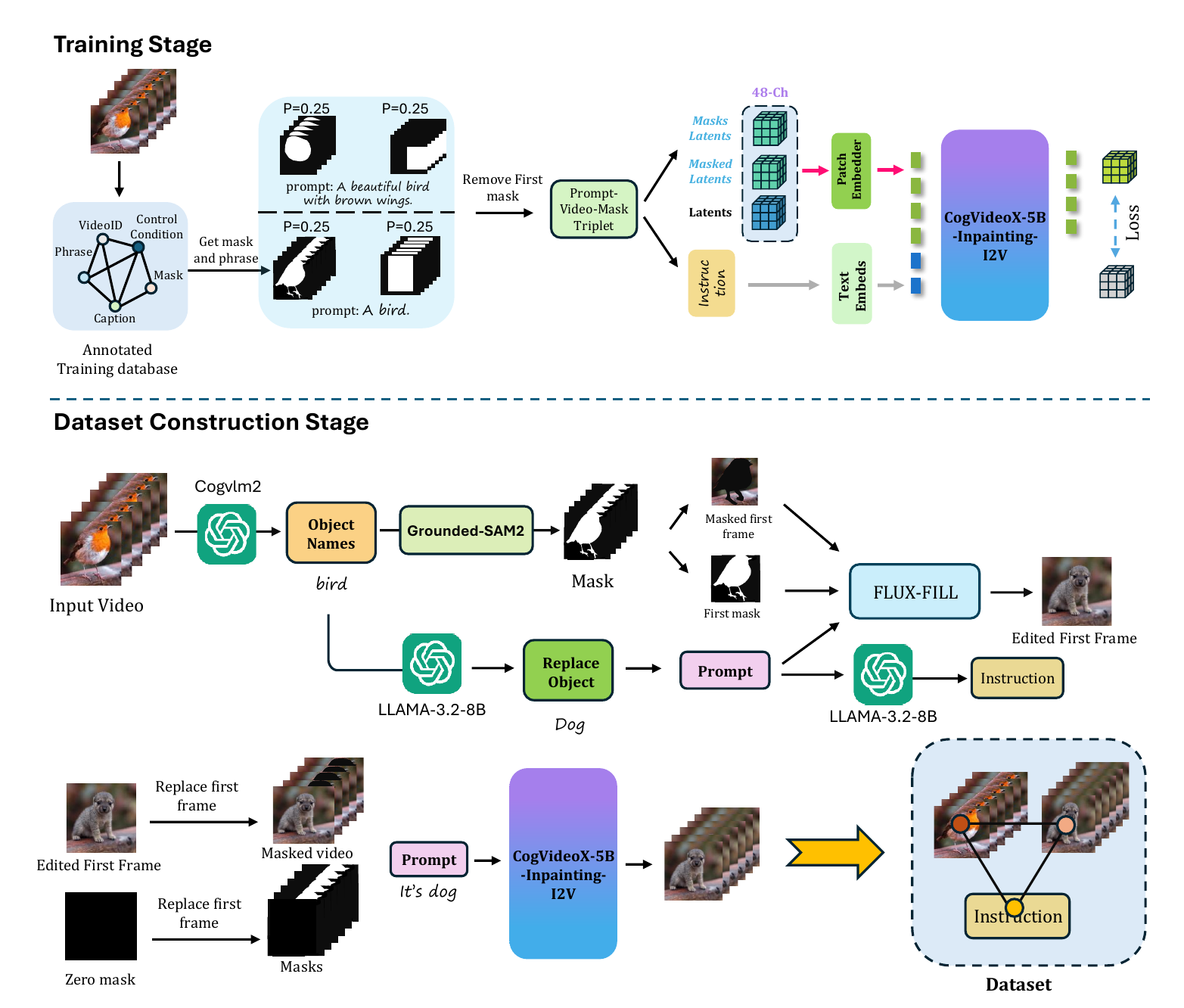}
    \caption{Top: The training pipeline of our inpainter. Bottom: The data construction pipeline for Señorita-2M using our inpainter.}
    \label{fig:inpainter}
\end{figure*}

\subsection{The Construction of Text-Guided Video Inpainter}
\label{sec_app_video_inpainter}

\begin{table*}[!ht]
\centering
\caption{Quantitative Comparison on Object Swap. The best results are \textbf{boldfaced}.}
\begin{tabular}{c|ccccccc} 
\toprule

\textbf{Methods} & Ewarp$(10^{-3})$ ($\downarrow$)& CLIPScore ($\uparrow$)& Temp-Cons ($\uparrow$)& PSNR ($\uparrow$)& SSIM ($\uparrow$)& LPIPS ($\downarrow$)& MSE ($\downarrow$)\\
\hline
Tokenflow & 17.21 & 0.3028 & 0.9752 & 18.70 & 0.8569 & 0.1447& 995.91 \\ 
Flatten & 17.91 & 0.2223 & 0.9744 & 18.80 & 0.8572 & 0.1350 & 1090.39 \\ 
InsV2V & \textbf{8.80} & 0.2733 & 0.9722 & 21.57 & 0.9204 & 0.0787 & 642.44 \\ 
AnyV2V & 13.49 & 0.2870 & 0.9741 & 19.78 & 0.8903 & 0.1197 & 777.86 \\
\hline
Our Expert & 12.06 & \textbf{0.3186} & \textbf{0.9782} & \textbf{25.59} & \textbf{0.9620} & \textbf{0.04} & \textbf{265.15}  \\ 

\bottomrule
\end{tabular} 
\label{tab:obj_swap_comparison} 
\end{table*}

Although many studies have explored text-guided video inpainting, such as AVID~\citep{avid_zhang2023avid} and COCOCO~\citep{avid_zhang2023avid}, most of these methods rely on outdated video foundation models, such as AnimateDiff~\citep{animatediff_guo2023animatediff}. Consequently, the generated videos often exhibit noticeable artifacts and inconsistencies. Recently,~\citet{vivid_10m_hu2024vivid} proposed the VIVID model, which trains an inpainter based on CogVideoX-5B-I2V. Unfortunately, their inpainter has not been open-sourced. Similar with the methods proposed by~\citet{vivid_10m_hu2024vivid}, we use the first-frame edited by a stable and well-performed image editor Flux-Fill to guide the inpainting process. 

\begin{figure*}[!ht]
  \centering
    \begin{subfigure}{0.485\textwidth}
      \animategraphics[width=1\textwidth]{8}{data/obj_swap/animal/best/6853331-hd_2048_1080_25fps_mask_id_1_0_0_guidance_scale_6.0_output79_org/image_}{0}{15}
    \end{subfigure}
    \begin{subfigure}{0.485\textwidth}
      \animategraphics[width=1\textwidth]{8}{data/obj_swap/animal/best/6853331-hd_2048_1080_25fps_mask_id_1_0_0_guidance_scale_6.0_output79/image_}{0}{15}
    \end{subfigure}

    \begin{subfigure}{0.24\textwidth}
      \animategraphics[width=1\textwidth]{8}{data/obj_swap/food/4978863-hd_1280_720_30fps_mask_id_1_0_0_guidance_scale_6.0_output22_org/image_}{0}{15}
    \end{subfigure}
    \begin{subfigure}{0.24\textwidth}
      \animategraphics[width=1\textwidth]{8}{data/obj_swap/food/4978863-hd_1280_720_30fps_mask_id_1_0_0_guidance_scale_6.0_output22/image_}{0}{15}
    \end{subfigure}
    \begin{subfigure}{0.24\textwidth}
      \animategraphics[width=1\textwidth]{8}{data/obj_swap/thing/7053533-hd_1280_720_60fps_mask_id_1_0_0_guidance_scale_6.0_output102_org/image_}{0}{15}
    \end{subfigure}
    \begin{subfigure}{0.24\textwidth}
      \animategraphics[width=1\textwidth]{8}{data/obj_swap/thing/7053533-hd_1280_720_60fps_mask_id_1_0_0_guidance_scale_6.0_output102/image_}{0}{15}
    \end{subfigure}

    \begin{subfigure}{0.24\textwidth}
      \animategraphics[width=1\textwidth]{8}{data/obj_swap/person/6306123-hd_1280_720_25fps_mask_id_1_0_0_guidance_scale_6.0_output55_org/image_}{0}{15}
    \end{subfigure}
    \begin{subfigure}{0.24\textwidth}
      \animategraphics[width=1\textwidth]{8}{data/obj_swap/person/6306123-hd_1280_720_25fps_mask_id_1_0_0_guidance_scale_6.0_output55/image_}{0}{15}
    \end{subfigure}
    \begin{subfigure}{0.24\textwidth}
      \animategraphics[width=1\textwidth]{8}{data/obj_swap/person/5046023-hd_1280_720_25fps_mask_id_1_0_0_guidance_scale_6.0_output74_org/image_}{0}{15}
    \end{subfigure}
    \begin{subfigure}{0.24\textwidth}
      \animategraphics[width=1\textwidth]{8}{data/obj_swap/person/5046023-hd_1280_720_25fps_mask_id_1_0_0_guidance_scale_6.0_output74/image_}{0}{15}
    \end{subfigure}

    \begin{subfigure}{0.24\textwidth}
      \animategraphics[width=1\textwidth]{8}{data/obj_swap/food/7185875-hd_1080_1920_30fps_mask_id_3_0_0_guidance_scale_6.0_output73_org/image_}{0}{15}
    \end{subfigure}
    \begin{subfigure}{0.24\textwidth}
      \animategraphics[width=1\textwidth]{8}{data/obj_swap/food/7185875-hd_1080_1920_30fps_mask_id_3_0_0_guidance_scale_6.0_output73/image_}{0}{15}
    \end{subfigure}
    \begin{subfigure}{0.24\textwidth}
      \animategraphics[width=1\textwidth]{8}{data/obj_swap/animal/best/17735534-hd_1080_1920_60fps_mask_id_2_0_0_guidance_scale_6.0_output192_org/image_}{0}{15}
    \end{subfigure}
    \begin{subfigure}{0.24\textwidth}
      \animategraphics[width=1\textwidth]{8}{data/obj_swap/animal/best/17735534-hd_1080_1920_60fps_mask_id_2_0_0_guidance_scale_6.0_output192/image_}{0}{15}
    \end{subfigure}

    \caption{The visual results of our inpainter. The video on the left depicts the original video, while the video on the right displays the edited videos. \emph{Best viewed with Acrobat Reader. Click the images to play the animation clips.}}
    
    \label{figure:object_swap}
    
\end{figure*}

The difference between VIVID and our inpainter lies in the following aspects. 
For training mask selection, we employ masks with random positions and shapes. We observed that the model tends to overfit to specific mask shapes during inpainting. To mitigate this, we generate masks with either random shapes or rectangles with varying aspect ratios in the first frame and periodically shift their locations in the subsequent frames. For both types of masks, we use video captions as prompts. Additionally, we introduce object-covering masks to enhance the model’s learning capacity. These masks are categorized into two types: (1) precise masks detected by Grounded-SAM2~\citep{groundingdino_liu2023grounding, sam2_ravi2024sam2} and (2) rectangular masks expanded from these precise masks. These masks are paired with structured prompts, which consist of pronouns and detected phrases, for training. Further details are provided in Figure~\ref{fig:inpainter}. Another key difference is in patch embedder initialization. Specifically, we initialize the first 16 channels of the patch embedder using parameters from the original patch embedders, while the remaining channels are zero-initialized.

For training, we set the first frame of the mask sequence to zeros to utilize the guidance of the edited image. The inpainter is initialized with the parameters of CogVideoX-5B-I2V. Unlike global stylizer methods, our inpainter does not require a control branch, allowing for a larger batch size. We trained for 1 epoch on our expert dataset with AdamW optimizer~\citep{adamw_loshchilov2017decoupled}, batch size of 16 and a learning rate of 1e-5. The resolution used during training was $336\times592$, and the number of frames was 33, the stride is 2. We freeze all FFN layers except for the first DiT block.

During inference, we input the prepared prompts, dilated precise masks, and videos to generate the inpainted video. The first frame is edited by Flux-Fill with a new object name. The new object name are generated by LLM. We use the classifier-free guidance of 6. The inference process can be finished on an Nvidia RTX 4090 GPU within 2 minutes, 33 frames and resolution of $336\times 592$.

Table~\ref{tab:obj_swap_comparison} compares different methods for object swap. Our expert model achieves the highest CLIPScore (0.3186) and Temporal Consistency (0.9782), indicating strong text alignment and frame coherence. Although InsV2V has the lowest Ewarp (8.80), suggesting minimal warping artifacts, it underperforms in CLIPScore, indicating poor alignment with text instructions. This discrepancy arises because InsV2V often fails to follow the given text instructions and does not successfully swap the object. As a result, many failure cases closely resemble the original video, leading to a lower warping error but also a lower CLIPScore.

For background preservation, our model outperforms all baselines, achieving the highest PSNR (25.59) and SSIM (0.9620), ensuring superior structural similarity. The lowest LPIPS (0.04) and MSE (265.15) further indicate minimal distortion. While AnyV2V and Tokenflow show competitive results, they fall short in maintaining both object swap fidelity and background consistency. These results demonstrate that our model effectively balances object replacement accuracy with background preservation.

\begin{figure*}[!ht]
    \centering
    \includegraphics[width=0.98\linewidth]{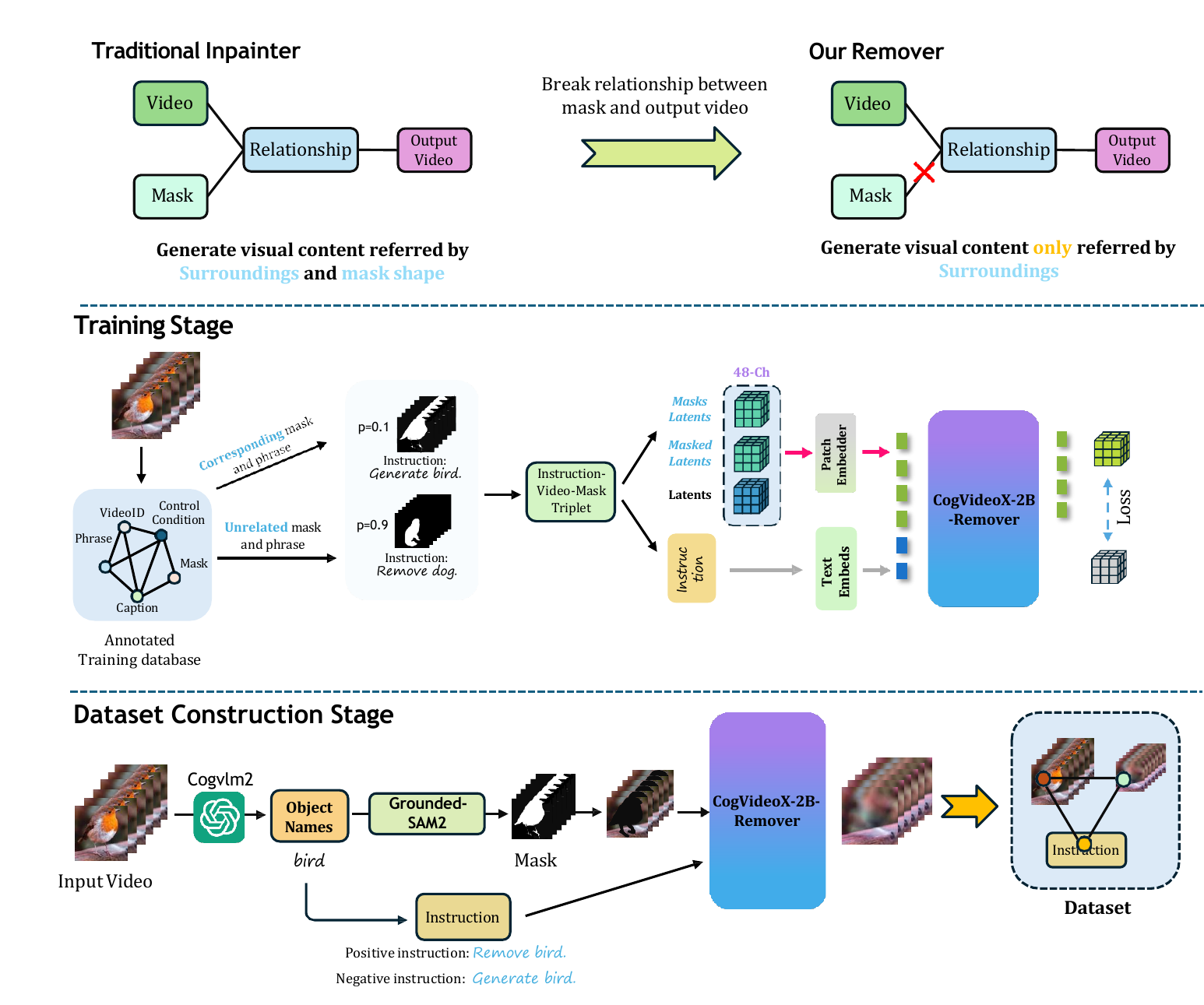}
    \caption{The framework of our remover and sub-dataset construction pipeline.}
    \label{fig:remover}
\end{figure*}

\begin{table*}[!ht]
\centering
\caption{Quantitative Comparison of Object Removal. To assess the performance of object removal, we calculate the CLIP similarity between the removal instruction and the edited video, denoted as \textbf{relevance}. A \textcolor{red}{lower} relevance score indicates \textcolor{red}{better} removal performance. The best results are \textbf{blodfaced}.}

\begin{tabular}{c|ccccccc} 
\toprule

\textbf{Methods} & Ewarp$(10^{-3})$ ($\downarrow$)& Relevance ($\downarrow$) & Temp-Cons ($\uparrow$) & PSNR ($\uparrow$) & SSIM ($\uparrow$) & LPIPS ($\downarrow$) & MSE ($\downarrow$) \\
\hline
Tokenflow & 16.34 & 0.1597 & 0.9786 & 18.38 & 0.8395 & 0.1639 & 1095.06 \\ 
Flatten & 11.18 & 0.2194 & 0.9759 & 18.87 & 0.8367 & 0.1529 & 1088.33 \\ 
InsV2V & 6.67 & 0.2134 & 0.9747 & 22.27 & 0.9187 & 0.0648 & 563.17 \\ 
AnyV2V & 13.14 & 0.1774 & 0.9765 & 19.80 & 0.8825 & 0.1290 & 800.56 \\ 
Propainter & 4.93 & 0.1685 & 0.9862 & \textbf{36.87} & \textbf{0.9978} & \textbf{0.0081} & \textbf{16.37} \\ 
\hline
Our Expert & \textbf{4.21} & \textbf{0.1554} & \textbf{0.9864} & 29.16 & 0.9863 & 0.031 & 89.62  \\ 

\bottomrule
\end{tabular} 
\label{tab:object_removal_comparison} 
\end{table*}

\subsection{The Construction of Remover}
\label{sec_app_remover}

\begin{figure*}[!ht]
  \centering

    \begin{subfigure}{0.49\textwidth}
      \animategraphics[width=1\textwidth]{8}{data/obj_removal/obj-3/4321851-hd_1280_720_25fps_mask_id_1_0_0_guidance_scale_2.0_output1961_org/image_}{0}{15}
    \end{subfigure}
    \begin{subfigure}{0.49\textwidth}
      \animategraphics[width=1\textwidth]{8}{data/obj_removal/obj-3/4321851-hd_1280_720_25fps_mask_id_1_0_0_guidance_scale_2.0_output1961/image_}{0}{15}
    \end{subfigure}

    \begin{subfigure}{0.24\textwidth}
      \animategraphics[width=1\textwidth]{8}{data/obj_removal/obj-1/3764305-hd_1280_720_30fps_mask_id_1_0_0_guidance_scale_2.0_output996_org/image_}{0}{15}
    \end{subfigure}
    \begin{subfigure}{0.24\textwidth}
      \animategraphics[width=1\textwidth]{8}{data/obj_removal/obj-1/3764305-hd_1280_720_30fps_mask_id_1_0_0_guidance_scale_2.0_output996/image_}{0}{15}
    \end{subfigure}
    \begin{subfigure}{0.24\textwidth}
      \animategraphics[width=1\textwidth]{8}{data/obj_removal/obj-2/4265399-hd_1280_720_30fps_mask_id_1_0_0_guidance_scale_2.0_output1080_org/image_}{0}{15}
    \end{subfigure}
    \begin{subfigure}{0.24\textwidth}
      \animategraphics[width=1\textwidth]{8}{data/obj_removal/obj-2/4265399-hd_1280_720_30fps_mask_id_1_0_0_guidance_scale_2.0_output1080/image_}{0}{15}
    \end{subfigure}

    \begin{subfigure}{0.24\textwidth}
      \animategraphics[width=1\textwidth]{8}{data/obj_removal/obj-11/6293014-hd_2048_1080_25fps_mask_id_1_0_0_guidance_scale_2.0_output1485_org/image_}{0}{15}
    \end{subfigure}
    \begin{subfigure}{0.24\textwidth}
      \animategraphics[width=1\textwidth]{8}{data/obj_removal/obj-11/6293014-hd_2048_1080_25fps_mask_id_1_0_0_guidance_scale_2.0_output1485/image_}{0}{15}
    \end{subfigure}
    \begin{subfigure}{0.24\textwidth}
      \animategraphics[width=1\textwidth]{8}{data/obj_removal/obj-10/6151144-hd_2048_1080_30fps_mask_id_3_0_0_guidance_scale_2.0_output1102_org/image_}{0}{15}
    \end{subfigure}
    \begin{subfigure}{0.24\textwidth}
      \animategraphics[width=1\textwidth]{8}{data/obj_removal/obj-10/6151144-hd_2048_1080_30fps_mask_id_3_0_0_guidance_scale_2.0_output1102/image_}{0}{15}
    \end{subfigure}

    \begin{subfigure}{0.24\textwidth}
      \animategraphics[width=1\textwidth]{8}{data/obj_removal/obj-8/5220276-hd_1280_720_25fps_mask_id_1_0_0_guidance_scale_2.0_output1803_org/image_}{0}{15}
    \end{subfigure}
    \begin{subfigure}{0.24\textwidth}
      \animategraphics[width=1\textwidth]{8}{data/obj_removal/obj-8/5220276-hd_1280_720_25fps_mask_id_1_0_0_guidance_scale_2.0_output1803/image_}{0}{15}
    \end{subfigure}
    \begin{subfigure}{0.24\textwidth}
      \animategraphics[width=1\textwidth]{8}{data/obj_removal/obj-9/5614850-hd_1280_720_25fps_mask_id_1_0_0_guidance_scale_2.0_output855_org/image_}{0}{15}
    \end{subfigure}
    \begin{subfigure}{0.24\textwidth}
      \animategraphics[width=1\textwidth]{8}{data/obj_removal/obj-9/5614850-hd_1280_720_25fps_mask_id_1_0_0_guidance_scale_2.0_output855/image_}{0}{15}
    \end{subfigure}

    \begin{subfigure}{0.24\textwidth}
      \animategraphics[width=1\textwidth]{8}{data/obj_removal/obj-9/5766010-hd_2048_1080_30fps_mask_id_2_0_0_guidance_scale_2.0_output1737_org/image_}{0}{15}
    \end{subfigure}
    \begin{subfigure}{0.24\textwidth}
      \animategraphics[width=1\textwidth]{8}{data/obj_removal/obj-9/5766010-hd_2048_1080_30fps_mask_id_2_0_0_guidance_scale_2.0_output1737/image_}{0}{15}
    \end{subfigure}
    \begin{subfigure}{0.24\textwidth}
      \animategraphics[width=1\textwidth]{8}{data/obj_removal/obj-11/6785358-hd_1280_720_25fps_mask_id_2_0_0_guidance_scale_2.0_output1061_org/image_}{0}{15}
    \end{subfigure}
    \begin{subfigure}{0.24\textwidth}
      \animategraphics[width=1\textwidth]{8}{data/obj_removal/obj-11/6785358-hd_1280_720_25fps_mask_id_2_0_0_guidance_scale_2.0_output1061/image_}{0}{15}
    \end{subfigure}

    \caption{The visual results of our object remover. The video on the left depicts the original video, while the video on the right displays the edited videos. \emph{Best viewed with Acrobat Reader. Click the images to play the animation clips.}}

    \label{figure:object_remover}
\end{figure*}

Traditional video inpainter, such as Propainter~\citep{zhou2023propainter}, uses optical flow to guide the completion. However, these methods show weaker performance than diffusion model~\citep{video_diffusion_inpainter_lee2024video}. Inpainters, such as CoCoCo~\citep{cococo_zi2024cococo}, AVID~\citep{avid_zhang2023avid} are designed to add objects. Recently, a new inpainting method, namely VIVID~\citep{vivid_10m_hu2024vivid} are designed to add, modify and remove video objects. We fully explored the CoCoCo and found it performs bad on object removal, since it has a high percentage to generate the object in the masked region, similar to the mask shape. To overcome this drawback, we design a training paradigm to break the correlation between generated content and mask shape. 

As shown in Figure~\ref{fig:remover}, our remover is trained by assuming that the input video contains objects from unrelated videos. The model is provided with an arbitrary mask from another video and learns to remove the assumed object while generating the object in the input video. Specifically, we randomly sample a mask and phrase from other video and used this mask to remove regions from the given video. We take 90\% unrelated masks with instruction ``\textit{Remove \textit{\{object name\}}}'', and 10\% masks corresponding to the input videos with instruction ``\textit{Generate \textit{\{object name\}}}'' This can be viewed as we use mask and the generate instruction corresponding to the input video as negative condition. During inference, the classifier-free guidance will steer the generation away from the negative condition, thus achieving the object removal.

We train the remover on our expert dataset for 1 epoch with AdamW optimizer, a batch size of 32, a learning rate of 1e-5, and a weight decay of 1e-4. For data sampling, we selected 90\% of the samples as task-irrelevant masks and 10\% as task-relevant masks. The video was sampled at 33 frames with a stride of 2, and the resolution was set to $336 \times 592$. Our Remover is built upon the CogVideoX-2B model and initialized with its pre-trained parameters. Similarly, to preserve generalization ability and accelerate training, we freeze the FFN layers except for the first DiT block.

During inference, we use classifier-free guidance scale of 2, the positive prompt is ``\textit{Remove \{object name\}}'', while the negative prompt is ``\textit{Generate \{object name\}}''. The frame number is 33 and the resolution of $336\times 592$. The removal process can be finished within 1 minute on an Nvidia RTX 4090 GPU. 

Table~\ref{tab:object_removal_comparison} compares different methods for object removal. Our expert model achieves the lowest Ewarp (4.21) and Relevance (0.1554), indicating minimal warping artifacts and strong removal effectiveness. Additionally, it attains the highest Temporal Consistency (0.9864), ensuring smooth and stable object removal across frames.

\begin{figure*}[!htp]
    \centering
    \begin{minipage}{0.02\textwidth}
        \rotatebox{90}{\small \centering \ \ \ \ Ours \ \ \ \ \ \ \ \ \ \ \ \ \ \ \ \ \ \ \ \ \ \ Propainter}
    \end{minipage}
    \begin{minipage}{0.90\textwidth}
        
        \begin{minipage}{0.98\textwidth}
        \includegraphics[width=0.32\textwidth]{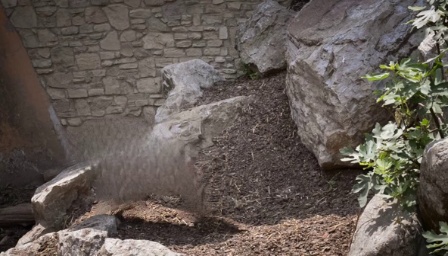}
        \includegraphics[width=0.32\textwidth]{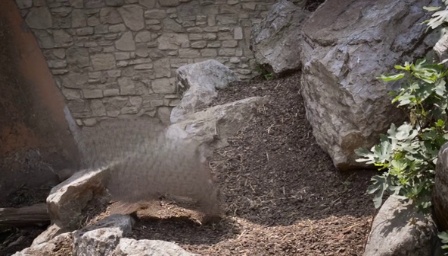}
        \includegraphics[width=0.32\textwidth]{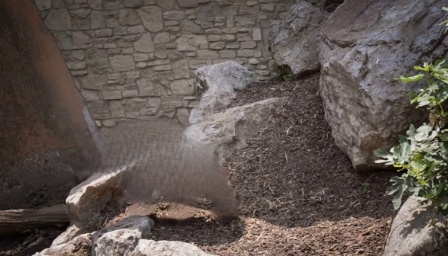}
        \end{minipage}

        \begin{minipage}{0.98\textwidth}
        \includegraphics[width=0.32\textwidth]{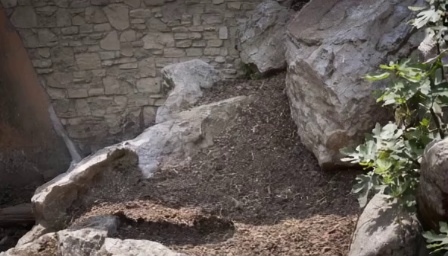}
        \includegraphics[width=0.32\textwidth]{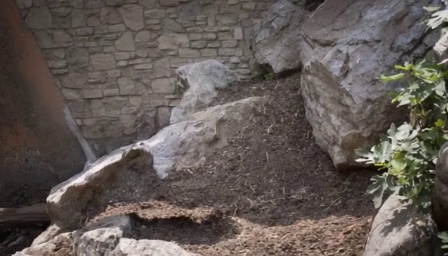}
        \includegraphics[width=0.32\textwidth]{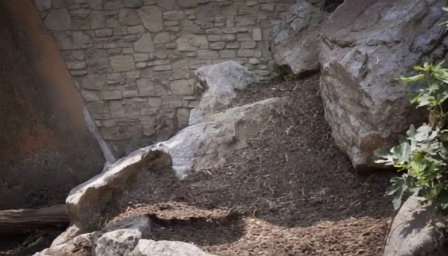}
        \end{minipage}
    \end{minipage}

    \begin{minipage}{0.02\textwidth}
        \rotatebox{90}{\small \centering \ \ \ \ Ours \ \ \ \ \ \ \ \ \ \ \ \ \ \ \ \ \ \ \ \ \ \ Propainter}
    \end{minipage}
    \begin{minipage}{0.90\textwidth}
    
    \begin{minipage}{0.98\textwidth}
        \includegraphics[width=0.32\textwidth]{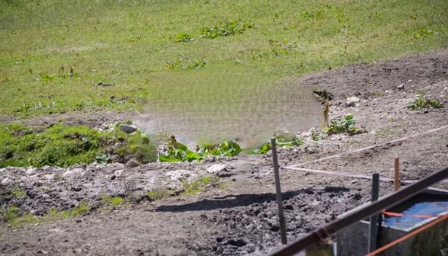}
        \includegraphics[width=0.32\textwidth]{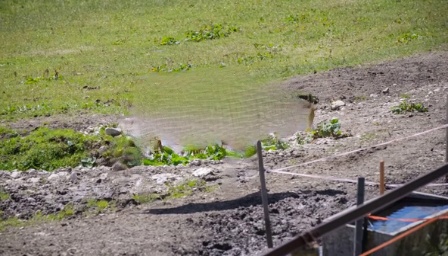}
        \includegraphics[width=0.32\textwidth]{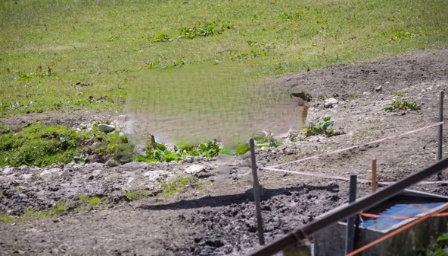}
    \end{minipage}
    
    \begin{minipage}{0.98\textwidth}
        \includegraphics[width=0.32\textwidth]{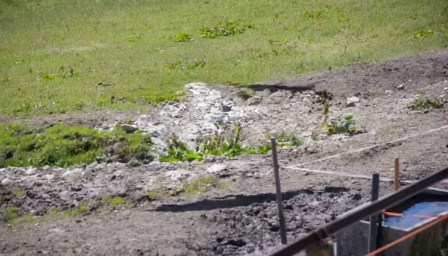}
        \includegraphics[width=0.32\textwidth]{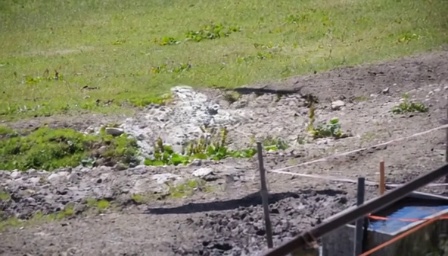}
        \includegraphics[width=0.32\textwidth]{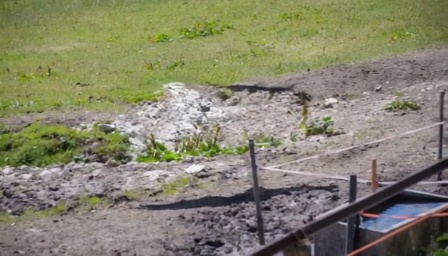}
    \end{minipage}
    \end{minipage}

    \begin{minipage}{0.02\textwidth}
        \rotatebox{90}{\small \centering \ \ \ \ Ours \ \ \ \ \ \ \ \ \ \ \ \ \ \ \ \ \ \ \ \ \ \ Propainter}
    \end{minipage}
    \begin{minipage}{0.90\textwidth}
        \begin{minipage}{0.98\textwidth}
        \includegraphics[width=0.32\textwidth]{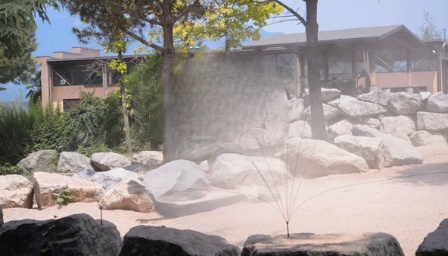}
        \includegraphics[width=0.32\textwidth]{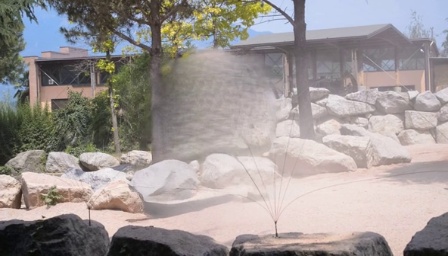}
        \includegraphics[width=0.32\textwidth]{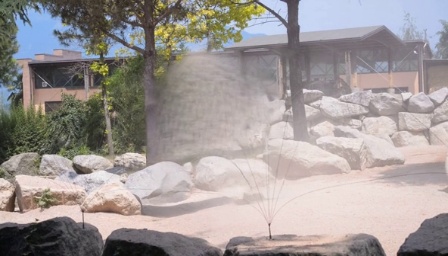}
        \end{minipage}
    
        \begin{minipage}{0.98\textwidth}
        \includegraphics[width=0.32\textwidth]{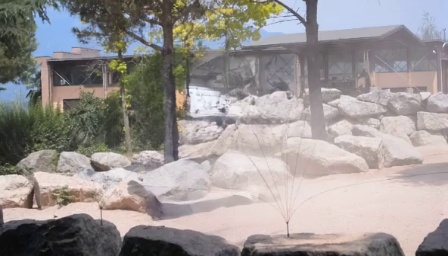}
        \includegraphics[width=0.32\textwidth]{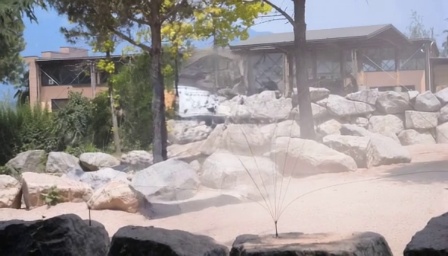}
        \includegraphics[width=0.32\textwidth]{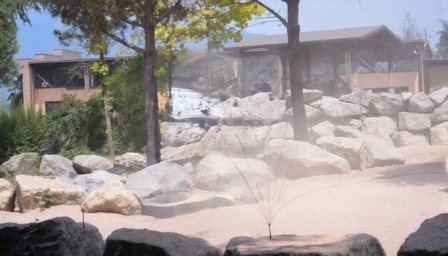}
        \end{minipage}
    \end{minipage}

    \begin{minipage}{0.02\textwidth}
        \rotatebox{90}{\small \centering \ \ \ \ Ours \ \ \ \ \ \ \ \ \ \ \ \ \ \ \ \ \ \ \ \ \ \ Propainter}
    \end{minipage}
    \begin{minipage}{0.90\textwidth}
        \begin{minipage}{0.98\textwidth}
        \includegraphics[width=0.32\textwidth]{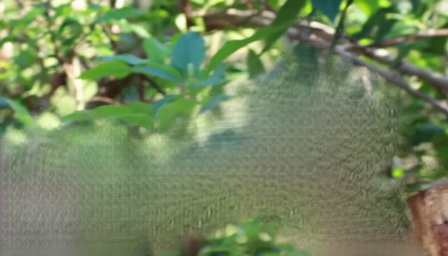}
        \includegraphics[width=0.32\textwidth]{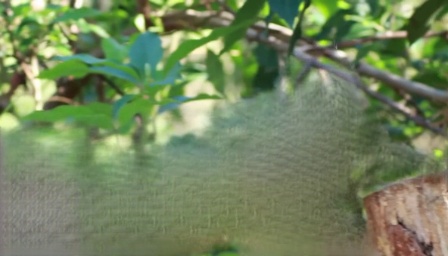}
        \includegraphics[width=0.32\textwidth]{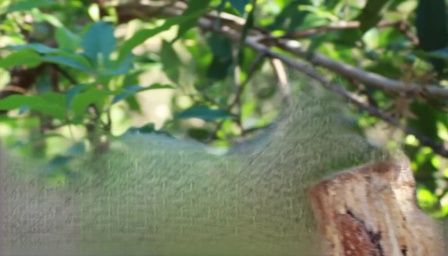}
        \end{minipage}
    
        \begin{minipage}{0.98\textwidth}
        \includegraphics[width=0.32\textwidth]{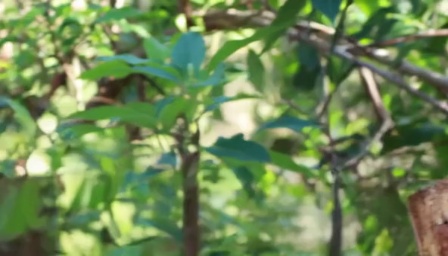}
        \includegraphics[width=0.32\textwidth]{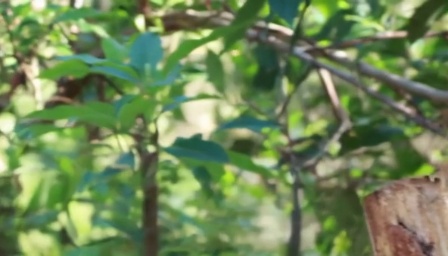}
        \includegraphics[width=0.32\textwidth]{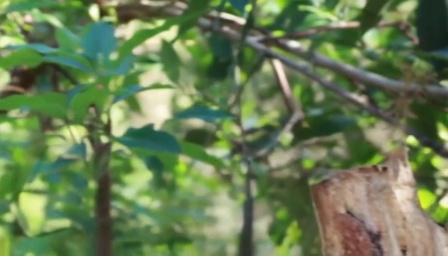}
        \end{minipage}
    
    \end{minipage}

    \caption{Editing results compared between different editing methods.}
    \label{fig:propainter_comparison}
\end{figure*}

While Propainter achieves exceptionally high PSNR (36.87) and SSIM (0.9978) with the lowest LPIPS (0.0081) and MSE (16.37), this is primarily because it does not alter background pixels. However, its object removal performance is poor, as the removed regions appear significantly blurry, which can be observed in qualitative examples~\ref{fig:propainter_comparison}. In contrast, our model effectively balances object removal with background preservation, maintaining both strong visual quality and semantic alignment.

\section{Construction of Señorita-2M Dataset}
\label{sec_app_senorita_2m_dataset}
\subsection{Source Data Collection}
\label{sec_data_collection}
We selected videos from Pexels~\citep{pexels}, a website that legally permits downloading and editing of videos. The dataset consists of a total of 388,909 videos. The resolution of these videos primarily ranges from 720p to 4K, with most videos containing more than 500 frames.

\subsection{Data Annotation}
\label{sec_data_annotation}
We use BLIP-2-opt-2B~\citep{blip2_li2023blip} to generate video captions while adhering to the length restrictions of CLIP~\citep{clip_ramesh2021zero}. For object recognition, we utilize CogVLM-video-llama3-chat~\citep{Cogvlm2_hong2024cogvlm2} with INT8 precision for efficient inference on Nvidia RTX 4090 GPUs. We set the maximum token length to 120 and use six frames per video. The videos and detected object names are then fed into Grounded-SAM2. Specifically, we employ the SWinB\_CogCoor model from Grounding-DINO~\citep{groundingdino_liu2023grounding} and the SAM2\_hiera\_tiny model from SAM2~\citep{sam2_ravi2024sam2}. As a result, the dataset contains approximately 800,000 mask sequences.

\subsection{Construction of Global Editing Video Pairs}
\label{sec_dataset_construction_global_editing}
Our global edit comprises several key components. First part is style transfer, we use 290 types of style prompt advised by Midjourney~\citep{midjourney}. To further enhance the object localization ability of the diffusion, we use the videos and masks given by Grounded-SAM2~\citep{groundingdino_liu2023grounding, sam2_ravi2024sam2} to compose video pairs. Besides, we also use the other control conditions (hed, depth, canny, etc.) and videos to make video pairs.

\subsubsection{Style Transfer}
As shown in Figure~\ref{fig:global-style-transfer}, we use ControlNet-SD1.5 to edit the first frame. Specifically, we append the style prompt to the captions to compose the new prompt. This prompt then is used for the style transfer for the first frame. After getting the first edited frame, we use this frame along with hed, depth, canny conditions and the new prompt to craft the rest frames. To accelerate inference, we reduce the resolution from $336\times 592$ to $256\times 448$, with $\times$2 times inference cost reduce. After inference, we upscale the frame to $336\times 592$. To get the instructions, we ask LLM~\citep{llama3_dubey2024llama} to generate instructions, by giving some examples. We use the original videos as the source videos, while the edited videos as the target videos, along with the instructions to build a (source, target, instruction) video editing triplet.

\subsubsection{Object Grounding.} 

We provide video pairs for object grounding, detected by Grounded-SAM2, to help the video editor accurately identify relevant regions in the video based on specific instructions. All areas unrelated to the input prompt are masked in black, while different object instances corresponding to the prompt are highlighted in distinct colors. To construct the initial instruction, we prepend words such as "Detect" or "Ground" before the object name. Finally, we use a LLM to refine and enhance these instructions. The resolution of the videos is $1120\times1984$, with 64 frames per sequence.

\subsubsection{Conditional generation.} This section comprises 10 tasks designed to aid video editors in video-to-video translation: Deblur, Canny-to-Video, Depth-to-Video, Depth Detection, Hed-to-Video, Hed Detection, Upscaling, FakeScribble-to-Video, FakeScribble Detection, and Colorization. For the deblur task, Gaussian blur is applied to create blurred source videos, with the original videos serving as target videos. Similarly, the tasks of Canny-to-Video, Depth-to-Video, Hed-to-Video, FakeScribble-to-Video, and Colorization use Canny, depth, hed, fake scribble, and grayscale videos as sources, while original videos are the targets. Conversely, the tasks of Depth Detection, Hed Detection, and FakeScribble Detection use controllable video conditions as target videos. The resolution of these tasks is $1120\times 1984$, 64 frames.

\subsection{Construction of Local Stylization}
\label{sec_dataset_construction_local_editing}
\subsubsection{Local Style Transfer}

As shown in Figure~\ref{fig:local_stylization2}, we use LLM~\citep{llama3_dubey2024llama} to modify the appearance, color, or style of detected phrases and convert them into prompts. These prompts, along with the three masked control conditions and the masked video, are used for object stylization. We use 33 frames and $336\times 592$ as the input resolution. For the instruction construction, we also use the LLM to transform the prompt to instruction, by showing some transfer examples to it. We use the original videos as the source videos, while the edited videos as the target videos, along with the instructions to build a (source, target, instruction) video editing triplet.

\subsubsection{Object Swap}
We use LLM~\citep{llama3_dubey2024llama} to find the a new object that has similar shape. With the new object, we use the Flux-Fill to edit first frame. With the guidance of the first frame, we use our inpainter to generate the rest frames.
The original videos are used to serve as the target videos, while the edited videos are used for source videos. We use LLM to generate the instructions by using both original and swapped object name. The prompts for instruction construction are shown in~\ref{prompts}.

\subsubsection{Object Removal and Addition}

We use object remover to remove the object from the videos. By taking both positive instructions and negative instructions with CFG of 2, we thus remove the object from the videos. The input frame number is 33, the resolution is $336\times 592$. For the object removal, the original videos are used as source videos, while the edited videos are used as the target videos. For the object addition, the edited videos are used as source videos, while the original videos are used as the target videos. We use LLM to generate instructions for two tasks.

\subsubsection{Video Inpainting and Outpainting}
We constructed approximately 60,000 video pairs to enhance the model's capabilities in video inpainting and outpainting. The masked regions are set to black, with pixel values of zero, while the unmasked regions remain unchanged. Both the video inpainting and outpainting processes have a resolution of $1280\times 1984$, with 64 frames.

\begin{tcolorbox}[colback=white,colframe=black!75!white,title=The Prompts for Instruction Construction]
\label{prompts}

  The prompt used for Global Stylization: \\
  \textit{Help me find the instruction of <input>. Don't give useless information, such as ``There be''. For example, <input> is ``sci-fi futurism, sleek spaceships, glowing cities, alien landscapes, advanced technology, cinematic visuals'', the answer is ``make it sci-fi futurism.''. <input> is ``warframe, warframe style, video game art style, the art created in warframe style'', the answer is ``make it chick warframe style.''. Don't give me descriptions. Please give me answer directly. Now, the <input> is ``\{style\_prompt\}'', the answer is:}\\

  The prompt used for Local Stylization: \\
  \textit{Help me find the instruction of <input>. Don't give useless information, such as ``There be''. For example, <input> is ``bird -> yellow bird'', the answer is ``make the bird yellow.''. <input> is ``chick -> green chick'', the answer is ``make the chick green.''. <input> is ``fox -> brown and furry fox'', the answer is ``make the fox brown and furry''. The <input> is ``pigeons -> gray pigeons'', the answer is ``make pigeons gray.''. Don't give me descriptions. Please give me answer directly.}
  \textit{Now, the <input> is ``\{object\_name\}''  ->  ``\{ text\_prompt \}'', the answer is:}\\

  The prompt used for Object Removal: \\
  \textit{Help me enhance the <input>. Don't give useless information, such as ``There be''. For example, <input> is ``Remove dog'', the answer is ``Delete the dog.''. <input> is ``Remove dog'', the answer is ``Remove the dog from this video.''. <input> is ``Remove dog'', the answer is ``Discard the dog from videos.''. <input> is ``Remove dog'', the answer is ``Eliminate the dog.''. Don't give me descriptions. Please give me answer directly. Now, the <input> is ``Remove a \{object\_name\}'', the answer is:}\\

  The prompt used for Object Addition: \\
  \textit{Help me enhance the <input>. Don't give useless information, such as ``There be''. For example, <input> is ``Add dog'', the answer is ``Insert a dog.''. <input> is ``Add dog\", the answer is ``Place a dog.''. <input> is ``Add dog\", the answer is ``Add a dog to this video.''. I will give you a negative instruction, <input> is ``Add dog'', the answer is ``Install a helicopter pad.''. (That's wrong) Don't give me descriptions. Please give me answer directly. Now, the <input> is ``Add a \{object\_name\}'', the answer is:}\\
  
  The prompt used for Object Swap: \\
  \textit{Help me rewrite the <input>. Don't change its meaning. Don't give useless information, such as ``There be''. For example, <input> is ``replace cat with dog.'', the answer is ``turn cat into dog.''. <input> is ``turn cat into dog.'', the answer is ``change the cat to dog''. <input> is ``replace cat with dog.'', the answer is ``Let there be a dog in the place of the cat.''. Don't give me descriptions. Please give me answer directly. Now, the <input> is ``Turn \{target\_name\} into \{object\_name\}'', the answer is:}
\end{tcolorbox}

\section{Data Selection and Cleaning}
\label{sec_data_selection}

We propose a comprehensive filtering pipeline to select high-quality, successfully edited videos. The process begins with a quality filter to identify successful edits. Next, videos with poor text alignment are detected and removed based on text-video similarity. Finally, videos that remain unchanged or show only minor modifications are excluded by comparing the original and edited versions using CLIP~\citep{clip_radford2021learning}.

\subsection{Quality Filtering}
\label{sec_quality_filtering}
Due to the existence of corrupted and failure cases in the generated samples, we train a quality classifier and propose a pipeline to filter out these failed samples. 

\textbf{Construction of training and validation set}.
We selected 5,000 edited videos and manually annotated approximately 1,000 of them as failed samples, while the remaining videos were labeled as successful samples. Additionally, we constructed a validation set consisting of 120 successful samples and 60 failed samples.

\textbf{The training details of quality classifier}.
We employed the vision encoder of CLIP, specifically ViT-Huge~\citep{clip_radford2021learning}, to extract video features. Frame-wise features were extracted from 17 frames per video. We utilized both the CLS token and the pooler output as two distinct feature representations. Based on these features, we trained two separate classifiers and subsequently combined them into an ensemble model to enhance classification performance. Each classifier comprises three MLP layers with ReLU activation functions applied between layers.

\textbf{The inference of quality classifier}. We apply a threshold of 0.6 to the classifiers, retaining samples with confidence scores above this threshold. Notably, for the object addition task, we set a slightly lower threshold, as the target videos correspond to the original videos and are not influenced by the quality of the edited outputs.

\subsection{Filtering Poor Text-alignment Videos}
\label{sec_poor_text_alignment}
Although some edited videos exhibit good visual quality, the content in the edited regions may be unrelated to the text prompt. Retaining these videos compromises dataset quality and hinders the effective training of video editors. To address this issue, we utilize CLIP to measure the similarity between edited videos and their corresponding text prompts.

We apply different similarity thresholds for different tasks. For global stylization, we compare the style prompt with the edited video. Since the prompt used in this comparison lacks detailed information, we set a lower threshold of 0.2. Additionally, for object swap and local stylization, edits are applied within a masked region, whereas text-video similarity is computed across the entire video, leading to a lower similarity score. To account for this discrepancy, we set thresholds of 0.2 and 0.22, respectively. For object removal, we rely on instructions rather than descriptive prompts. As a result, we do not apply similarity-based filtering for this task.

\subsection{Filtering Unchanged Video Pairs}
\label{sec_quality_unchanged_video_pairs}
After applying the aforementioned filtering methodologies, we retained videos with high text alignment and superior visual quality. However, some edited videos may contain only subtle modifications compared to the original footage. This issue arises due to factors such as small edited regions or the generation of visually similar content within masked areas.

To address this, we utilize the CLIP vision encoder to extract features and conduct a frame-by-frame comparison between the original and edited videos. Video pairs with a similarity score exceeding 0.95 are subsequently removed.

\section{Style Prompts}
\label{sec_style_prompts}
\begin{tcolorbox}[colback=white,colframe=black!75!white,title=Style Prompts]
  Michelangelo, Michelangelo style, Renaissance style, the art created by Michelangelo\\
Monet, Monet style, Impressionist style, the art created by Monet\\
Paul Cézanne, Cézanne style, Post-Impressionist style, the art created by Paul Cézanne\\
Mark Rothko, Rothko style, Abstract Expressionist style, the art created by Mark Rothko\\
Paul Klee, Klee style, Abstract style, Bauhaus style, the art created by Paul Klee\\
Picasso, Picasso style, Cubist style, the art created by Picasso\\
Piet Mondrian, Mondrian style, De Stijl style, the art created by Piet Mondrian\\
Pierre-Auguste Renoir, Renoir style, Impressionist style, the art created by Pierre-Auguste Renoir\\
Rembrandt, Rembrandt style, Baroque style, the art created by Rembrandt\\
René Magritte, Magritte style, Surrealist style, the art created by René Magritte\\
Roy Lichtenstein, Lichtenstein style, Pop Art style, the art created by Roy Lichtenstein\\
Salvador Dalí, Dalí style, Surrealist style, the art created by Salvador Dalí\\
Sandro Botticelli, Botticelli style, Early Renaissance style, the art created by Sandro Botticelli\\
Takashi Murakami, Murakami style, Superflat style, the art created by Takashi Murakami\\
Van Gogh, Van Gogh style, Post-Impressionist style, the oil painting style, the oil painting created by Van Gogh\\
Wassily Kandinsky, Kandinsky style, Abstract style, Bauhaus style, the art created by Wassily Kandinsky\\
Mat Collishaw, Collishaw style, Contemporary Art style, the art created by Mat Collishaw\\
Yayoi Kusama, Kusama style, Contemporary Art style, Pop Art style, the art created by Yayoi Kusama\\
Igor Morski, Morski style, Surrealist style, Fantasy Art style, the art created by Igor Morski\\
Shinkai Makoto, Shinkai style, Anime style, Cinematic style, the art created by Shinkai Makoto\\
Pixar, Pixar style, 3D Animation style, CGI style, the animation created by Pixar\\
Kyoto Animation, Kyoto Animation style, Anime style, the animation created by Kyoto Animation\\
Jerry Pinkney, Pinkney style, Illustration style, Children's Books style, the illustrations created by Jerry Pinkney\\
Hayao Miyazaki, Miyazaki style, Anime style, Ghibli style, the animation created by Hayao Miyazaki\\
Beatrix Potter, Potter style, Illustration style, Children's Books style, the illustrations created by Beatrix Potter\\
Jon Klassen, Klassen style, Children's Books style, Illustration style, the illustrations created by Jon Klassen\\
Kay Sage, Sage style, Surrealist style, the art created by Kay Sage\\
Jeffrey Catherine Jones, Jones style, Fantasy Art style, Illustration style, the art created by Jeffrey Catherine Jones\\
Yaacov Agam, Agam style, Kinetic Art style, Op Art style, the art created by Yaacov Agam\\
David Hockney, Hockney style, Pop Art style, Contemporary Art style, the art created by David Hockney\\
Victor Moscoso, Moscoso style, Psychedelic Art style, Graphic Art style, the art created by Victor Moscoso\\
Raphaelite, Pre-Raphaelite Brotherhood style, the art created by Raphaelite\\
Stefan Koid, Koid style, Contemporary Art style, the art created by Stefan Koid\\
Sui Ishida, Ishida style, Manga style, the art created by Sui Ishida\\
Swoon, Swoon style, Street Art style, Contemporary Art style, the art created by Swoon\\
Tasha Tudor, Tudor style, Illustration style, Children's Books style, the illustrations created by Tasha Tudor\\
Tintoretto, Tintoretto style, Mannerism style, Late Renaissance style, the art created by Tintoretto\\
Theodore Robinson, Robinson style, Impressionist style, the art created by Theodore Robinson\\
Titian, Titian style, Renaissance style, the art created by Titian\\
WLOP, WLOP style, Digital Art style, Fantasy Art style, the art created by WLOP\\
Yanjun Cheng, Cheng style, Contemporary Art style, the art created by Yanjun Cheng\\
Yoji Shinkawa, Shinkawa style, Video Game Art style, Concept Art style, the art created by Yoji Shinkawa\\
Alena Aenami, Aenami style, Digital Art style, the art created by Alena Aenami\\
Anton Fadeev, Fadeev style, Concept Art style, Digital Art style, the art created by Anton Fadeev\\
Charlie Bowater, Bowater style, Concept Art style, Digital Art style, the art created by Charlie Bowater\\
Cory Loftis, Loftis style, Concept Art style, Digital Art style, the art created by Cory Loftis\\
Fenghua Zhong, Zhong style, Digital Art style, Illustration style, the art created by Fenghua Zhong\\
Greg Rutkowski, Rutkowski style, Digital Painting style, Fantasy Art style, the art created by Greg Rutkowski
\end{tcolorbox}
\begin{tcolorbox}[colback=white,colframe=black!75!white,title=Style Prompts]
Anton Pieck, Pieck style, Illustration style, Fairy Tale Art style, the art created by Anton Pieck\\
Carl Barks, Barks style, Comic Book Art style, the art created by Carl Barks\\
Alphonse Mucha, Mucha style, Art Nouveau style, the art created by Alphonse Mucha\\
Andy Warhol, Warhol style, Pop Art style, the art created by Andy Warhol\\
Banksy, Banksy style, Street Art style, Contemporary Art style, the art created by Banksy\\
Francisco de Goya, Goya style, Romanticism style, the art created by Francisco de Goya\\
Caravaggio, Caravaggio style, Baroque style, the art created by Caravaggio\\
Diego Rivera, Rivera style, Muralism style, the art created by Diego Rivera\\
Marc Chagall, Chagall style, Modern Art style, Surrealist style, the art created by Marc Chagall\\
Edgar Degas, Degas style, Impressionist style, the art created by Edgar Degas\\
Eugène Delacroix, Delacroix style, Romanticism style, the art created by Eugène Delacroix\\
Francis Bacon, Bacon style, Expressionist style, Modern Art style, the art created by Francis Bacon\\
Frida Kahlo, Kahlo style, Surrealist style, Modern Art style, the art created by Frida Kahlo\\
Gerald Brom, Brom style, Dark Fantasy Art style, the art created by Gerald Brom\\
Gustav Klimt, Klimt style, Symbolist style, Art Nouveau style, the art created by Gustav Klimt\\
Henri Matisse, Matisse style, Fauvist style, the art created by Henri Matisse\\
J.M.W. Turner, Turner style, Romanticism style, the art created by J.M.W. Turner\\
Jack Kirby, Kirby style, Comic Book Art style, the art created by Jack Kirby\\
Jackson Pollock, Pollock style, Abstract Expressionist style, the art created by Jackson Pollock\\
Johannes Vermeer, Vermeer style, Baroque style, the art created by Johannes Vermeer\\
Jean-Michel Basquiat, Basquiat style, Neo-Expressionist style, the art created by Jean-Michel Basquiat\\
Marcel Duchamp, Duchamp style, Dada style, the art created by Marcel Duchamp\\
Traditional Chinese Ink Painting, Chinese Art style, the art created in Traditional Chinese Ink Painting style\\
Japanese Ukiyo-e, Ukiyo-e style, Japanese Art style, the art created in Japanese Ukiyo-e style\\
Japanese comics/manga, Manga style, the art created in Japanese comics/manga style\\
Stock illustration style, Illustration style, the illustrations created in Stock illustration style\\
CGSociety, CGSociety style, Digital Art style, CGI style, the art created by CGSociety\\
DreamWorks Pictures, DreamWorks style, 3D Animation style, CGI style, the animation created by DreamWorks Pictures\\
Fashion, Fashion Illustration style, Runway Art style, the art created in Fashion style\\
Poster of Japanese graphic design, Japanese Graphic Design style, the art created in Poster of Japanese graphic design style\\
90s video game, Retro Game Art style, the art created in 90s video game style\\
French art, Various Styles (Impressionism, Romanticism, etc.), the art created in French art style\\
Bauhaus, Bauhaus style, Modernist Art style, the art created in Bauhaus style\\
Anime, Anime style, Japanese Animation style, the art created in Anime style\\
Pixel Art, Pixel Art style, Digital Art style, Retro Art style, the art created in Pixel Art style\\
Vintage, Vintage style, Retro Art style, the art created in Vintage style\\
Pulp Noir, Pulp Noir style, Pulp Art style, Noir Art style, the art created in Pulp Noir style\\
Country style, Folk Art style, the art created in Country style\\
Abstract, Abstract style, Abstract Art style, the art created in Abstract style\\
Risograph, Risograph style, Printmaking style, Graphic Art style, the art created in Risograph style\\
Graphic, Graphic style, Graphic Design style, the art created in Graphic style\\
Ink render, Ink render style, Ink Art style, the art created in Ink render style\\
Ethnic Art, Ethnic Art style, Folk Art style, the art created in Ethnic Art style\\
Retro dark vintage, Retro dark vintage style, Gothic Art style, Dark Art style, the art created in Retro dark vintage style\\
Traditional Chinese Ink Painting style, Chinese Art style, the art created in Traditional Chinese Ink Painting style\\
Steampunk, Steampunk style, Steampunk Art style, the art created in Steampunk style\\
Film photography, Film photography style, Photography style, the photographs taken in Film photography style\\
Concept art, Concept art style, Conceptual Art style, the art created in Concept art style
\end{tcolorbox}
\begin{tcolorbox}[colback=white,colframe=black!75!white,title=Style Prompts]
Gothic gloomy, Gothic gloomy style, Gothic Art style, the art created in Gothic gloomy style\\
Realism, Realism style, Realist Art style, the art created in Realism style\\
Black and white, Black and white style, Monochrome Art style, the art created in Black and white style\\
Unity Creations, Unity Creations style, Digital Art style, CGI style, the art created by Unity Creations\\
Baroque, Baroque style, Baroque Art style, the art created in Baroque style\\
Impressionism, Impressionist style, the art created in Impressionism style\\
Art Nouveau, Art Nouveau style, the art created in Art Nouveau style
Rococo, Rococo style, Rococo Art style, the art created in Rococo style\\
Adrian Donohue, Donohue style, Photography style, the photographs taken by Adrian Donohue\\
Adrian Tomine, Tomine style, Comic Art style, Illustration style, the art created by Adrian Tomine\\
Akihiko Yoshida, Yoshida style, Video Game Art style, Concept Art style, the art created by Akihiko Yoshida\\
Akira Toriyama, Toriyama style, Manga style, Anime style, the art created by Akira Toriyama\\
Cai Guo-Qiang, Cai style, Contemporary Art style, the art created by Cai Guo-Qiang\\
Drew Struzan, Struzan style, Poster Art style, Illustration style, the art created by Drew Struzan\\
Hans Arp, Arp style, Dada style, Abstract Art style, the art created by Hans Arp\\
Ilya Kuvshinov, Kuvshinov style, Manga style, Anime style, the art created by Ilya Kuvshinov\\
James Jean, Jean style, Illustration style, Fine Art style, the art created by James Jean\\
Jasmine Becket-Griffith, Becket-Griffith style, Pop Surrealism style, the art created by Jasmine Becket-Griffith\\
Jean Giraud, Giraud style, Comic Art style, Illustration style, the art created by Jean Giraud\\
Partial anatomy, Anatomical Art style, the art created in Partial anatomy style\\
Color ink on paper, Ink Art style, the art created with color ink on paper\\
Doodle, Doodle style, Illustration style, Sketch Art style, the art created in Doodle style\\
Voynich manuscript, Manuscript Art style, Historical Art style, the art created in Voynich manuscript style\\
Book page, Book page style, Illustration style, Typography Art style, the art created in Book page style\\
Realistic, Realism style, the art created in Realistic style\\
3D, 3D Art style, CGI style, the art created in 3D style\\
Sophisticated, Fine Art style, the art created in Sophisticated style\\
Photoreal, Photorealism style, the art created in Photoreal style\\
Character concept art, Character concept art style, Concept Art style, the art created in Character concept art style\\
Renaissance, Renaissance style, Renaissance Art style, the art created in Renaissance style\\
Fauvism, Fauvist style, the art created in Fauvism style\\
Cubism, Cubist style, the art created in Cubism style\\
Abstract Art, Abstract Art style, the art created in Abstract Art style\\
Surrealism, Surrealist style, the art created in Surrealism style\\
Op Art / Optical Art, Optical Art style, the art created in Op Art / Optical Art style\\
Victorian, Victorian style, Victorian Art style, the art created in Victorian style\\
Futuristic, Futuristic style, Sci-Fi Art style, the art created in Futuristic style\\
Minimalist, Minimalist style, the art created in Minimalist style\\
Brutalist, Brutalist style, the art created in Brutalist style\\
Constructivist, Constructivist style, the art created in Constructivist style\\
BOTW, BOTW style, Video Game Art style (Breath of the Wild), the art created in BOTW style\\
Warframe, Warframe style, Video Game Art style, the art created in Warframe style\\
Pokémon, Pokémon style, Anime style, Video Game Art style, the art created in Pokémon style\\
APEX, APEX style, Video Game Art style, the art created in APEX style\\
The Elder Scrolls, Elder Scrolls style, Video Game Art style, the art created in The Elder Scrolls style\\
From Software, From Software style, Video Game Art style, the art created by From Software\\
Detroit: Become Human, Detroit: Become Human style, Video Game Art style, the art created in Detroit: Become Human style\\
AFK Arena, AFK Arena style, Video Game Art style, the art created in AFK Arena style\\
Hong SoonSang, SoonSang style, Animation style, Concept Art style, the art created by Hong SoonSang
\end{tcolorbox}
\begin{tcolorbox}[colback=white,colframe=black!75!white,title=Style Prompts]
CookieRun Kingdom, CookieRun Kingdom style, Video Game Art style, the art created in CookieRun Kingdom style\\
League of Legends, League of Legends style, Video Game Art style, the art created in League of Legends style\\
Jojo's Bizarre Adventure, Jojo's Bizarre Adventure style, Manga style, Anime style, the art created in Jojo's Bizarre Adventure style\\
Makoto Shinkai, Shinkai style, Anime style, Cinematic style, the art created by Makoto Shinkai\\
Poster of Japanese graphic design, Japanese Graphic Design style, the art created in Poster of Japanese graphic design style\\
90s video game, Retro Game Art style, the art created in 90s video game style\\
French art, Various Styles (Impressionism, Romanticism, etc.), the art created in French art style\\
Bauhaus, Bauhaus style, Modernist Art style, the art created in Bauhaus style\\
Anime, Anime style, Japanese Animation style, the art created in Anime style\\
Pixel Art, Pixel Art style, Digital Art style, Retro Art style, the art created in Pixel Art style\\
Vintage, Vintage style, Retro Art style, the art created in Vintage style\\
Pulp Noir, Pulp Noir style, Pulp Art style, Noir Art style, the art created in Pulp Noir style\\
Country style, Folk Art style, the art created in Country style\\
Abstract, Abstract style, Abstract Art style, the art created in Abstract style\\
Risograph, Risograph style, Printmaking style, Graphic Art style, the art created in Risograph style\\
Graphic, Graphic style, Graphic Design style, the art created in Graphic style\\
Ink render, Ink render style, Ink Art style, the art created in Ink render style\\
Ethnic Art, Ethnic Art style, Folk Art style, the art created in Ethnic Art style\\
Retro dark vintage, Retro dark vintage style, Gothic Art style, Dark Art style, the art created in Retro dark vintage style\\
Traditional Chinese Ink Painting style, Chinese Art style, the art created in Traditional Chinese Ink Painting style\\
Steampunk, Steampunk style, Steampunk Art style, the art created in Steampunk style\\
Film photography, Film photography style, Photography style, the photographs taken in Film photography style\\
Concept art, Concept art style, Conceptual Art style, the art created in Concept art style\\
Montage, Montage style, Collage Art style, the art created in Montage style\\
Full details, Full details style, Realism style, Hyperrealism style, the art created in Full details style\\
Gothic gloomy, Gothic gloomy style, Gothic Art style, the art created in Gothic gloomy style\\
Realism, Realism style, Realist Art style, the art created in Realism style\\
Black and white, Black and white style, Monochrome Art style, the art created in Black and white style\\
Unity Creations, Unity Creations style, Digital Art style, CGI style, the art created by Unity Creations\\
Baroque, Baroque style, Baroque Art style, the art created in Baroque style\\
Impressionism, Impressionist style, the art created in Impressionism style\\
Art Nouveau, Art Nouveau style, the art created in Art Nouveau style\\
Rococo, Rococo style, Rococo Art style, the art created in Rococo style\\
Adrian Donohue, Donohue style, Photography style, the photographs taken by Adrian Donohue\\
Adrian Tomine, Tomine style, Comic Art style, Illustration style, the art created by Adrian Tomine\\
Akihiko Yoshida, Yoshida style, Video Game Art style, Concept Art style, the art created by Akihiko Yoshida\\
Akira Toriyama, Toriyama style, Manga style, Anime style, the art created by Akira Toriyama\\
Cai Guo-Qiang, Cai style, Contemporary Art style, the art created by Cai Guo-Qiang\\
Drew Struzan, Struzan style, Poster Art style, Illustration style, the art created by Drew Struzan\\
Hans Arp, Arp style, Dada style, Abstract Art style, the art created by Hans Arp\\
Ilya Kuvshinov, Kuvshinov style, Manga style, Anime style, the art created by Ilya Kuvshinov\\
James Jean, Jean style, Illustration style, Fine Art style, the art created by James Jean\\
Jasmine Becket-Griffith, Becket-Griffith style, Pop Surrealism style, the art created by Jasmine Becket-Griffith\\
Jean Giraud, Giraud style, Comic Art style, Illustration style, the art created by Jean Giraud\\
Partial anatomy, Anatomical Art style, the art created in Partial anatomy style\\
Color ink on paper, Ink Art style, the art created with color ink on paper\\
Doodle, Doodle style, Illustration style, Sketch Art style, the art created in Doodle style\\
Voynich manuscript, Manuscript Art style, Historical Art style, the art created in Voynich manuscript style
\end{tcolorbox}
\begin{tcolorbox}[colback=white,colframe=black!75!white,title=Style Prompts]
Book page, Book page style, Illustration style, Typography Art style, the art created in Book page style\\
Realistic, Realism style, the art created in Realistic style\\
3D, 3D Art style, CGI style, the art created in 3D style\\
Sophisticated, Fine Art style, the art created in Sophisticated style\\
Photoreal, Photorealism style, the art created in Photoreal style\\
Character concept art, Character concept art style, Concept Art style, the art created in Character concept art style\\
Renaissance, Renaissance style, Renaissance Art style, the art created in Renaissance style\\
Fauvism, Fauvist style, the art created in Fauvism style\\
Cubism, Cubist style, the art created in Cubism style\\
Abstract Art, Abstract Art style, the art created in Abstract Art style
Surrealism, Surrealist style, the art created in Surrealism style\\
Op Art / Optical Art, Optical Art style, the art created in Op Art / Optical Art style\\
Futuristic, Futuristic style, Sci-Fi Art style, the art created in Futuristic style\\
Minimalist, Minimalist style, the art created in Minimalist style\\
Brutalist, Brutalist style, the art created in Brutalist style\\
Constructivist, Constructivist style, the art created in Constructivist style\\
BOTW, BOTW style, Video Game Art style (Breath of the Wild), the art created in BOTW style\\
Warframe, Warframe style, Video Game Art style, the art created in Warframe style\\
Pokémon, Pokémon style, Anime style, Video Game Art style, the art created in Pokémon style\\
APEX, APEX style, Video Game Art style, the art created in APEX style\\
The Elder Scrolls, Elder Scrolls style, Video Game Art style, the art created in The Elder Scrolls style\\
From Software, From Software style, Video Game Art style, the art created by From Software\\
Detroit: Become Human, Detroit: Become Human style, Video Game Art style, the art created in Detroit: Become Human style\\
AFK Arena, AFK Arena style, Video Game Art style, the art created in AFK Arena style\\
CookieRun Kingdom, CookieRun Kingdom style, Video Game Art style, the art created in CookieRun Kingdom style\\
League of Legends, League of Legends style, Video Game Art style, the art created in League of Legends style\\
Jojo's Bizarre Adventure, Jojo's Bizarre Adventure style, Manga style, Anime style, the art created in Jojo's Bizarre Adventure style\\
Makoto Shinkai, Shinkai style, Anime style, Cinematic style, the art created by Makoto Shinkai\\
Soejima Shigenori, Shigenori style, Video Game Art style, the art created by Soejima Shigenori\\
Yamada Akihiro, Akihiro style, Manga style, Anime style, the art created by Yamada Akihiro\\
Munashichi, Munashichi style, Concept Art style, Digital Art style, the art created by Munashichi\\
Watercolor Children's Illustration, Watercolor Children's Illustration style, Watercolor Art style, Children's Books style, the art created in Watercolor Children's Illustration style\\
Ghibli Studio, Ghibli style, Anime style, the animation created by Ghibli Studio\\
Stained Glass Window, Stained Glass style, the art created in Stained Glass Window style\\
Ink Illustration, Ink Illustration style, Ink Art style, the art created in Ink Illustration style\\
Miyazaki Hayao Style, Miyazaki style, Anime style, Ghibli style, the animation created in Miyazaki Hayao style\\
Vincent van Gogh, Van Gogh style, Post-Impressionist style, the oil painting style, the oil painting created by Van Gogh\\
Leonardo da Vinci, Da Vinci style, Renaissance style, the art created by Leonardo da Vinci\\
Manga, Manga style, the art created in Manga style\\
Pointillism, Pointillist style, the art created in Pointillism style\\
Claude Monet, Monet style, Impressionist style, the art created by Claude Monet\\
Johannes Itten, Itten style, Bauhaus style, the art created by Johannes Itten\\
John Harris, Harris style, Sci-Fi Art style, Illustration style, the art created by John Harris\\
Jon Klassen, Klassen style, Children's Books style, Illustration style, the art created by Jon Klassen\\
Junji Ito, Ito style, Horror Manga style, the art created by Junji Ito
Koe no Katachi, Koe no Katachi style, Anime style, Manga style, the art created in Koe no Katachi style
\end{tcolorbox}
\begin{tcolorbox}[colback=white,colframe=black!75!white,title=Style Prompts]
Osamu Tezuka, Tezuka style, Manga style, Anime style, the art created by Osamu Tezuka\\
Rob Gonsalves, Gonsalves style, Surrealist style, Magic Realism style, the art created by Rob Gonsalves\\
Sol LeWitt, LeWitt style, Minimalist style, Conceptual Art style, the art created by Sol LeWitt\\
Yusuke Murata, Murata style, Manga style, Anime style, the art created by Yusuke Murata\\
Antonio Mora, Mora style, Surrealist style, Photo Manipulation style, the art created by Antonio Mora\\
Yoji Shinkawa, Shinkawa style, Video Game Art style, Concept Art style, the art created by Yoji Shinkawa\\
National Geographic, National Geographic style, Photography style, the photographs taken for National Geographic\\
Hyperrealism, Hyperrealism style, the art created in Hyperrealism style
Cinematic, Cinematic style, Cinematic Art style, the art created in Cinematic style\\
Architectural Sketching, Architectural Sketching style, Architecture Art style, the art created in Architectural Sketching style\\
Clear Facial Features, Clear Facial Features style, Portrait Art style, the art created with Clear Facial Features\\
Interior Design, Interior Design style, Interior Art style, the art created in Interior Design style\\
Weapon Design, Weapon Design style, Concept Art style, the art created in Weapon Design style\\
Subsurface Scattering, Subsurface Scattering style, Digital Art style, CGI style, the art created with Subsurface Scattering\\
Game Scene Graph, Game Scene Graph style, Video Game Art style, the art created in Game Scene Graph style\\
Cyberpunk style, neon-lit dystopian cityscape, futuristic skyscrapers, dark rain-soaked streets, glowing holograms, cybernetic characters, high-tech and gritty, rebellion theme, vibrant colors, immersive detail\\
Ultra-detailed photorealistic image, realistic lighting and textures, high-resolution, cinematic quality\\
Cyberpunk style, neon-lit cityscape, futuristic tech, dark atmosphere, glowing holograms, high-tech low-life\\
Epic fantasy scene, magical landscapes, mythical creatures, intricate details, vibrant colors, ethereal lighting\\
Anime-style illustration, vibrant colors, dynamic poses, sharp line art, expressive characters, 2D aesthetics\\
Concept art, highly detailed environments, creative landscapes, futuristic design, cinematic lighting, imaginative visuals\\
Watercolor painting, soft textures, pastel colors, flowing brushstrokes, dreamy and artistic\\
Steampunk aesthetic, Victorian-era technology, brass and gears, intricate machinery, retro-futuristic design\\
Abstract art, vibrant colors, geometric patterns, fluid forms, modern artistic expression, minimalistic or chaotic\\
Noir style, black-and-white, dramatic shadows, moody atmosphere, vintage detective aesthetics\\
Pixel art, retro 8-bit style, vibrant blocky colors, low-resolution, game-like visuals, nostalgic charm\\
Professional studio portrait, dramatic lighting, high detail, shallow depth of field, realistic skin textures\\
Isometric perspective, detailed environments, vibrant colors, 3D-inspired flat design, intricate details\\
Baroque art, elaborate and ornate details, dramatic compositions, rich textures, classical European aesthetics\\
Dark fantasy setting, eerie atmosphere, mystical creatures, gothic architecture, muted tones, ominous lighting\\
Low poly 3D art, simplified geometric shapes, bright pastel colors, minimalist style, game aesthetic\\
Impressionist painting, soft brushstrokes, vivid colors, natural light, artistic and emotional style\\
Sci-fi futurism, sleek spaceships, glowing cities, alien landscapes, advanced technology, cinematic visuals\\
Pop art style, bold colors, comic book aesthetics, stylized patterns, retro 1960s look\\
Vaporwave style, retro-futuristic design, pastel neon colors, 1980s aesthetics, surreal landscapes\\
Surrealist art, dreamlike scenes, unexpected juxtapositions, imaginative landscapes, abstract and symbolic\\
Medieval-inspired style, illuminated manuscripts, intricate patterns, historical scenes, muted tones\\
Graffiti art, vibrant spray paint textures, urban street style, bold typography, dynamic and expressive\\
Art Nouveau style, flowing organic lines, floral motifs, intricate patterns, pastel and earthy tones\\
Cinematic lighting, moody atmosphere, dramatic shadows, high contrast, film-like quality\\
Minimalist design, clean and simple lines, muted colors, open space, modern and abstract\\
Retro-futurism, 1950s sci-fi style, sleek spaceships, vintage design, bold colors, nostalgic aesthetics\\
Fantasy map style, hand-drawn cartography, intricate details, parchment textures, medieval aesthetic\\
Glitch art, pixelated visuals, distorted and fragmented images, neon and dark tones, digital chaos\\
Nature photography, high detail, realistic textures, vibrant landscapes, soft natural light
\end{tcolorbox}
\begin{tcolorbox}[colback=white,colframe=black!75!white,title=Style Prompts]
Full details, Full details style, Realism style, Hyperrealism style, the art created in Full details style\\
Chibi-style characters, exaggerated cute proportions, vibrant colors, anime-inspired, playful and adorable\\
Dennis Stock, Stock style, Photography style, the photographs taken by Dennis Stock\\
Michal Lisowski, Lisowski style, Digital Art style, Illustration style, the art created by Michal Lisowski\\
Paul Lehr, Lehr style, Science Fiction Art style, Illustration style, the art created by Paul Lehr\\
Ross Tran, Tran style, Digital Art style, Concept Art style, the art created by Ross Tran\\
Montage, Montage style, Collage Art style, the art created in Montage style
\end{tcolorbox}
\end{document}